\documentclass[journal]{IEEEtran}

\usepackage{cite}
\usepackage{hyperref}
\usepackage[pdftex]{graphicx}
\usepackage{amsmath}
\usepackage[linesnumbered, ruled, vlined]{algorithm2e}
\usepackage{array}
\usepackage[caption=false,font=footnotesize]{subfig}
\usepackage{multirow}
\usepackage{xcolor}
\usepackage{soul}

\newtheorem{lemma}{Lemma}

\begin{document}

\title{Detecting and Identifying Optical Signal Attacks on Autonomous Driving Systems}

\author{Jindi~Zhang,
        Yifan~Zhang,
        Kejie~Lu,
        Jianping~Wang,
        Kui~Wu,
        Xiaohua~Jia,
        and~Bin~Liu%
\thanks{This work was supported in part by Hong Kong Research Grant Council under GRF 11200220, and in part by the National Science Foundation (NSF) under grant CNS-1730325.}%
\thanks{J. Zhang, Y. Zhang, J. Wang and X. Jia are with the Department of Computer Science, City University of Hong Kong, Hong Kong (e-mail: \href{mailto:jd.zhang@my.cityu.edu.hk}{jd.zhang@my.cityu.edu.hk}; \href{mailto:yif.zhang@my.cityu.edu.hk}{yif.zhang@my.cityu.edu.hk}; \href{mailto:jianwang@cityu.edu.hk}{jianwang@cityu.edu.hk}; \href{mailto:csjia@cityu.edu.hk}{csjia@cityu.edu.hk}).}%
\thanks{K. Lu is with the Department of Computer Science and Engineering, University of Puerto Rico at Mayag{\"u}ez, Mayag{\"u}ez, Puerto Rico 00682 (e-mail: \href{mailto:kejie.lu@upr.edu}{kejie.lu@upr.edu}).}%
\thanks{K. Wu is with the Department of Computer Science, University of Victoria, Victoria BC V8P 5C2, Canada (e-mail: \href{mailto:wkui@uvic.ca}{wkui@uvic.ca}).}%
\thanks{B. Liu is with the Department of Computer Science and Technology, Tsinghua University, Beijing, China (e-mail: \href{mailto:liub@tsinghua.edu.cn}{liub@tsinghua.edu.cn}).}%
\thanks{This article has been accepted for publication in a future issue of this journal, but has not been fully edited. Content may change prior to final publication. Citation information: DOI \url{10.1109/JIOT.2020.3011690}, IEEE Internet of Things Journal. 2327-4662 (c) 2021 IEEE. Personal use is permitted, but republication/redistribution requires IEEE permission. See \url{http://www.ieee.org/publications standards/publications/rights/index.html} for more information.}}

\maketitle

\begin{abstract}
For autonomous driving, an essential task is to detect surrounding objects accurately. To this end, most existing systems use optical devices, including cameras and light detection and ranging (LiDAR) sensors, to collect environment data in real-time. In recent years, many researchers have developed advanced machine learning models to detect surrounding objects. Nevertheless, the aforementioned optical devices are vulnerable to optical signal attacks, which could compromise the accuracy of object detection. To address this critical issue, we propose a framework to detect and identify sensors that are under attack. Specifically, we first develop a new technique to detect attacks on a system that consists of three sensors. Our main idea is to (1) use data from three sensors to obtain two versions of depth maps (i.e., disparity) and (2) detect attacks by analyzing the distribution of disparity errors. In our study, we use real datasets and the state-of-the-art machine learning model to evaluate our attack detection scheme and the results confirm the effectiveness of our detection method. Based on the detection scheme, we further develop an identification model that is capable of identifying up to $n-2$ attacked sensors in a system with one LiDAR and $n$ cameras. We prove the correctness of our identification scheme and conduct experiments to show the accuracy of our identification method. At last, we investigate the overall sensitivity of our framework.
\end{abstract}


\vspace{-0.1in}
\section{Introduction}

\IEEEPARstart{I}{n} the past few years, autonomous driving has attracted significant attention from both academia and industry. Recent advances in artificial intelligence and machine learning technologies enable accurate object and event detection and response (OEDR)~\cite{SAE18taxonomy}. The technology advances, together with great commercial potentials and incentives, quickly pushed the adoption of autonomous driving. For instance, Waymo launched a driverless taxi service in Arizona in 2018~\cite{davies18waymo}. Tesla announced that the full self-driving feature of its products would be available by the end of 2020~\cite{hawkins19tesla}.

To facilitate accurate OEDR tasks, autonomous vehicles (AVs) are usually equipped with a number of sensors, including GPS, inertial measurement unit, radar, sonar, camera, light detection and ranging (LiDAR), etc.~\cite{apollo19perception}. Among these sensors, optical devices (LiDAR and camera) have become more and more important because they can provide object detection in a large range and also because many emerging machine learning models proposed in the past few years can accurately measure the depth of objects and detect objects. Due to the importance of these optical devices, in this paper, we focus on their security aspects, particularly on the mitigation of potential attacks on these optical devices.

Despite the importance of the security of optical devices in an autonomous driving system, it was investigated only in a few previous studies. In~\cite{wyglinski13security,ren19security}, researchers summarized several categories of vulnerabilities in autonomous vehicles. In~\cite{petit15remote, yan16can, shin17illusion}, researchers demonstrated through experiments that LiDAR can be attacked by sending spoofed and/or delaying optical pulses. They also demonstrated that a camera can be blinded if it receives an intense light beam.

Although these pioneer studies are important, there is a lack of a comprehensive mechanism to detect and identify such attacks. In this paper, we propose a novel framework to tackle this important issue by (1) detecting the optical attacks using data from multiple sensors and (2) identifying the sensors that are under attack. To achieve accuracy in both detection and identification, there are two major challenges:

\begin{itemize}
\item
The optical signals can be processed by many advanced machine learning models, each of which can generate various features. Moreover, optical signal attack causes different consequences on camera images and LiDAR point clouds. Therefore, an appropriate type of feature needs to be chosen as the common ground where both attacks can be detected.

\item
The size and the position of the damaged area caused by optical signal attack in images and point clouds are unpredictable. The damaged area can appear anywhere in the sensor view. Detection method must perform fine-grained detection across the whole sensor view in order to be invariant to the size and position of the damaged area and distinguish the feature differences in non-attack scenarios and attack scenarios.

\end{itemize}

To address the first challenge, the proposed framework includes an optical attack detection method that extracts depth information (i.e., disparity) from two sets of sensor data respectively and then uses depth information as the common ground to detect attacks on both images and point clouds. To address the second challenge, our method detects attacks by analyzing the distribution of disparity errors that measure \emph{pixel-level} disparity inconsistencies in the whole sensor view. Thus, the detection method is robust.

The main contributions of this study can be summarized as follows:

\begin{itemize}
\item
We develop a new technique to detect optical attacks on a system that consists of three sensors, including two possible cases (1) one LiDAR and two cameras, or (2) three cameras. Specifically, we first use data from three sensors to obtain two versions of depth maps (i.e., disparity) and then detect attacks by analyzing the distribution of disparity errors. In our study, we use real datasets of KITTI~\cite{geiger12are, geiger13vision} and the state-of-the-art machine learning model PSMNet~\cite{chang18pyramid} to evaluate our attack detection scheme and the results confirm the effectiveness of our detection method.

\item
Based on the detection scheme, we further develop an identification model that is capable of identifying up to $n-2$ attacked sensors in a system with one LiDAR and $n$ cameras. In our study, we prove the correctness of our identification scheme and conduct experiments to show the accuracy of our identification method.

\item
At last, we investigate the sensitivity of our framework to optical attacks with more diverse settings. We use experiments to show its excellence in this aspect.

\end{itemize}

The rest of this paper is organized as follows. In Section~\ref{sec.relatedwork}, we first introduce the studies related to our work. In Section~\ref{sec.system}, we discuss the system models, including the optical sensors and attack models, and our attack mitigation framework. In Section~\ref{sec.detection}, we elaborate on the attack detection schemes. In Section~\ref{sec.identification}, we further investigate the attack identification issue. In Section~\ref{sec.sensitivity}, we examine the overall sensitivity of our framework. Finally, we conclude the paper in Section~\ref{sec.conclusion}.

\vspace{-0.1in}
\section{Related Work}
\label{sec.relatedwork}

The methods for attacking optical sensors (LiDAR and camera) gradually become more and more advanced. In surveys of studies~\cite{wyglinski13security,ren19security}, researchers introduced the vulnerability that perceptual sensors of AVs could be compromised via physical channels at a close distance. In~\cite{petit15remote}, the authors showed several effective and realistic methods to compromise a 2D LiDAR and a camera. Particularly, in their experiments, they managed to relay and spoof LiDAR signals, as well as blind the camera using strong light beams. Adversary attacks against camera with intensive lights were also studied in~\cite{yan16can} and even caused irrecoverable damages to the camera. Later, Shin~\textit{et al.} demonstrated the attacks against Velodyne VLP-16, one of the most popular top-sale 3D LiDARs in the market, by producing fake signals~\cite{shin17illusion}. Based on the previous work, the researchers dug even deeper in~\cite{cao19adversarial}, in which the authors designed an optimization-based strategy to produce more bogus dots to compromise a 3D LiDAR with a much higher success rate, and they constructed new attacking scenarios to study the impact on the decision making of AVs. Despite their importance, existing studies in optical attacks only give some rough countermeasure ideas, such as redundancy of sensors and randomization of LiDAR pulse.

In the literature, there are only a few studies for systematically defending optical sensors of AVs, but these studies focus on other types of sensor attack. For example, the authors of~\cite{rofail19multi} claimed that the attacks against a camera could be wise enough to erase only objects from pictures or modify their positions. By using an additional LiDAR as a reference, they proposed to extract object features from images and LiDAR point cloud, and then detect the attacks via mismatches of the two sets of features. In~\cite{changalvala19lidar}, Changalvala~\textit{et al.} investigated an internal attack that can tamper a point cloud from the inside of an AV system, and they tackled the detection problem by adding a watermark to the LiDAR point cloud. Different from~\cite{rofail19multi} and~\cite{changalvala19lidar}, our work targets at defending against the optical attacks on LiDAR and cameras of AVs. We follow the idea of redundancy of sensors and design a countermeasure framework that not only can detect optical attacks via analyzing the inconsistency of depth information (i.e., disparity) from different sources, but also can identify the compromised sensors of an AV system.

As for estimating depth using images, there are two main categories of algorithms: monocular-vision based and stereo-vision based. The current methods of the two categories all adopt deep neural networks, but the monocular ones consider the task as a dense matrix regression problem and focus on minimizing the error of predictions, while the stereo-vision based algorithms formulate it as a problem of matching pixels from two images~\cite{bhoi19monocular}. As a result, DORN~\cite{fu18deep}, one of the best monocular methods, can only predict depth with an error of around $9\%$. In contrast, as a representative algorithm of the latter category, the state-of-the-art PSMNet~\cite{chang18pyramid} achieves the task with an error that is less than $2\%$. In this study, we choose PSMNet over other methods because of its better accuracy.

\vspace{-0.1in}
\section{System Models}
\label{sec.system}

In this section, we first explain the main optical sensors in an autonomous driving system and their normal operations. We then elaborate on the attack models on LiDAR and camera, with some numerical results that illustrate the impacts of optical attacks. Finally, we briefly explain the main idea of the proposed framework for attack detection and identification.

\vspace{-0.15in}
\subsection{Optical Sensors}
\label{subsec:architecture}

In this paper, we consider a general autonomous driving system, and we focus on the optical devices, in particular, LiDAR and camera.

\subsubsection{LiDAR}

A LiDAR sensor can send and receive specific optical pulses in certain directions. By comparing the incoming reflected signals with the transmitted ones, LiDAR can provide an accurate estimation of the distance between the LiDAR and an object in a specific direction. The output of LiDAR consists of a set of points in 3D space, which is known as a point cloud. By clustering these points, the object detection models applied in AV systems can locate obstacles in the real world.

\subsubsection{Camera}

Cameras are very common in existing autonomous driving systems. AVs are usually equipped with more than two of them. The produced images are useful to several perception functions, such as obstacle detection and road/lane detection.

Specifically, similar to human eyes, two cameras can be used to form a stereo vision system that can estimate the depth of an object. As a simple example, if a real-world point is captured at pixel $P_{l}=(x_{l},y)$ in the left image and at pixel $P_{r}=(x_{r},y)$ in the right image, then the \textit{disparity} $d$ is defined as $d=x_{l}-x_{r}$. We can calculate the depth $z$ using
\begin{equation}\label{eq:triangulation}
z=f\times\frac{b}{d},
\end{equation}
where $f$ is the focal length and $b$ is the distance between the two cameras. 

In general, we can obtain the depth of a real-world point using disparity, once the pair of corresponding pixels ($P_{l}$ and $P_{r}$) are located in two images. Therefore, the main goal of depth estimation algorithms, such as PSMNet~\cite{chang18pyramid}, is to identify pairs of pixels in two images that are corresponding to the same real-world points. Finally, a \textit{disparity map} is generated by computing disparity for every pixel in an image.

\begin{figure}[!t]
    \centering
    \includegraphics[width=0.48\textwidth]{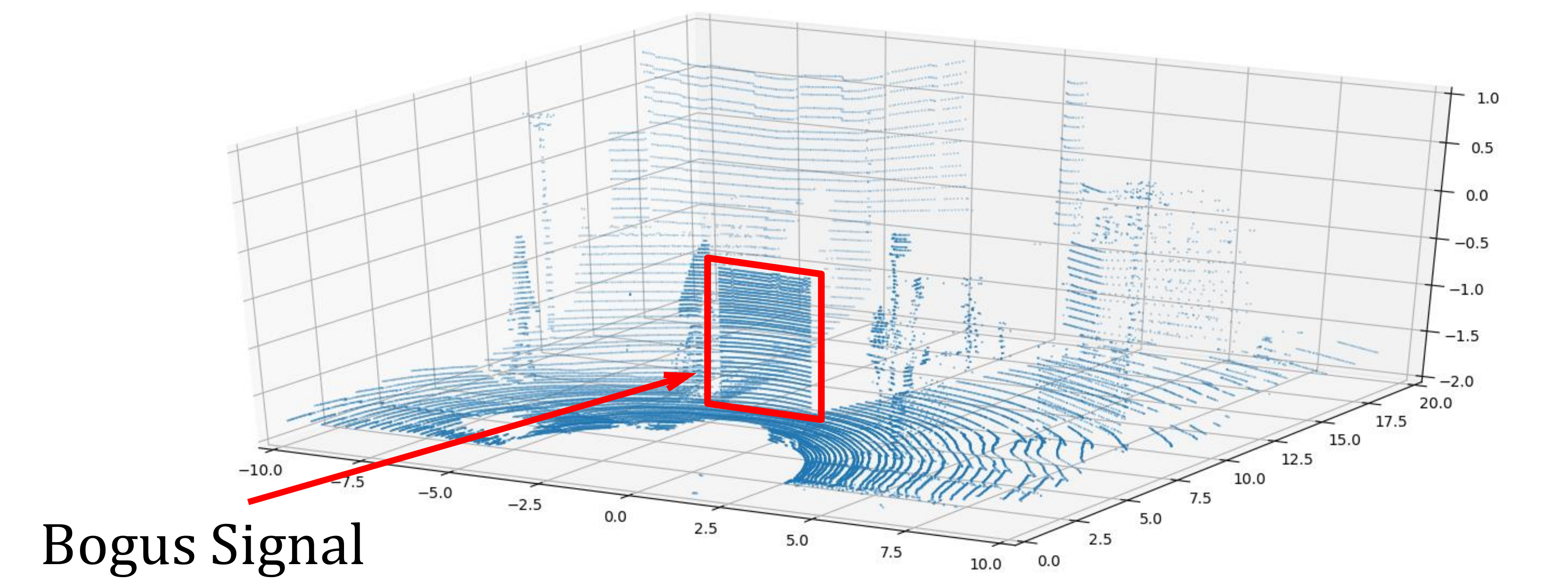}
    \caption{An example of a compromised point cloud that contains bogus signals in a region.}
    \label{fig:lidar_attack}
    \vspace{-0.1in}
\end{figure}

\begin{figure}[!t]
    \centering
    \includegraphics[width=0.48\textwidth]{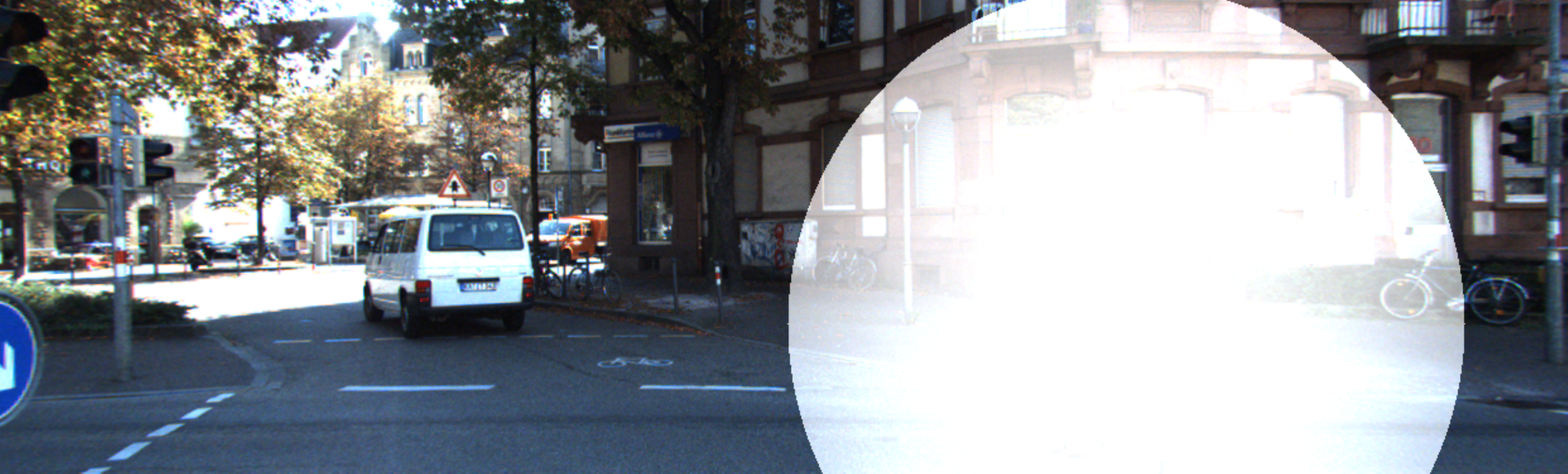}
    \caption{An example of a compromised camera image that contains a round Gaussian facula.}
    \label{fig:camera_attack}
    \vspace{-0.15in}
\end{figure}

\vspace{-0.15in}
\subsection{Attacks on Optical Sensors}

\subsubsection{Attacks against LiDAR}

In~\cite{petit15remote} and~\cite{shin17illusion}, the authors demonstrated several methods to attack LiDARs. The main idea in these attacks is to generate or relay the legitimate optical pulses so as to mislead the perception module in the system model.

Although the existing LiDAR spoofing attacks can only generate a limited number of fake points, we believe that a more powerful attacker can generate a larger number of spoofed points in the point cloud with several advanced attacking sources. Therefore, in this paper, we produce the compromised point clouds by generating spoofed signals for a region, so that the perception module of AVs may detect a fake object, as shown in Fig.~\ref{fig:lidar_attack}.

\subsubsection{Attacks against Camera}

The attacks against camera have been studied in~\cite{petit15remote} and~\cite{yan16can}. The main idea in these studies is to generate strong light signals so as to blind the cameras. According to~\cite{petit15remote}, to blind a camera, the power of light source must increase exponentially with the growth of the distance between the light source and the camera. The effectiveness of the attacks is also affected by the environment light conditions. Therefore, when LED is used, in order to form effective attacks, the distance between the light source and the camera must be within a few meters, and the attacks must be conducted in dark environments, which is less practical. By comparison, attacks using lasers seem to be more realistic.

In our study, we believe that the attackers do not need to completely blind the camera. Instead, their main objective is to mislead the perception module in the autonomous driving system. To this end, we consider that the attacking light source is a laser and the distance between the attacking source and the cameras can be sufficiently large. As a result, the attacks from a laser result in a contaminated area with certain size at a random position in images. Therefore, we generate the affected camera data by overlaying a Gaussian facula on them, as shown in Fig.~\ref{fig:camera_attack}. The affected images we generate are equivalent to the results in~\cite{petit15remote} and~\cite{yan16can}.

\vspace{-0.15in}
\subsection{Impact of the Optical Attacks}

\begin{figure}[!t]
    \centering
    \includegraphics[width=0.48\textwidth]{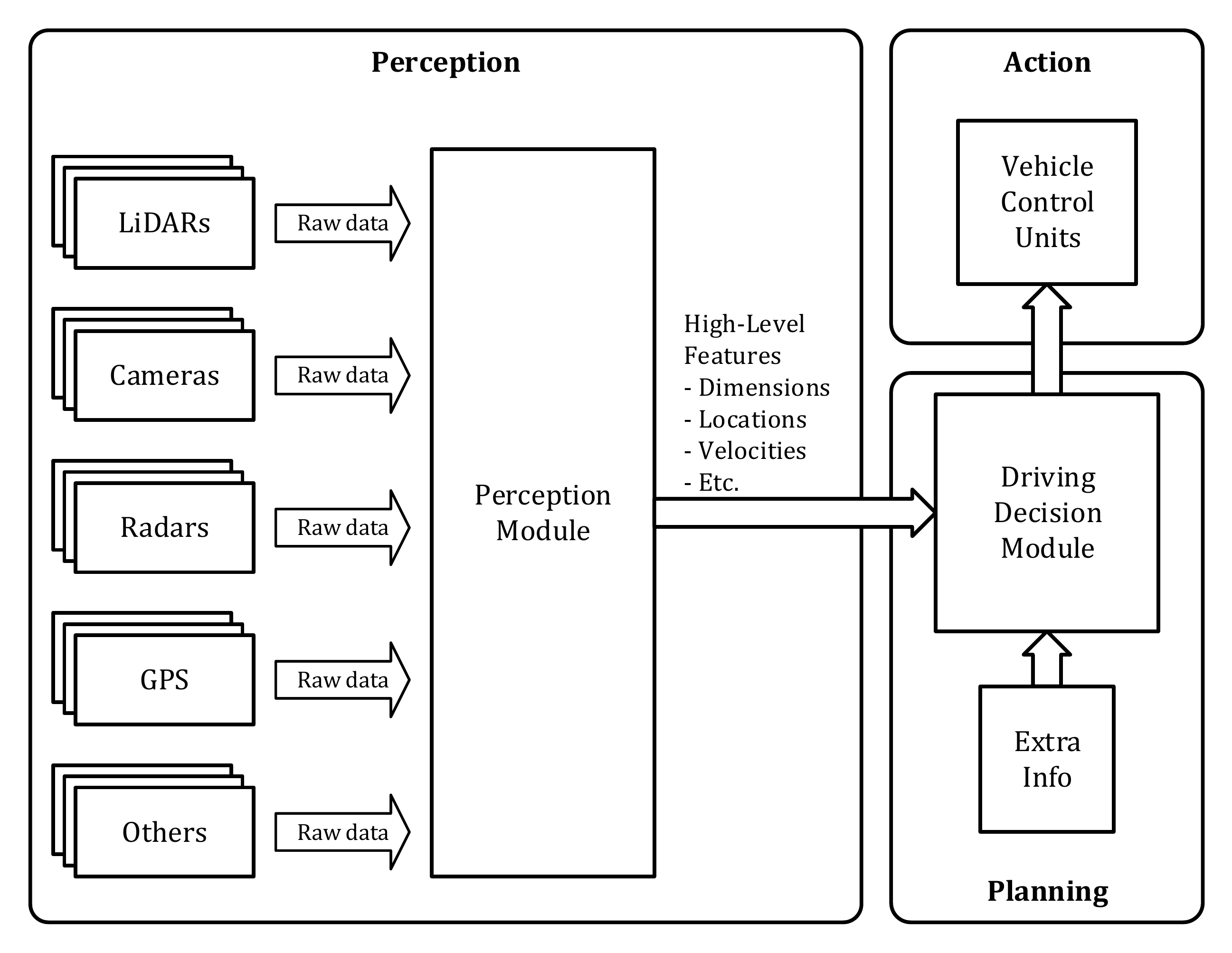}
    \caption{In AV systems, the perception module processes the raw sensor data and generates the environmental high-level information for driving decision making. Then, the driving commands are sent to control units.}
    \label{fig:system_model}
    \vspace{-0.15in}
\end{figure}

To understand the impacts of the aforementioned optical attacks, we conduct extensive experiments testing the object detection algorithms for AVs with the compromised sensor data. Next, we briefly introduce the common experimental setup in this paper, which is also used in the experiments of other sections.

\subsubsection{Common Experimental Setup}
\label{subsubsec:setup}

In this paper, we use two datasets. The first one is the KITTI raw dataset~\cite{geiger13vision}, which includes data of one LiDAR and four cameras in different environmental conditions for autonomous driving, such as \emph{City}, \emph{Residential}, \emph{Road}, \emph{Campus}, etc. We customize it by selecting 1000 sets of sensor data.

The second dataset we use is the one provided in the KITTI object detection benchmark~\cite{geiger12are}, which contains sensor data of one LiDAR and two cameras. We divide the labeled part of the dataset into a training set and a test set according to~\cite{chen153d}. The two sets include 3712 sets of sensor data and 3769 sets of sensor data, respectively.

To produce the compromised LiDAR data, we generate a bogus signal with a height of $1.5$ meters and a width of $2.5$ meters, which is equal to the width of typical highway lanes, at a random distance of $6$ to $10$ meters away from the LiDAR sensor in point clouds.

To generate the compromised camera images, we overlay a round Gaussian facula with a random radius of $187$ pixels to $375$ pixels on images that have a size of $1242$ pixels by $375$ pixels.

\subsubsection{Attack Experiments and Results}

Here, we first conduct experiments on the customized KITTI raw dataset~\cite{geiger13vision}. To evaluate the impact of the optical attacks on LiDAR, we use a pre-trained model based on PIXOR~\cite{yang18pixor}, which is a 3D object detection method using LiDAR data. In our experiments, we generate a compromised point cloud for each one of the 1000 sets of sensor data and feed it to the PIXOR model. We observe that the model falsely considers the bogus signals as obstacles in 986 cases out of 1000.

To measure the impact of the attacks on camera, we use the standard performance metric, \emph{average precision} (AP), where the prediction is considered accurate if and only if the \emph{Intersection over Union} (IoU) is larger than $50\%$. In addition, we use a pre-trained model provided in TensorFlow API~\cite{huang17speed} that can detect vehicles from images. Specifically, the model is based on Faster R-CNN~\cite{ren15faster} with a ResNet-101 architecture~\cite{he16deep}. In our experiments, we produce a compromised image for each set of sensor data in the dataset and feed it to the Faster R-CNN model. The numerical results show that the AP for detecting cars is $84.62\%$ when there are no attacks. By comparison, the AP drops dramatically to $61.53\%$ when there are optical attacks against the camera.

To briefly summarize, we observe that the aforementioned attacks on optical devices can significantly compromise the accuracy of object detection, which is a fundamental task of perception module in autonomous driving. As shown in Fig.~\ref{fig:system_model}, the results of environment perception are passed to the driving decision module that directly sends commands to the vehicle control units, such as the engine and brake. Therefore, we believe that optical attacks are extremely hazardous because it is highly possible that an inaccurate perception due to optical attacks can lead to wrong driving decisions and can cause catastrophic outcomes.

\vspace{-0.1in}
\subsection{A Mitigation Framework Against Optical Attacks}

To defend against such attacks, in this paper, we propose a framework to mitigate optical attacks. The main idea of our framework is to detect optical attacks and then identify the affected sensors. In this manner, the autonomous driving system can choose to use signals from sensors that are not under attack to perform accurate perception.

Specifically, our framework consists of two main procedures. The first procedure is for attack detection. To this end, we consider a system that consists of three sensors in two scenarios, (1) one LiDAR and two cameras, and (2) three cameras. In both cases, we use data from three sensors to obtain two versions of disparity maps and then detect attacks by analyzing the distribution of disparity errors. Based on the first procedure, we design the second procedure to identify up to $n-2$ affected sensors in a system that consists of one LiDAR and $n$ cameras. In Section~\ref{sec.detection} and Section~\ref{sec.identification}, we introduce the two procedures in more detail.

\begin{figure*}[!t]
    \centering
    \includegraphics[width=0.98\textwidth]{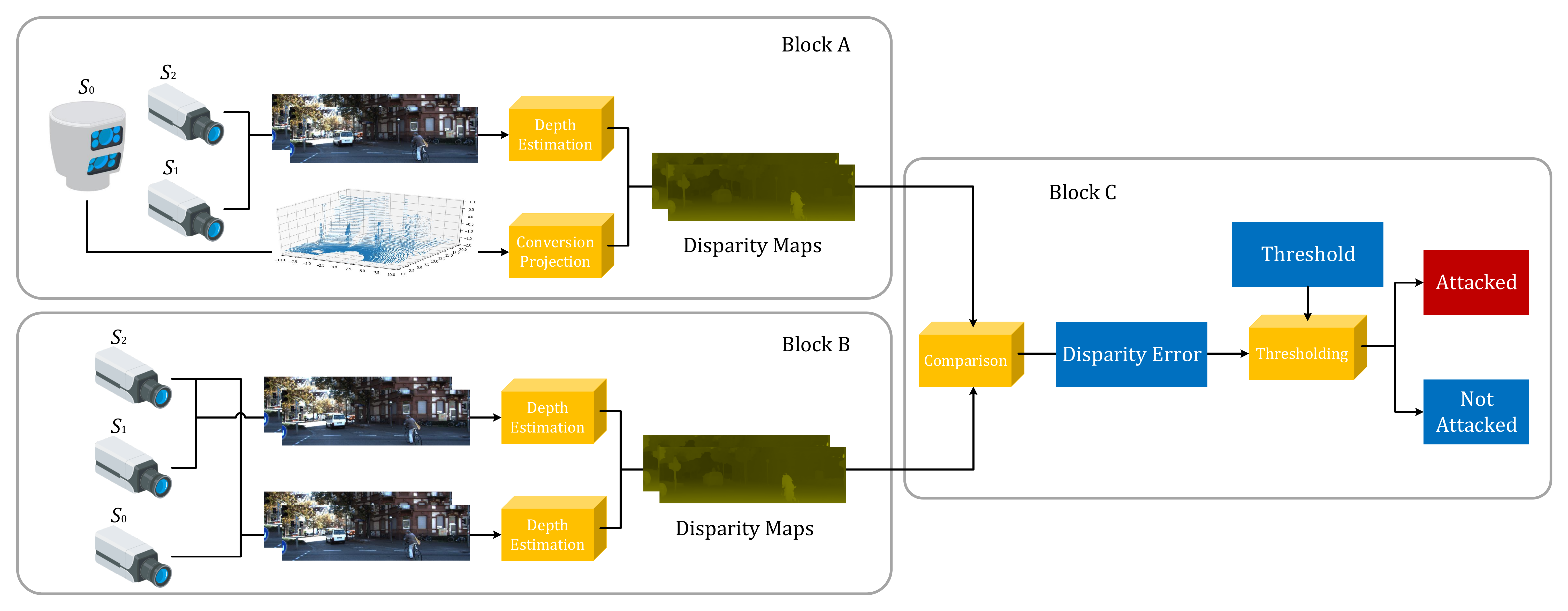}
    \caption{The detection method is designed for two three-sensor systems. For Scenario 1, the detection method structure involves Block A and Block C. For Scenario 2, the detection method structure involves Block B and Block C.}
    \label{fig:detection}
    \vspace{-0.15in}
\end{figure*}

\vspace{-0.15in}
\section{Attack Detection} \label{sec.detection}

In this section, we first explain why we target a system that consists of three sensors and then make a hypothesis about the feasibility of the detection task in a three-sensor system. Next, we focus on the disparity error and how to detect attacks by analyzing the disparity error distribution. Specifically, we elaborate on the calculation of disparity error for two main scenarios of the three-sensor system and conduct extensive experiments on the real dataset to prove the hypothesis and show the accuracy of our method.

\vspace{-0.15in}
\subsection{Three-Sensor Systems and a Hypothesis}

For attack detection, we aim to detect attacks with sufficient accuracy using the \emph{smallest} number of sensors. Due to the trade-off between cost and performance of object detection, usually, there is one primary LiDAR mounted on the roof of an autonomous vehicle which is also equipped with multiple cameras~\cite{apollo19perception}. To obtain two versions of a depth map from an AV system like this, we at least need three sensors: one LiDAR and two cameras, or three cameras.

For the first case, we notice that LiDAR can produce accurate depth maps in point clouds. On the other hand, stereo-vision based depth estimation algorithms can also generate depth maps out of stereo images. Intuitively, if we compare a depth map produced by LiDAR and another generated by two stereo images, we may be able to detect the distortion of depth information caused by optical attacks on such a three-sensor system. Consequently, the first three-sensor system that we consider consists of one LiDAR and two cameras.

For the second case, it is obvious that we can use the image taken by one camera as the reference, and then use images taken by two other cameras to produce two depth maps using a depth estimation model. By comparing the two depth maps, we may be able to detect attacks on the second three-sensor system that consists of three cameras.

To briefly summarize, in this paper, we consider two three-sensor systems that are practical in autonomous driving systems. Furthermore, we make a \emph{hypothesis} that, with appropriate design, we can accurately detect the optical attacks on both of the three-sensor systems. In the following discussions, we verify this hypothesis by elaborating on the mechanisms to detect attacks on the two three-sensor systems, respectively.

\vspace{-0.1in}
\subsection{Scenario 1: One LiDAR and Two Cameras}

For this scenario, we denote the LiDAR as sensor $S_{0}$, and denote two cameras from the right to the left as $S_{1}$ and $S_{2}$, respectively. Accordingly, the data generated by the sensors are denoted as $D_{0}$, $D_{1}$, and $D_{2}$. The detection system we design for this scenario is shown as the combination of Block A and Block C in Fig.~\ref{fig:detection}.

In our system illustrated in Fig.~\ref{fig:detection} (Block A \& Block C), we designate camera $S_2$ as the reference camera to generate two disparity maps. Specifically, we set the image taken by the reference camera (i.e., $D_{2}$) as the reference image, and then feed it with the image taken by the other camera (i.e., $D_{1}$) to a depth estimation model to produce the first disparity map, denoted as $DM_{1,2}$, in which we include the disparity information at each pixel of the reference image.

Here, we note that many algorithms have been developed in recent years that can generate accurate disparity maps with camera images. For instance, PSMNet~\cite{chang18pyramid} and DORN~\cite{fu18deep} are two recent algorithms based on deep learning. In this paper, we use the former one to produce disparity maps, since the PSMNet model gives more accurate results over others.

In addition to $DM_{1,2}$, we also project the depth information (i.e., point cloud $D_{0}$) obtained by LiDAR onto the reference image $D_{2}$ to generate the second disparity map, denoted as $DM_{0,2}$. In this procedure, the depth information in the point cloud $D_{0}$ is converted to disparities by using Eqn.~(\ref{eq:triangulation}). To generate all disparity maps in the same scale, $f$ in the equation is set to be the same as the focal length of the cameras, and $b$ is set to be equal to the baseline of the stereo vision formed by $S_{1}$ and $S_{2}$. Then, disparities are projected onto $D_{2}$.

\begin{figure*}[t]
	\centering
	\subfloat[]{
		\label{fig:s0_1_d}
		\includegraphics[width=0.23\textwidth]{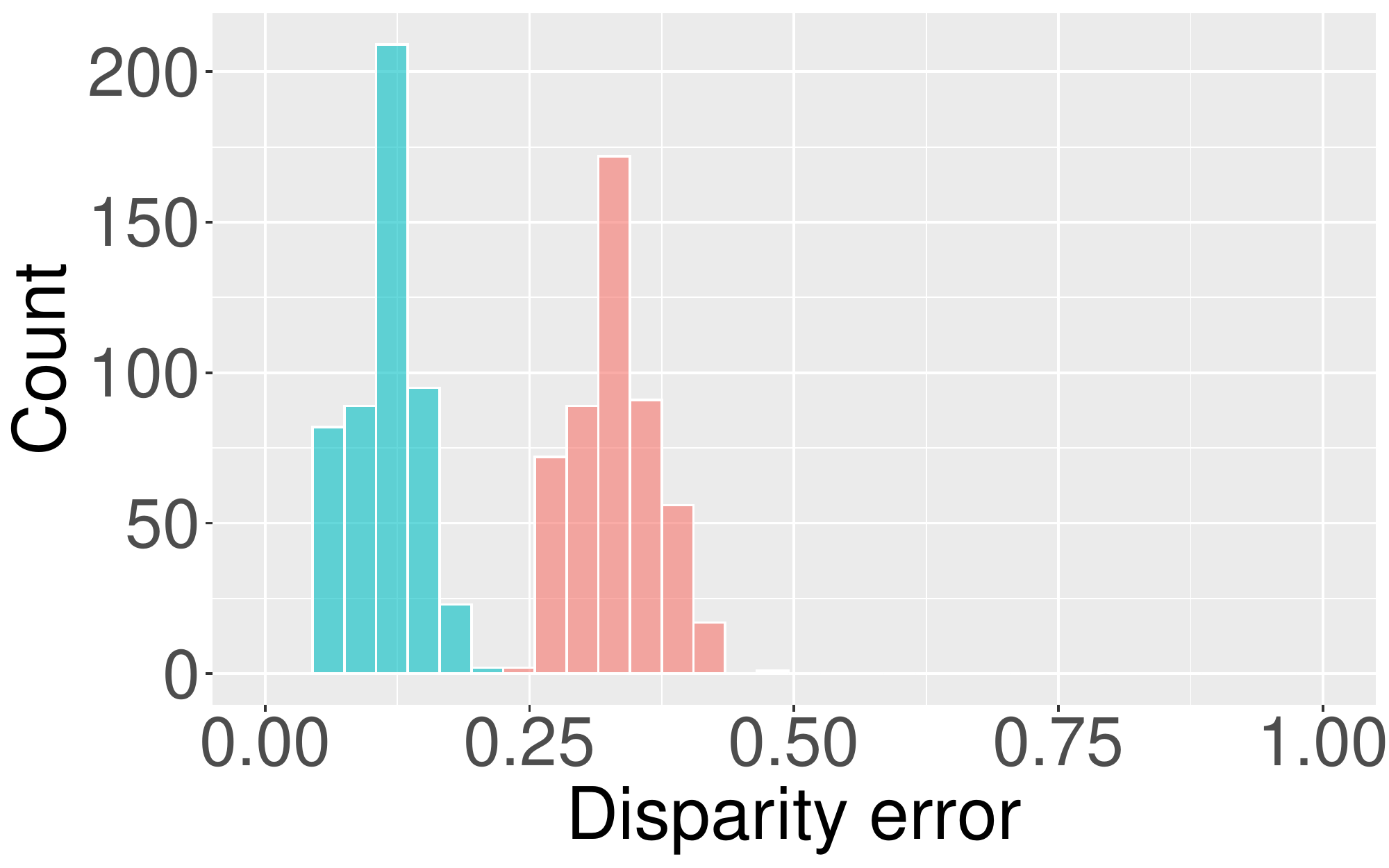}}
	\hfil
	\subfloat[]{
		\label{fig:s1_1_d}
		\includegraphics[width=0.23\textwidth]{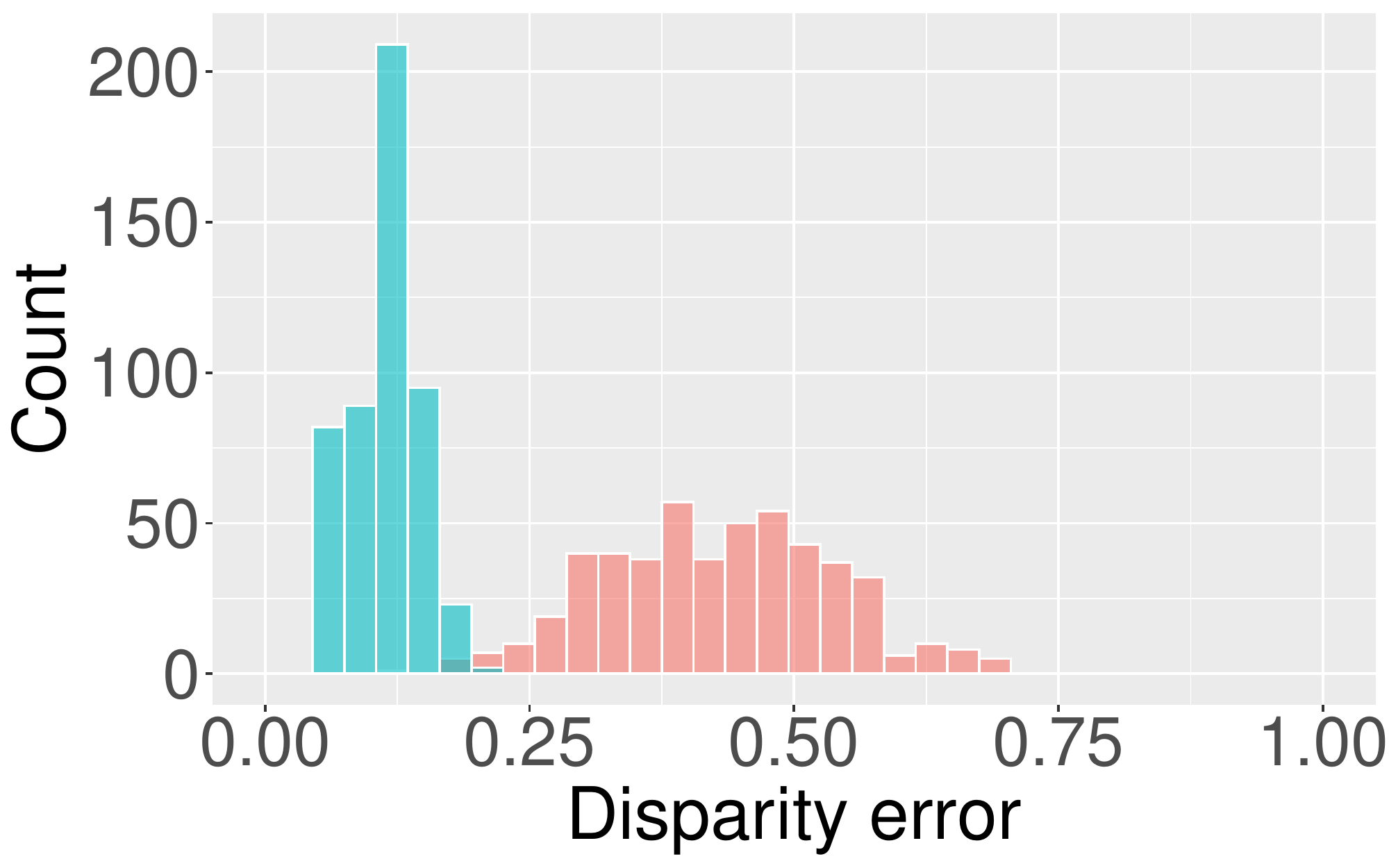}}
	\hfil
	\subfloat[]{
		\label{fig:s2_1_d}
		\includegraphics[width=0.23\textwidth]{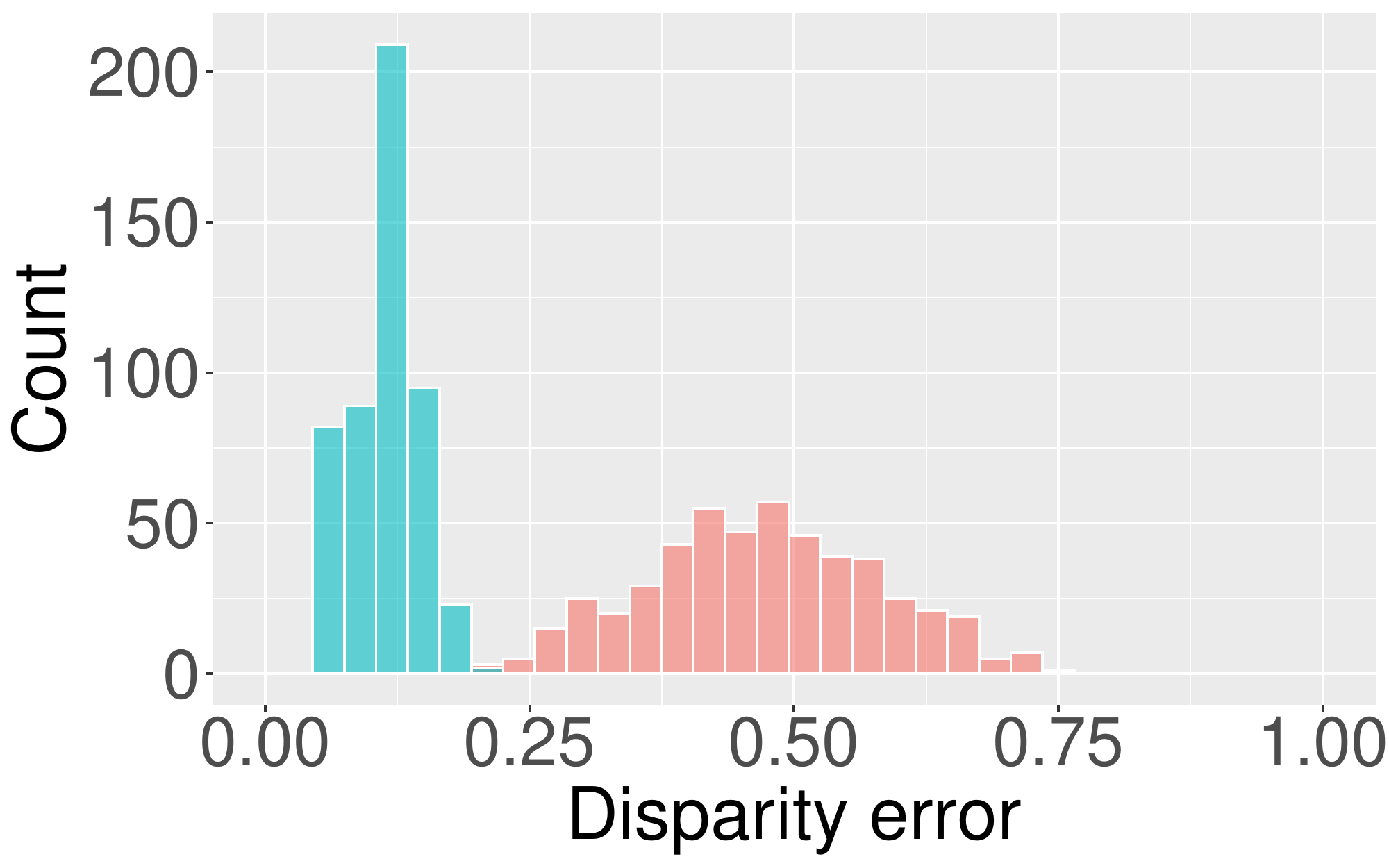}}
	\hfil
	\subfloat[]{
		\label{fig:s0_s1_1_d}
		\includegraphics[width=0.23\textwidth]{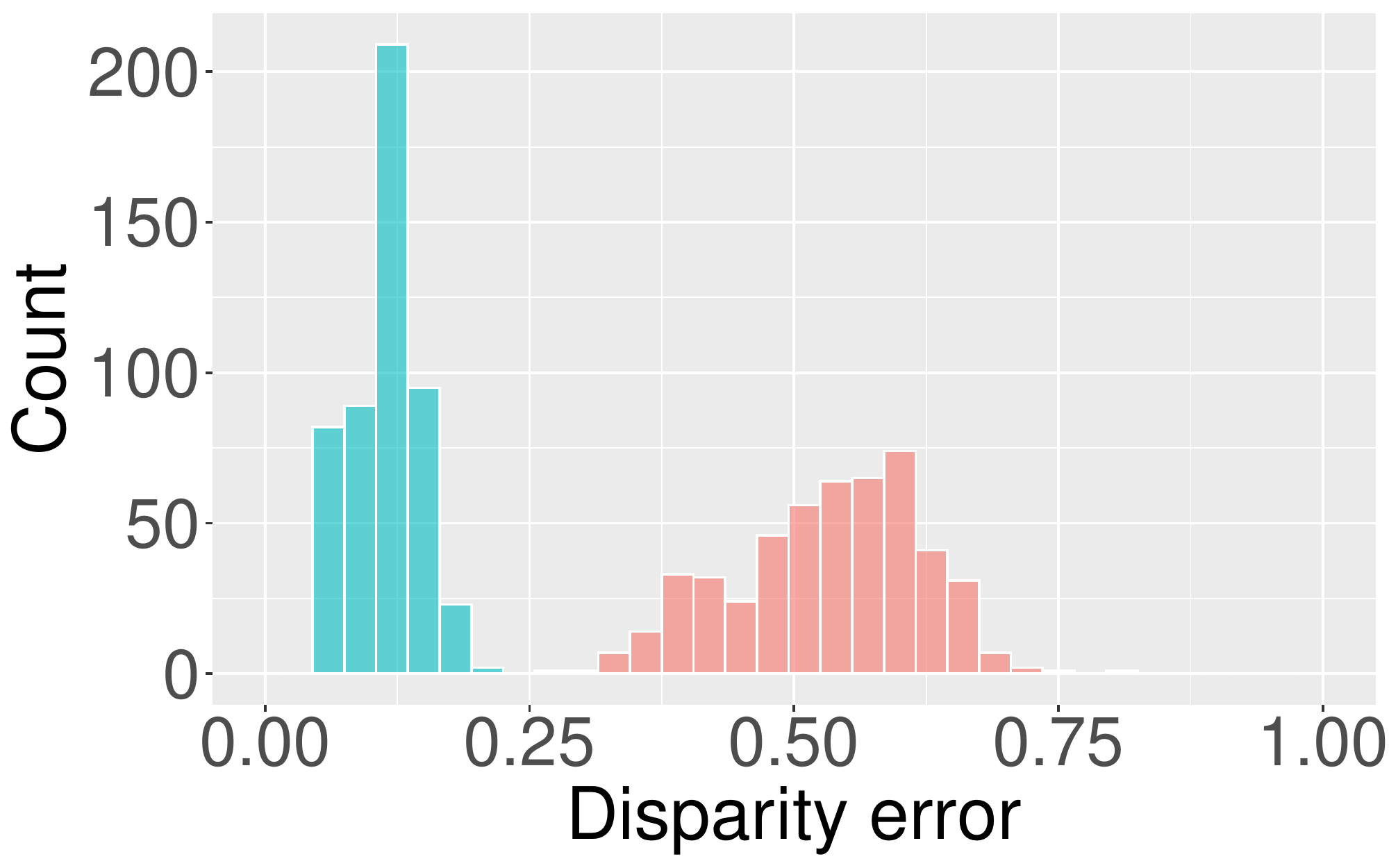}}
	\hfil
	\vspace{-0.1in}
	\subfloat[]{
		\label{fig:s0_s2_1_d}
		\includegraphics[width=0.23\textwidth]{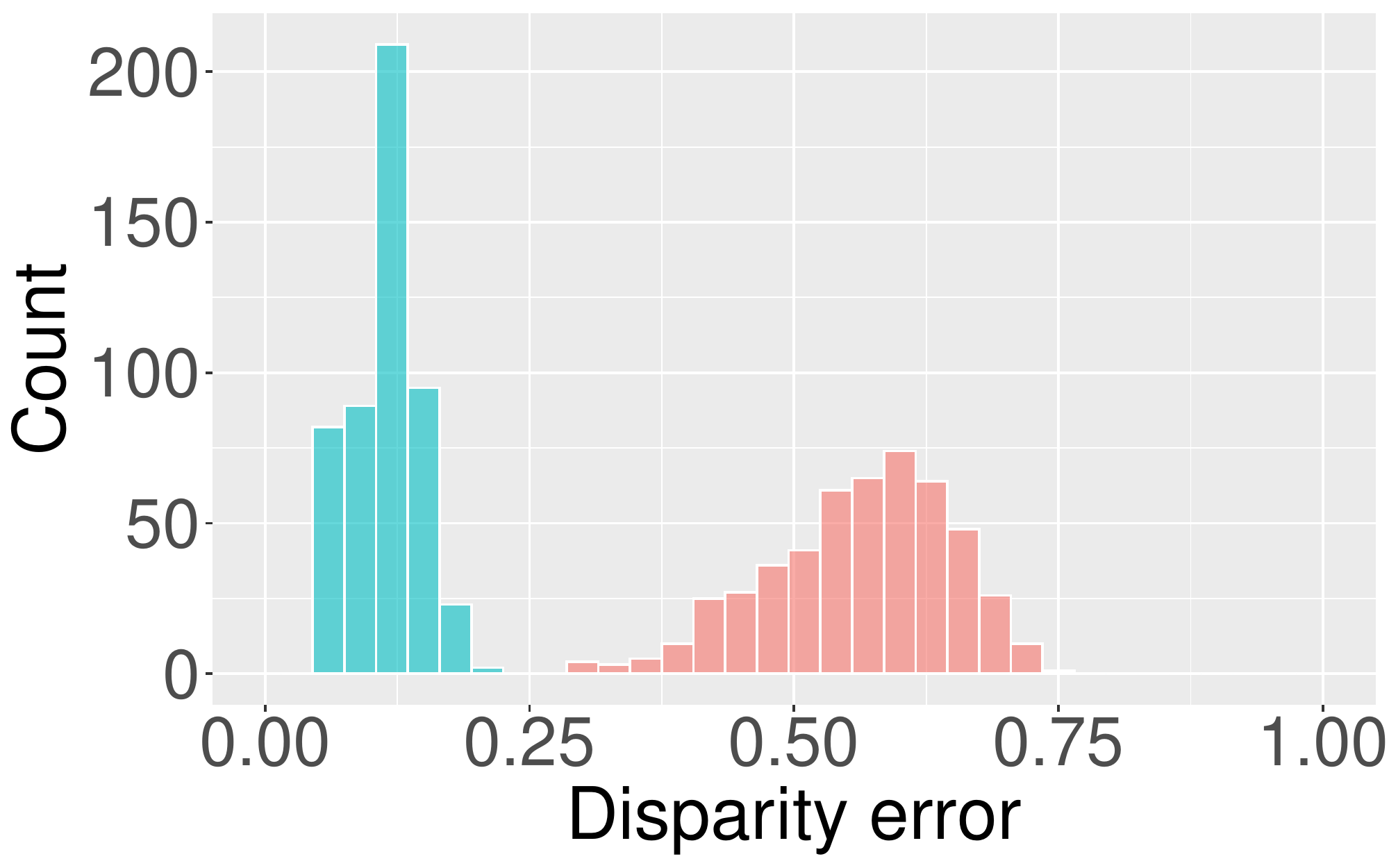}}
	\hfil
	\subfloat[]{
		\label{fig:s1_s2_1_d}
		\includegraphics[width=0.23\textwidth]{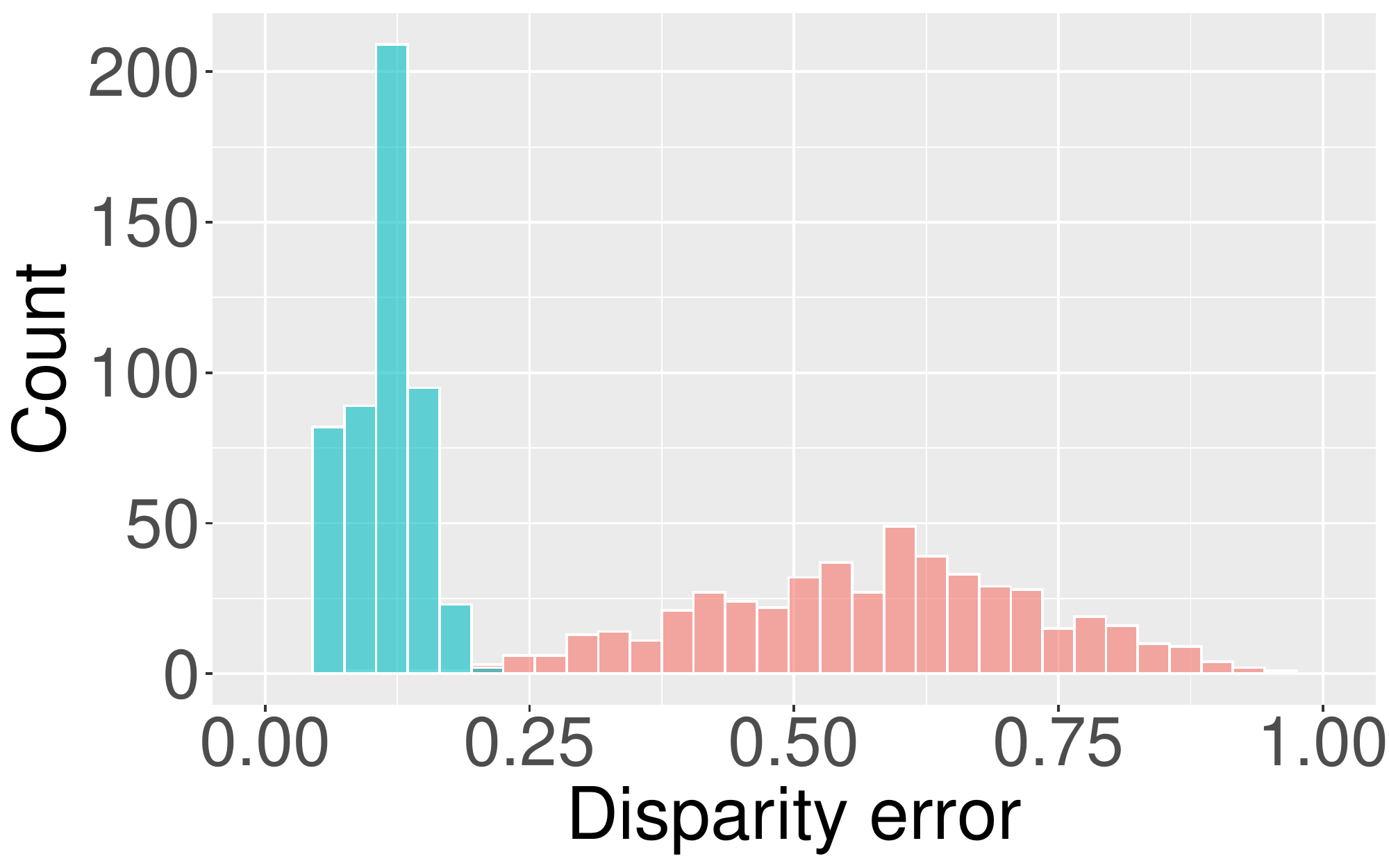}}
	\hfil
	\subfloat[]{
		\label{fig:s0_s1_s2_1_d}
		\includegraphics[width=0.23\textwidth]{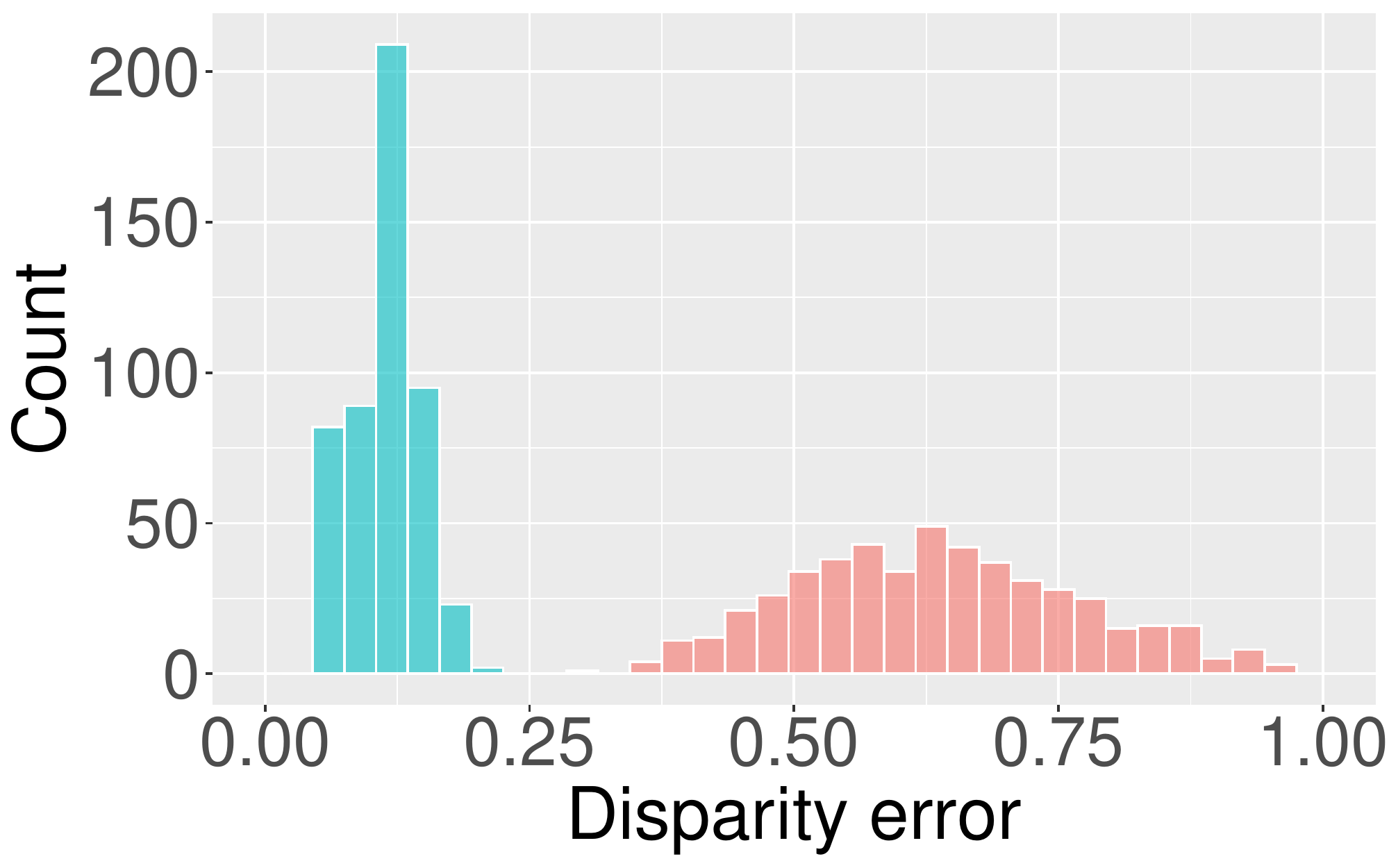}}
	\caption{Distributions of disparity error $E_{0,1,2}$ in normal case (cyan bars) and in attack cases (red bars) for Scenario 1. (a)~No attack vs. $S_{0}$ attacked; (b)~No attack vs. $S_{1}$ attacked; (c)~No attack vs. $S_{2}$ attacked; (d)~No attack vs. $S_{0},S_{1}$ attacked; (e)~No attack vs. $S_{0},S_{2}$ attacked; (f)~No attack vs. $S_{1},S_{2}$ attacked; (g)~No attack vs. $S_{0},S_{1},S_{2}$ attacked.}
	\label{fig:scen_1_exp_d}
	\vspace{-0.15in}
\end{figure*}

\begin{figure*}
	\centering
	\subfloat[]{
		\label{fig:s0_1_r}
		\includegraphics[width=0.23\textwidth]{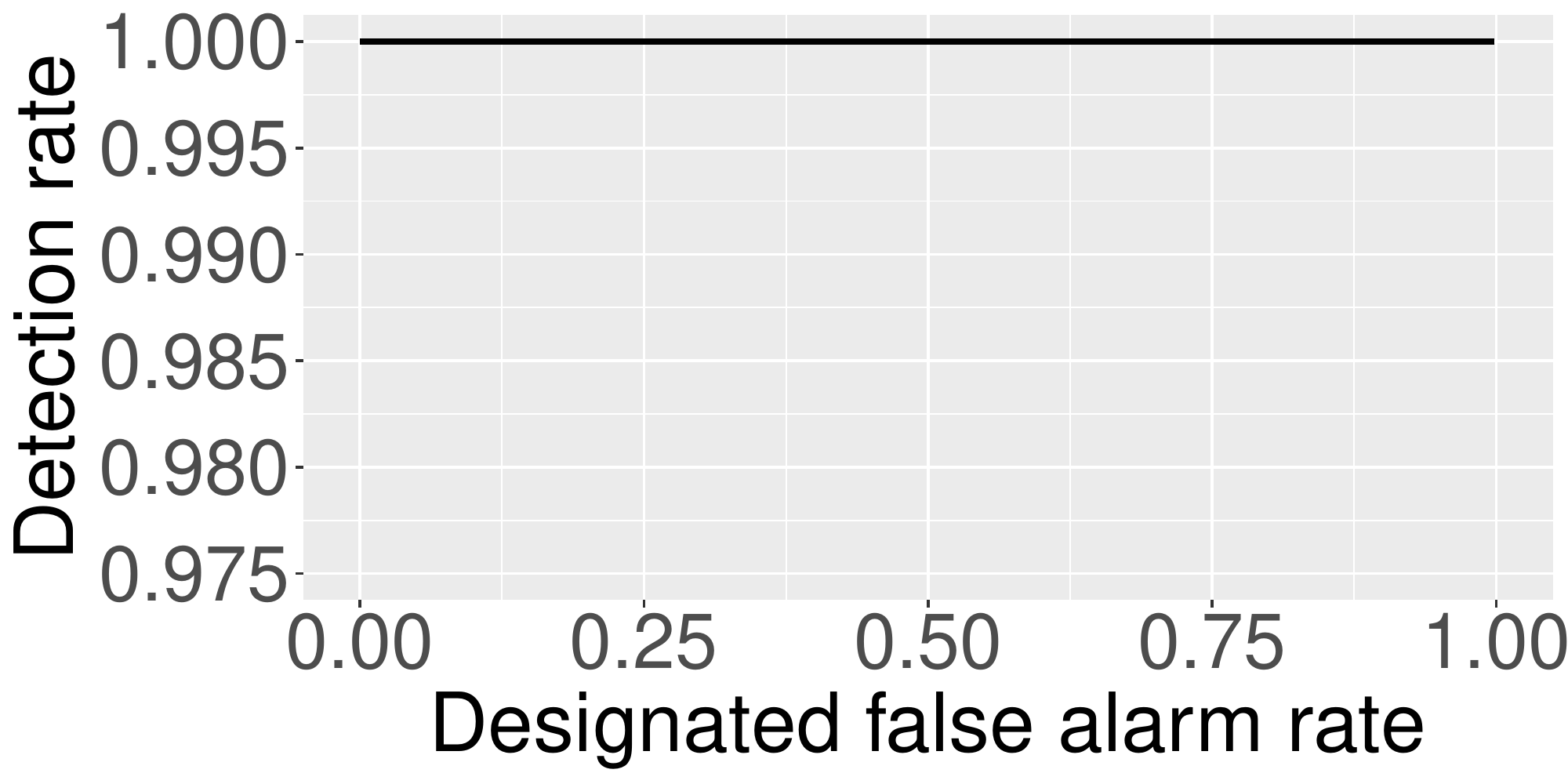}}
	\hfil
	\subfloat[]{
		\label{fig:s1_1_r}
		\includegraphics[width=0.23\textwidth]{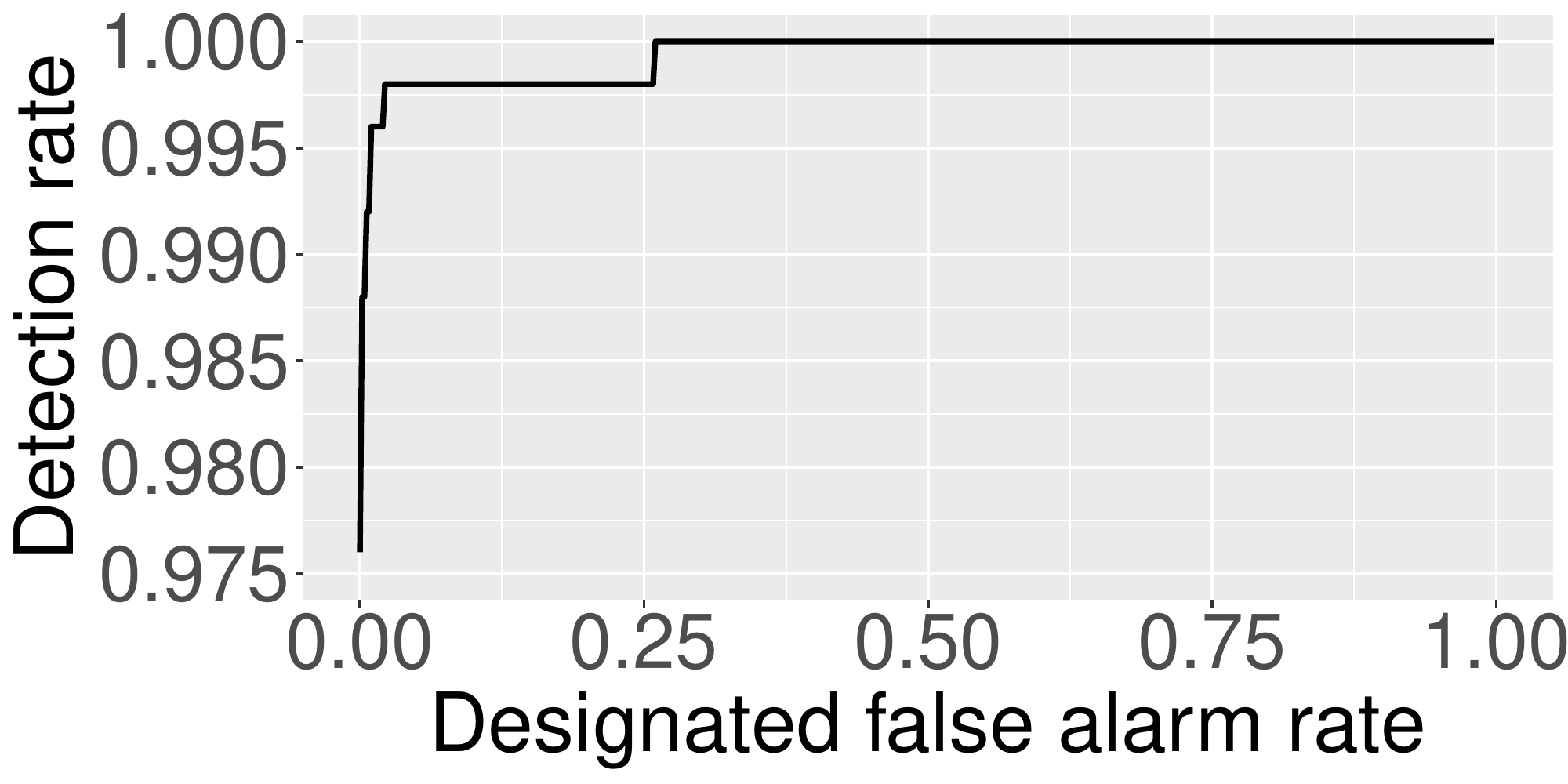}}
	\hfil
	\subfloat[]{
		\label{fig:s2_1_r}
		\includegraphics[width=0.23\textwidth]{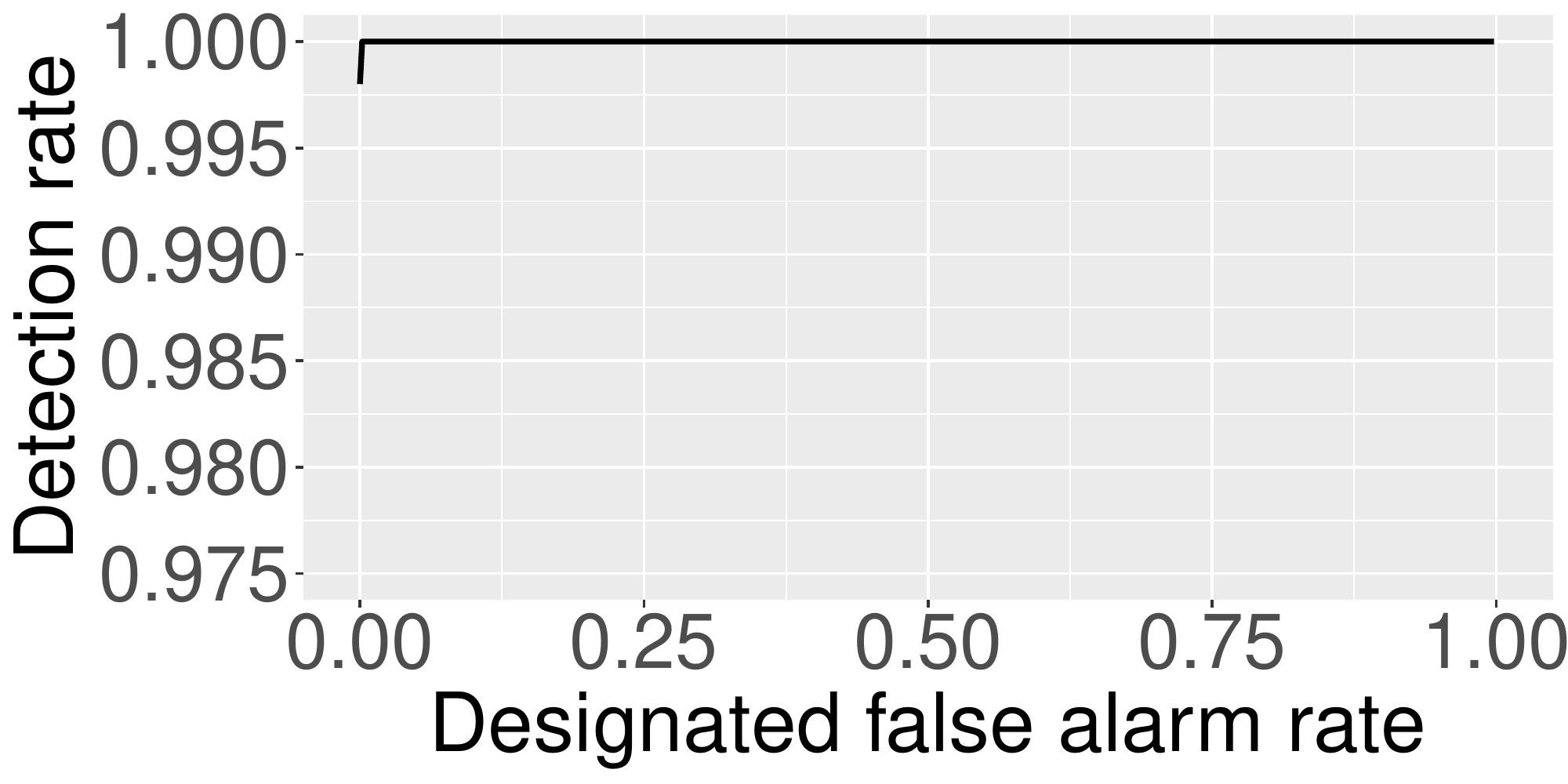}}
	\hfil
	\subfloat[]{
		\label{fig:s0_s1_1_r}
		\includegraphics[width=0.23\textwidth]{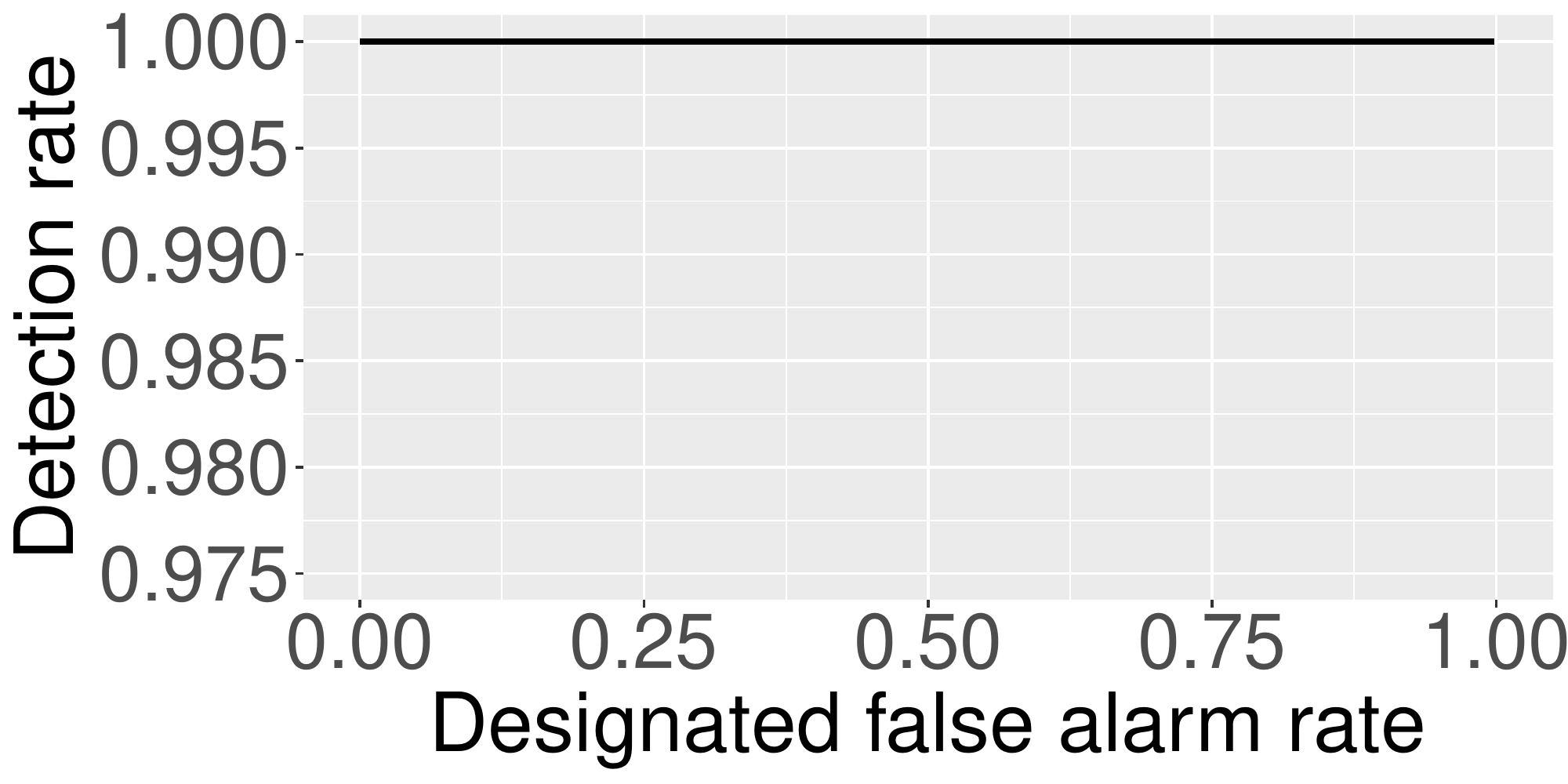}}
	\hfil
	\vspace{-0.1in}
	\subfloat[]{
		\label{fig:s0_s2_1_r}
		\includegraphics[width=0.23\textwidth]{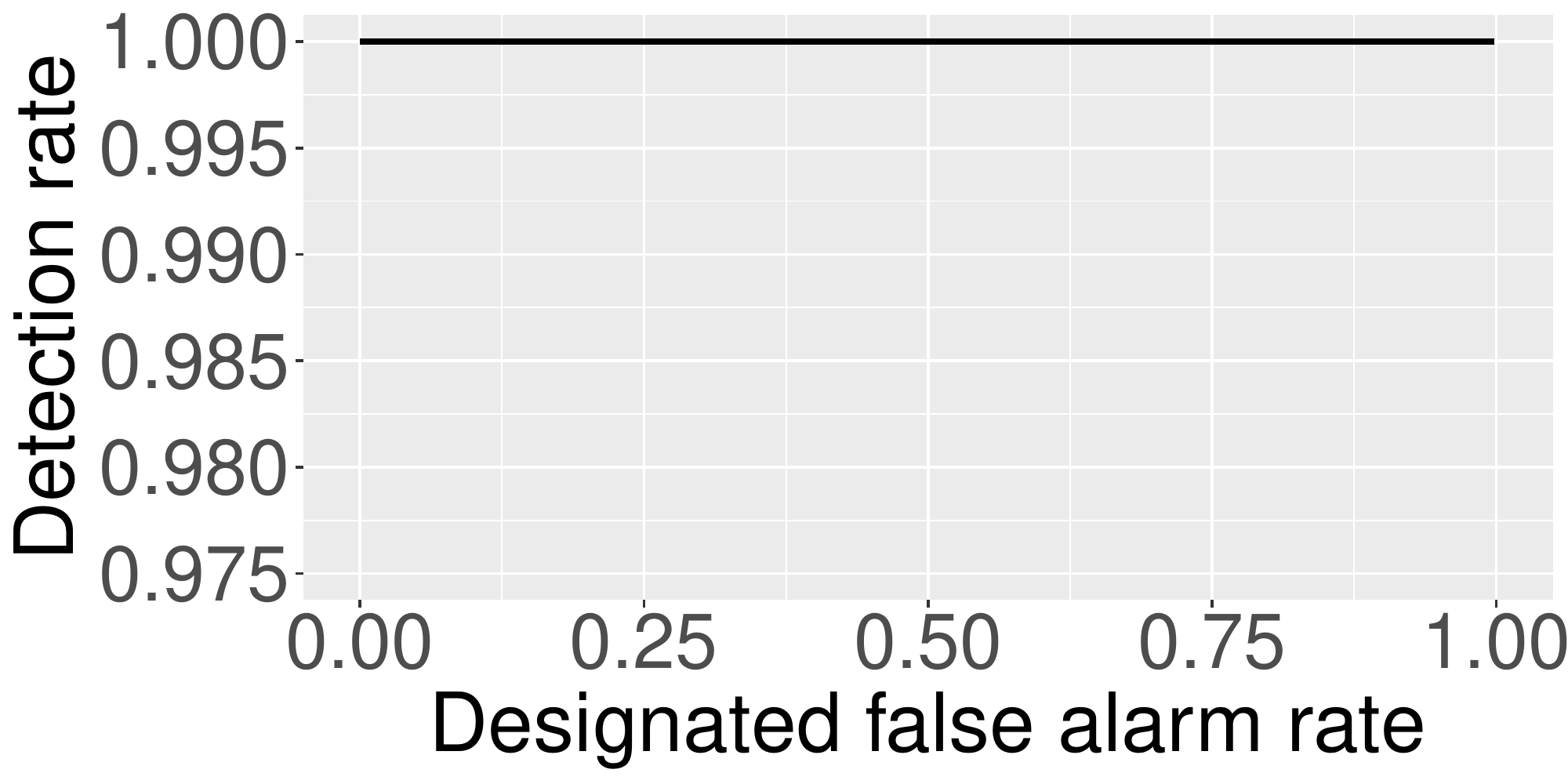}}
	\hfil
	\subfloat[]{
		\label{fig:s1_s2_1_r}
		\includegraphics[width=0.23\textwidth]{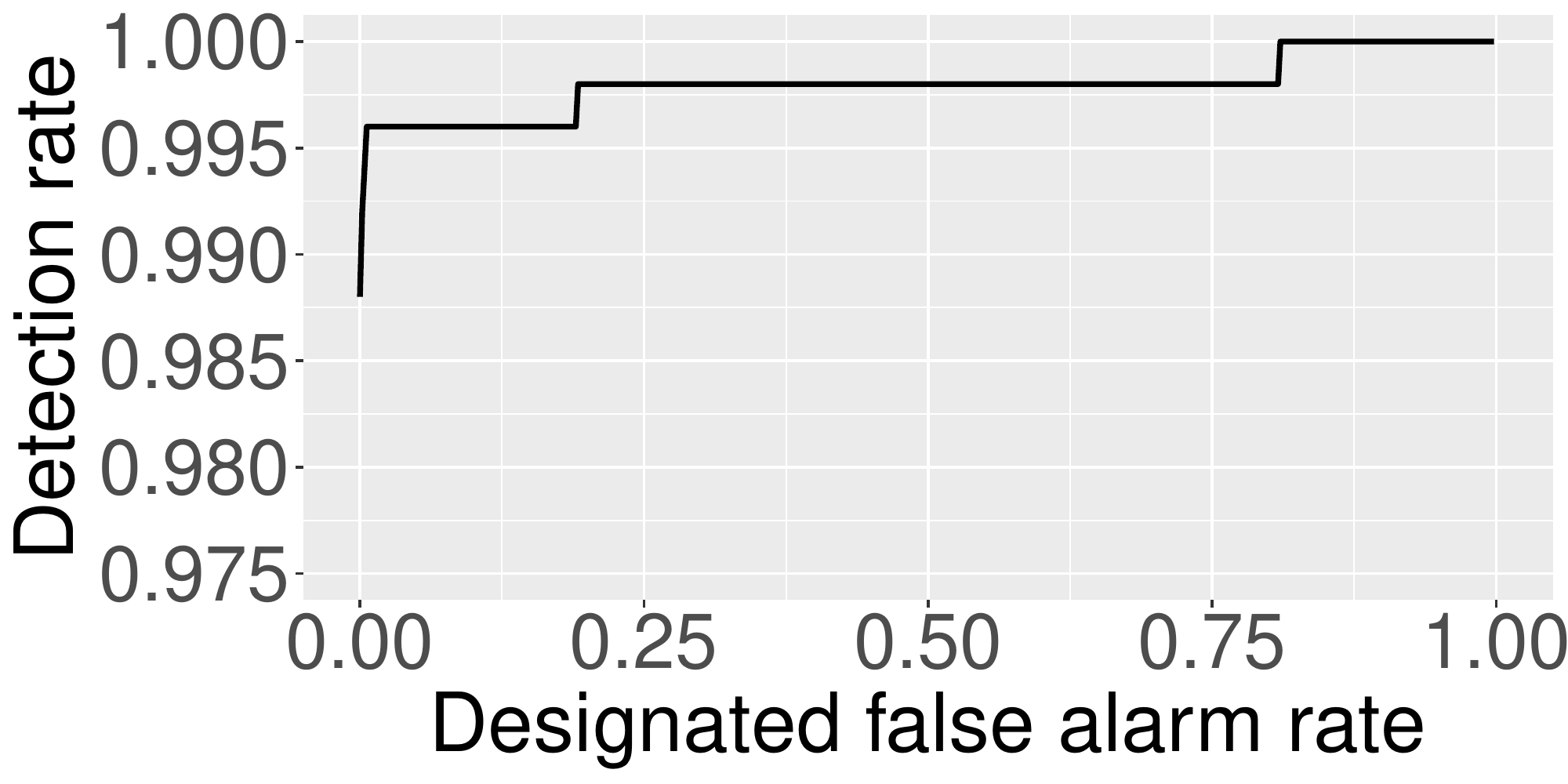}}
	\hfil
	\subfloat[]{
		\label{fig:s0_s1_s2_1_r}
		\includegraphics[width=0.23\textwidth]{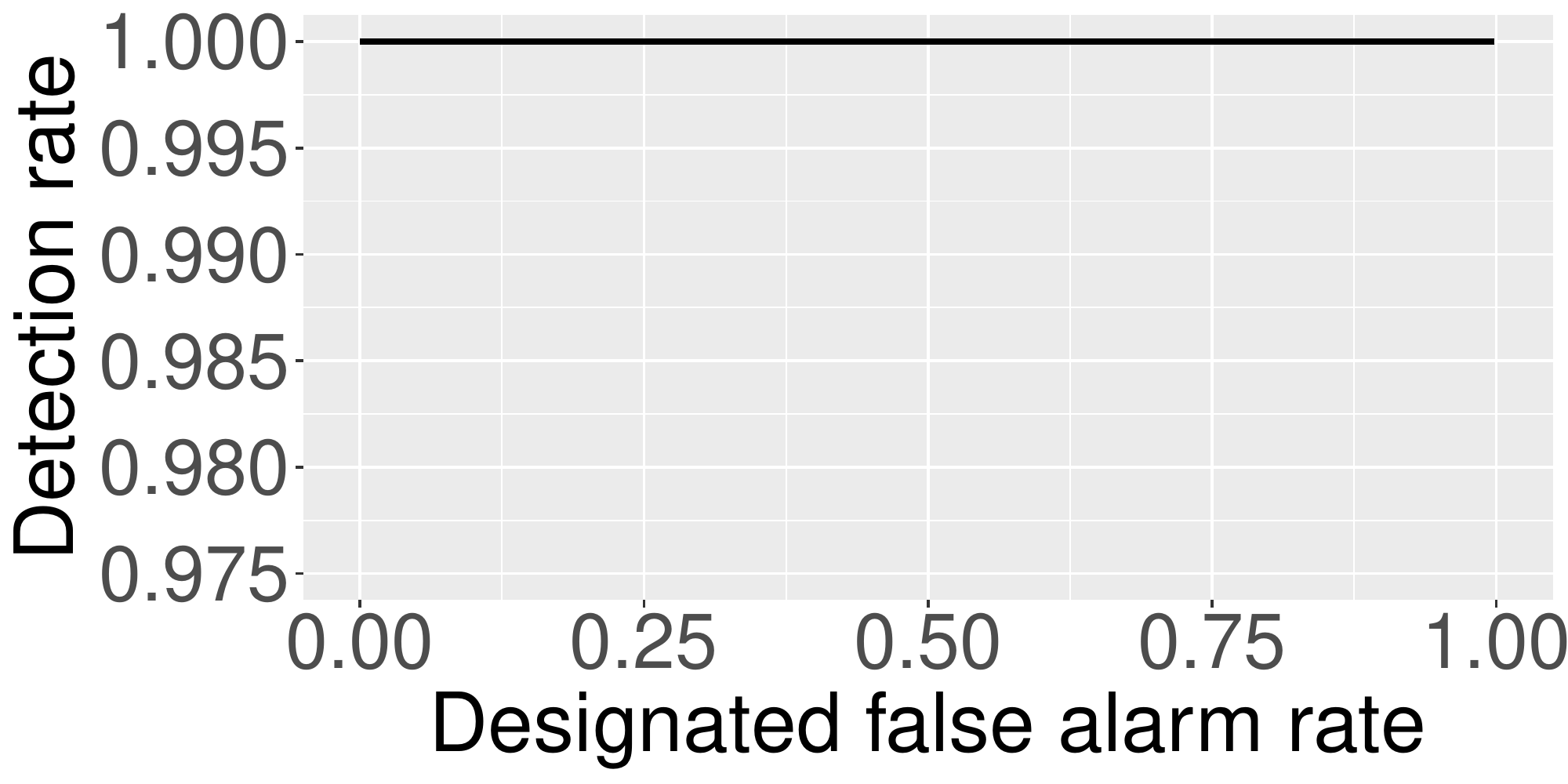}}
	\caption{Detection rate varies with the designated false alarm rate $r$ in each attack case for Scenario 1. (a)~$S_{0}$ is attacked; (b)~$S_{1}$ is attacked; (c)~$S_{2}$ is attacked; (d)~$S_{0}$ and $S_{1}$ are attacked; (e)~$S_{0}$ and $S_{2}$ are attacked; (f)~$S_{1}$ and $S_{2}$ are attacked; (g)~$S_{0}$, $S_{1}$ and $S_{2}$ are attacked.}
	\label{fig:scen_1_exp_r}
	\vspace{-0.15in}
\end{figure*}

To detect optical attacks, we compare the two disparity maps: $DM_{1,2}$ and $DM_{0,2}$. Since in the two procedures described above we use $D_2$ as the reference image, the two produced disparity maps have the same scale and share the same view. Thus, we can compare them directly. Here, it shall be noted that $DM_{0,2}$ contains sparse disparity information, since the distances in the point cloud $D_0$ are not densely measured. Therefore, in this comparison procedure, we only compare pixels that have valid disparity in $DM_{0,2}$. For valid pixels, we take the KITTI stereo benchmark~\cite{menze15object} as a reference and design our own standard, in which a disparity inconsistency for pixel $a$ is counted if and only if
\begin{equation}\label{eq:disparity_error}
\begin{cases}
    |DM_{0,2}(a)-DM_{1,2}(a)|>3,\\
    \frac{|DM_{0,2}(a)-DM_{1,2}(a)|}{min(DM_{0,2}(a),DM_{1,2}(a))}>0.05.
\end{cases}
\end{equation}

Based on the \emph{pixel-level} disparity inconsistencies, we evaluate the disparity error, denoted as $E_{0,1,2}$, between $DM_{0,2}$ and $DM_{1,2}$. In particular, the \emph{disparity error} is defined as the ratio of the quantity of \emph{pixel-level} disparity inconsistencies over the total number of valid pixels.

Finally, to detect an optical attack, we need to evaluate the ranges of disparity errors in normal cases and attack cases. We believe that the two ranges are distinguishable and we can then use a threshold $\theta_{0,1,2}$ to determine whether there is an optical attack. In particular, we determine that one of the three sensors is under attack if and only if $E_{0,1,2}>\theta_{0,1,2}$.

The threshold $\theta_{0,1,2}$ is determined offline based on the value distribution of $E_{0,1,2}$ when the three-sensor system is in a safe environment, since only the correct data of optical sensors are available on an autonomous vehicle in normal conditions. According to the statistical law, we use a large number of samples of the disparity error to represent its real distribution in normal cases and define a designated false alarm rate $r$ to arbitrarily set $r \times 100\%$ of them as virtual outliers, where $r \in [0,1]$. Then, the critical value separating inliers from outliers is $\theta_{0,1,2}$:
\begin{equation}\label{eq:detection_threshold}
    \frac{\text{\# samples of } E_{0,1,2} > \theta_{0,1,2}}{\text{\# samples of } E_{0,1,2}} = r.
\end{equation}
In this manner, the threshold, which only moves within the value range of disparity error samples, is determined by the value of $r$. Intuitively, to obtain the best detection performance, we should maximize the detection rate and minimize the designated false alarm rate. Hence, we show how the detection rate varies when adjusting the threshold via $r$.

\vspace{-0.1in}
\subsection{Experiments for Scenario 1}

\begin{table}[t]
    \centering
	\caption{Detection rate comparison between the method in our framework and the baseline}\label{tbl:dete_exp_r}
	\begin{tabular}{|c|c|c|c|}
	    \hline
	        \multirow{2}{*}{Method} & \multirow{2}{*}{Granularity} & \multicolumn{2}{c|}{Avg. Detection Rate} \\
		\cline{3-4}
		    && Scenario 1 & Scenario 2 \\
		\hline
    	\hline
    	Ours ($r=0\%$) & Pixel-level & $99.46\%$ & $99.94\%$ \\
    	Ours ($r=1\%$) & Pixel-level & $99.89\%$ & $99.97\%$ \\
    	Ours ($r=2\%$) & Pixel-level & $99.89\%$ & $100\%$ \\
    	Ours ($r=3\%$) & Pixel-level & $99.91\%$ & $100\%$ \\
    	Ours ($r=5\%$) & Pixel-level & $99.91\%$ & $100\%$ \\
    	\hline
    	Baseline (IoU $=0.5$) & Object-level & $65.32\%$ & $67.17\%$ \\
    	Baseline (IoU $=0.7$) & Object-level & $59.54\%$ & $63.39\%$\\
    	\hline
	\end{tabular}
	\vspace{-0.15in}
\end{table}

\subsubsection{Setup}

To validate the hypothesis for this scenario, we conduct extensive experiments. We consider all possible attack cases where any sensor or any combination of the three sensors gets attacked. We use the data of one LiDAR and two cameras from the customized KITTI raw dataset~\cite{geiger13vision} to produce affected sensor data for each attack case. The production scheme is described in Section~\ref{subsubsec:setup}. The PSMNet model used in the experiments is provided by Wang \textit{et al.}~\cite{wang19pseudo}, which is trained on Scene Flow dataset~\cite{mayer16large} and KITTI object detection dataset~\cite{geiger12are}. As for metrics, we measure the disparity error distribution and the rate of correct detection for each attack case.

In the literature, there is no existing solution for optical attack detection. Therefore, to compare our scheme with possible solutions, we implement a possible baseline solution to optical attack detection that first extracts \emph{object-level} features from the data of two individual sensors respectively, and then detects attacks by measuring the mismatches between the two sets of features. Specifically for Scenario 1, we implement the baseline with the backbone of PIXOR~\cite{yang18pixor} for extracting \emph{object-level} features from point clouds and the backbone of Faster R-CNN~\cite{ren15faster} for extracting from images. We set IoU to $0.5$ and $0.7$ for determining feature mismatches.

\begin{figure*}[!t]
	\centering
	\subfloat[]{
		\label{fig:s0_2_d}
		\includegraphics[width=0.23\textwidth]{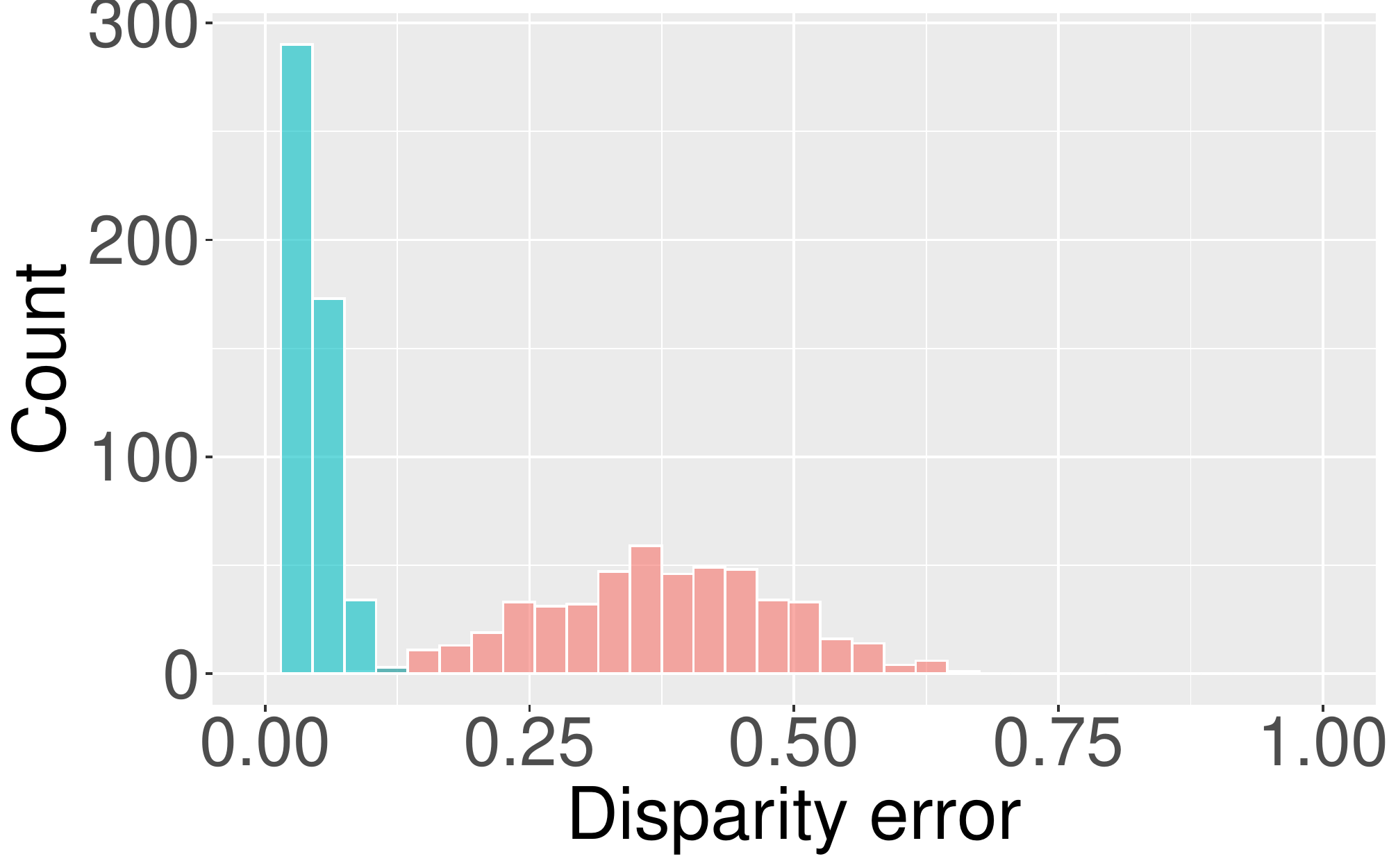}}
	\hfil
	\subfloat[]{
		\label{fig:s1_2_d}
		\includegraphics[width=0.23\textwidth]{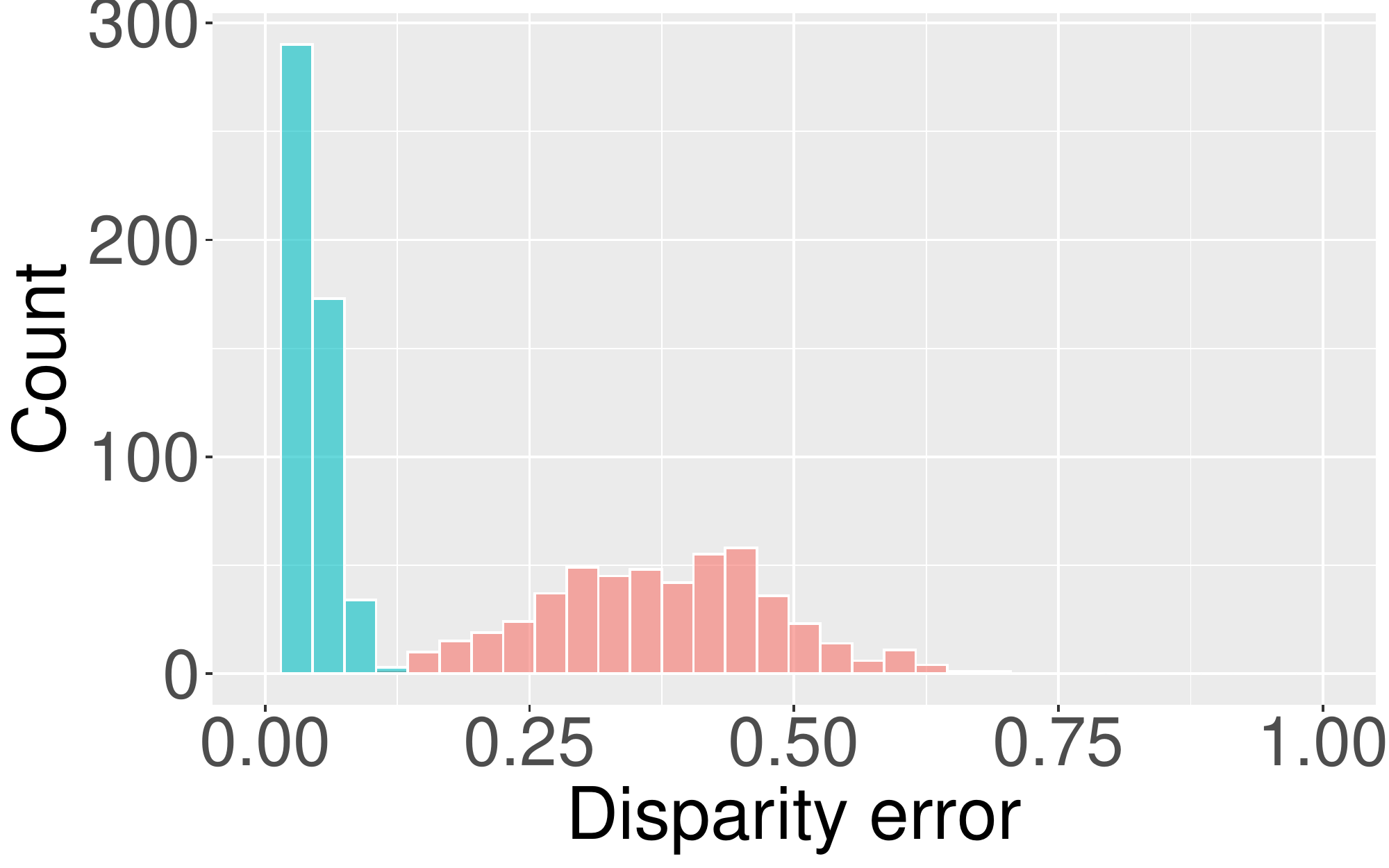}}
	\hfil
	\subfloat[]{
		\label{fig:s2_2_d}
		\includegraphics[width=0.23\textwidth]{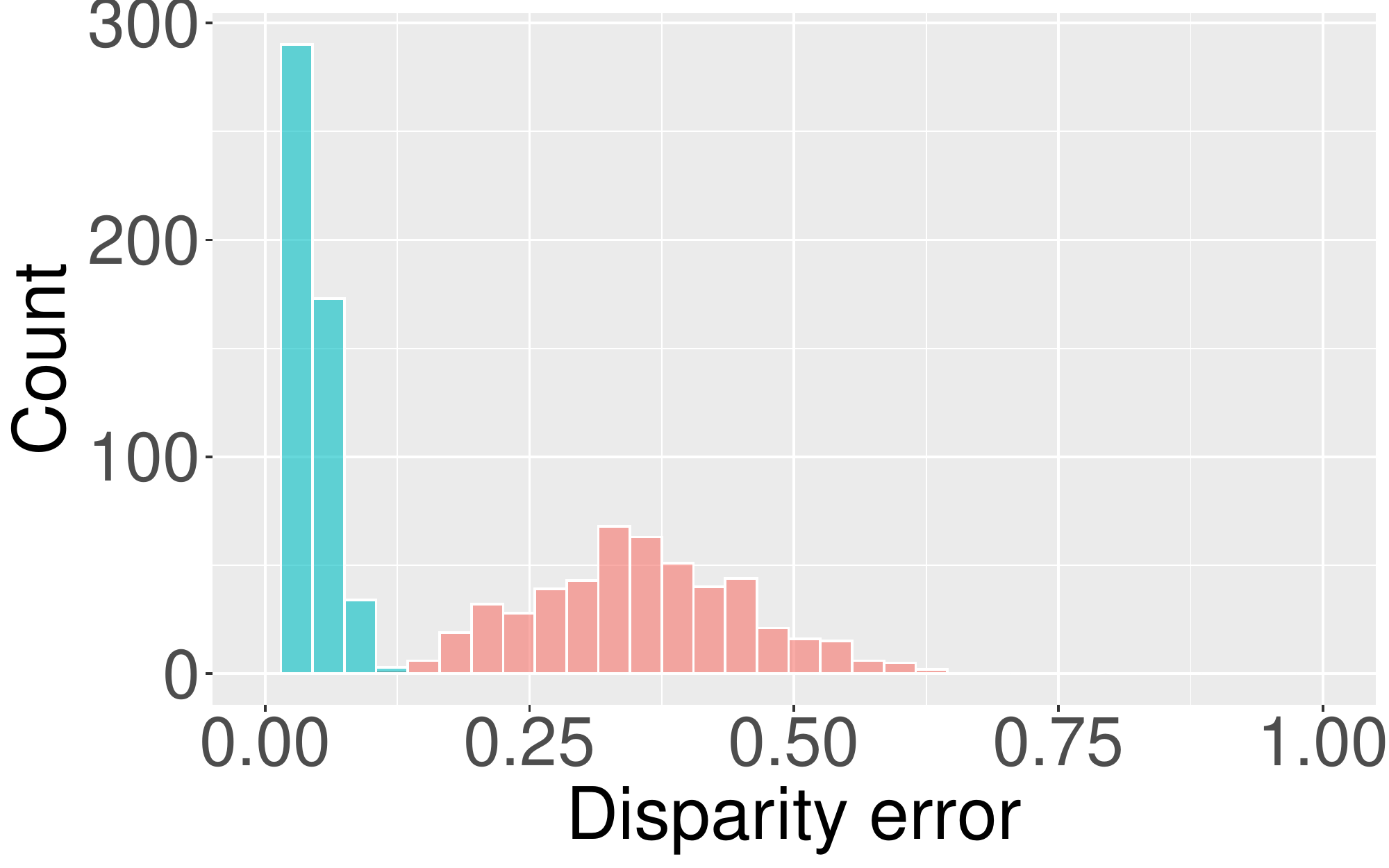}}
	\hfil
	\subfloat[]{
		\label{fig:s0_s1_2_d}
		\includegraphics[width=0.23\textwidth]{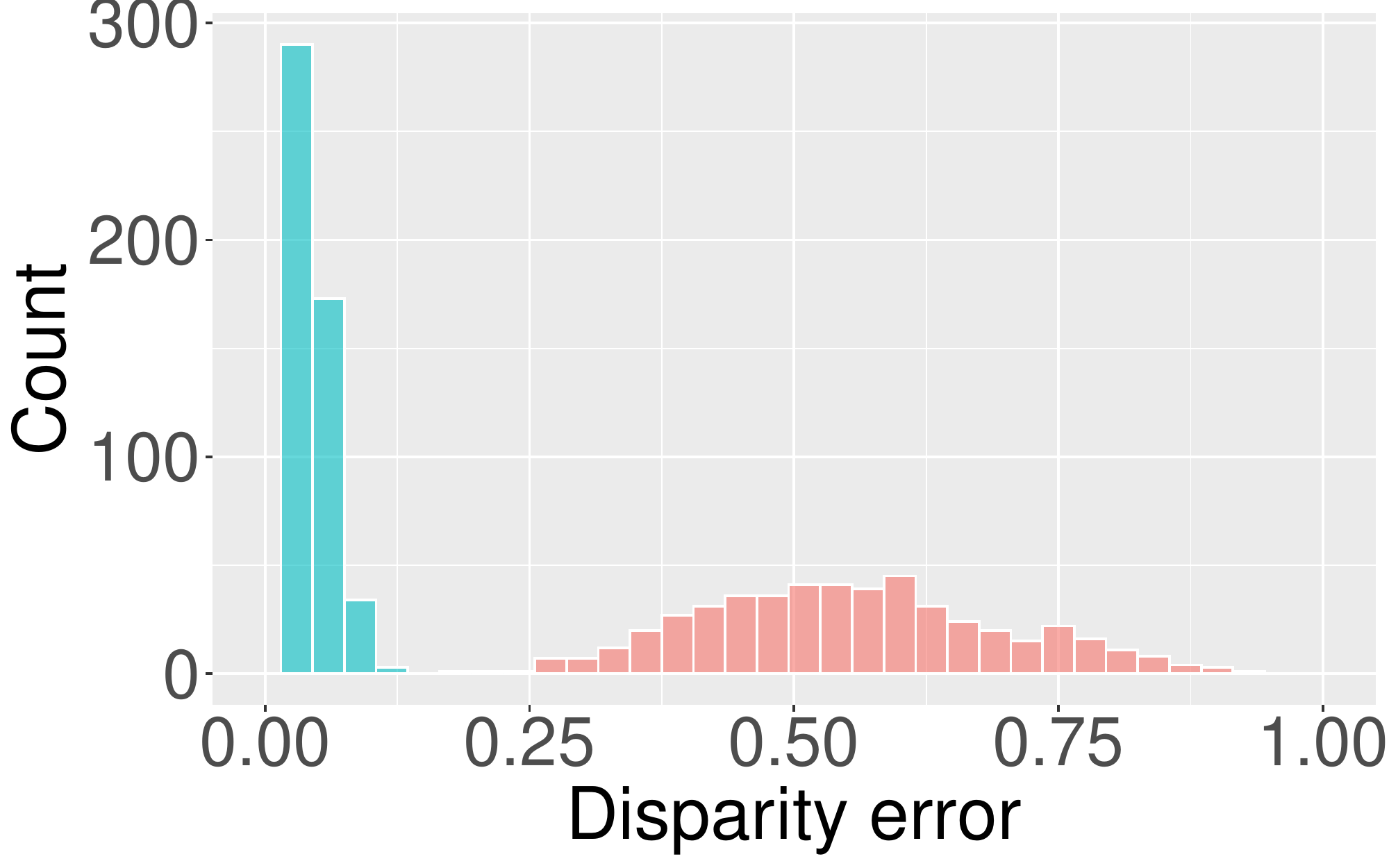}}
	\hfil
	\vspace{-0.1in}
	\subfloat[]{
		\label{fig:s0_s2_2_d}
		\includegraphics[width=0.23\textwidth]{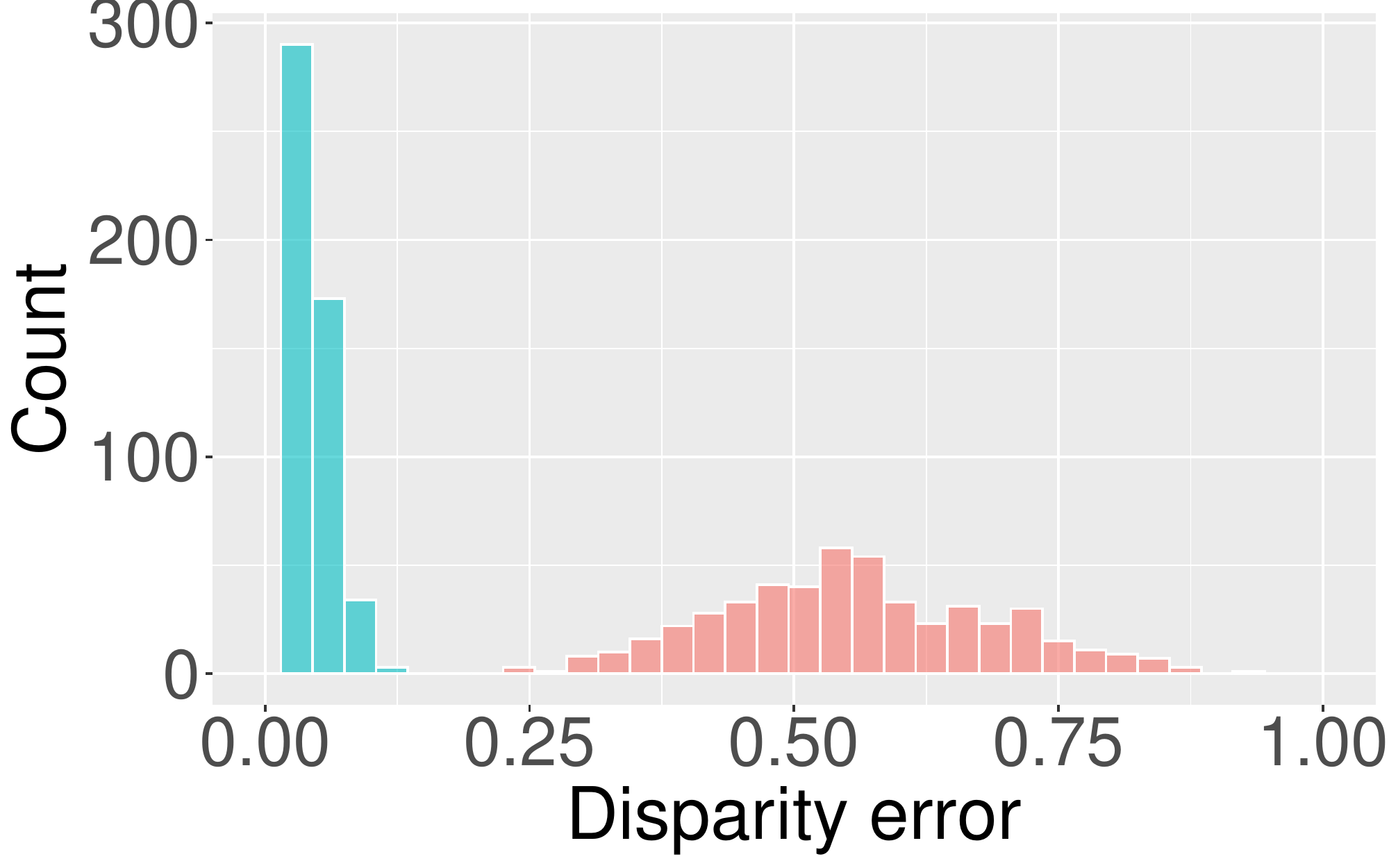}}
	\hfil
	\subfloat[]{
		\label{fig:s1_s2_2_d}
		\includegraphics[width=0.23\textwidth]{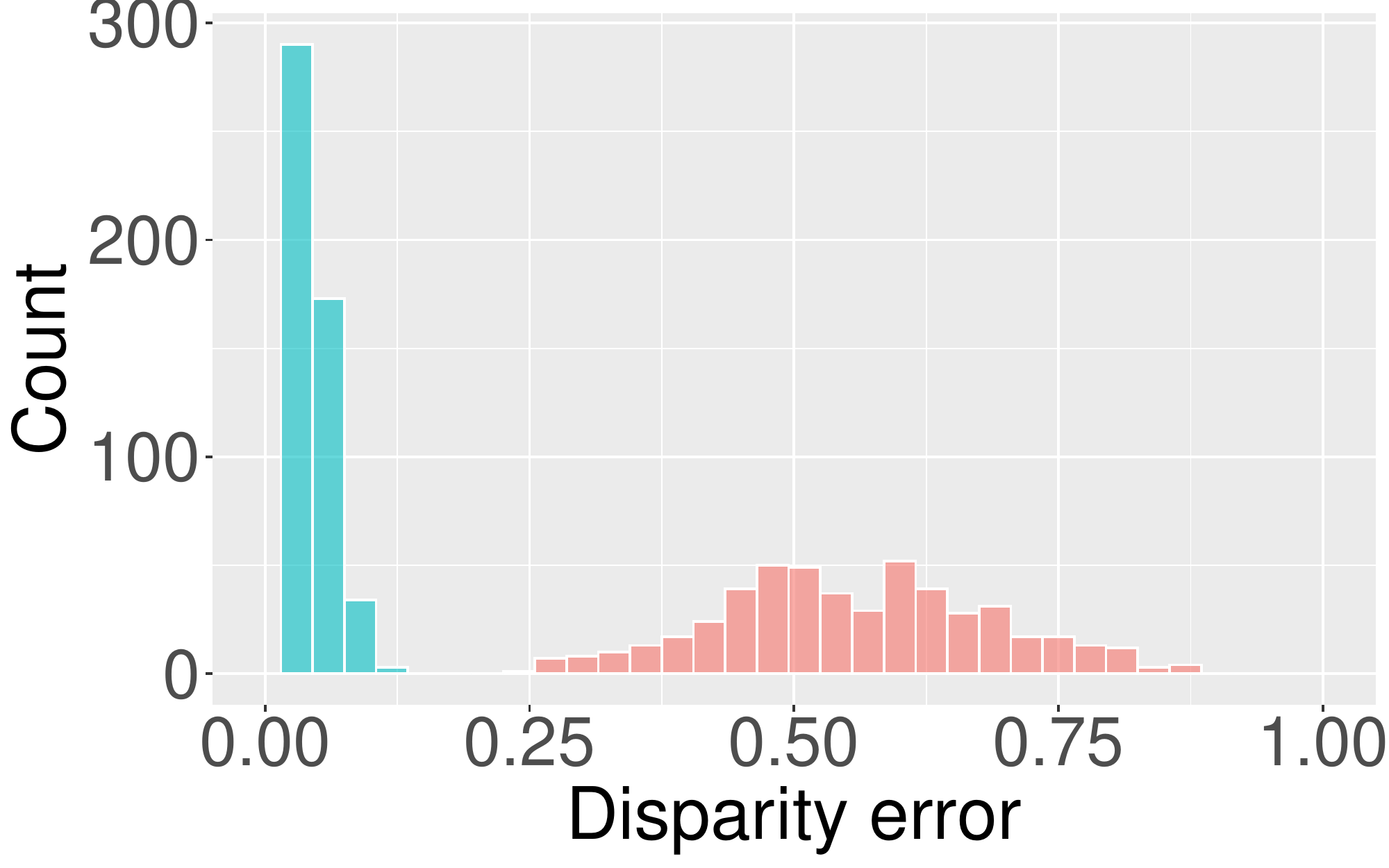}}
	\hfil
	\subfloat[]{
		\label{fig:s0_s1_s2_2_d}
		\includegraphics[width=0.23\textwidth]{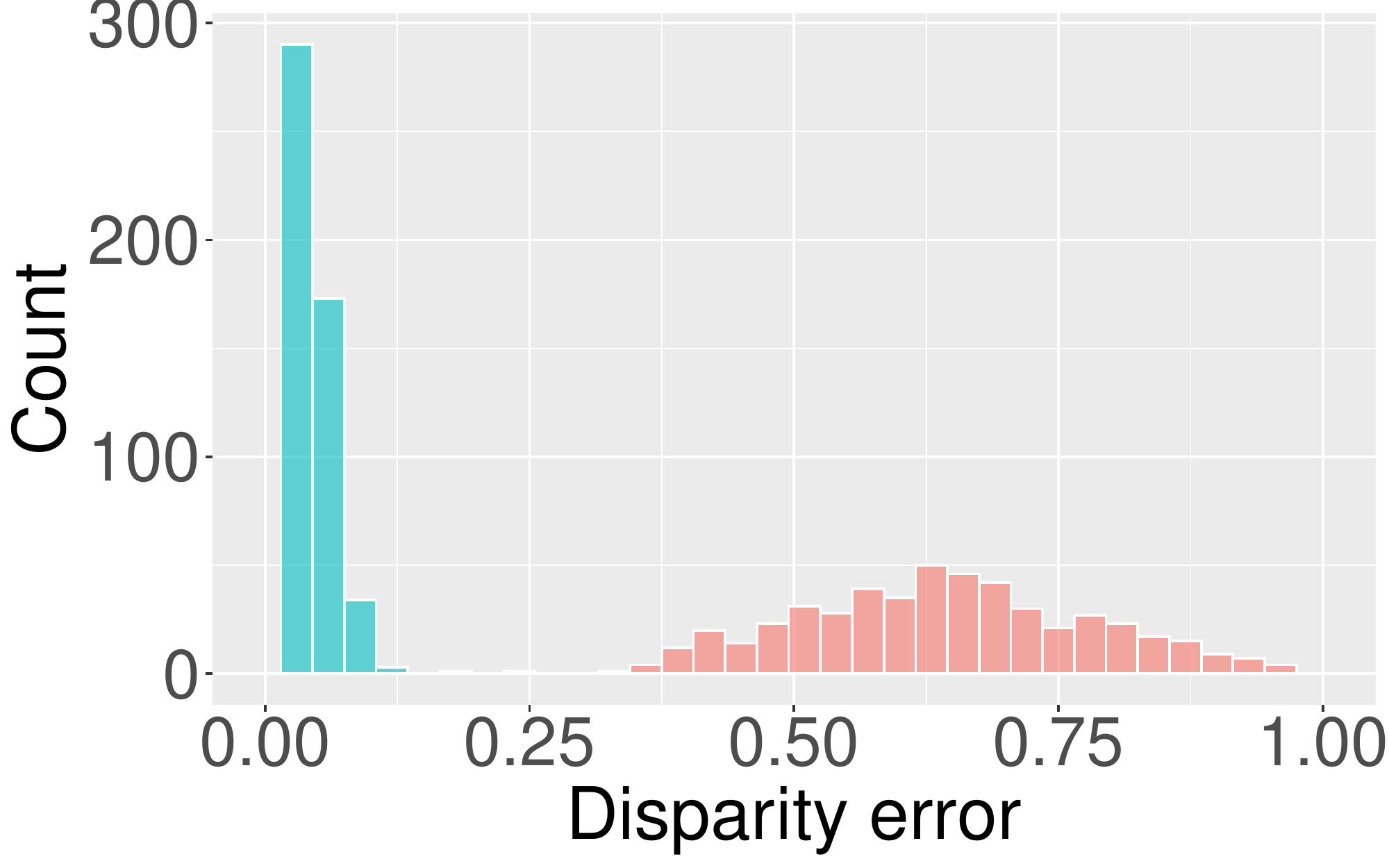}}
	\caption{Distributions of disparity error $E_{0,1,2}$ in normal case (cyan bars) and in attack cases (red bars) for Scenario 2. (a)~No attack vs. $S_{0}$ attacked; (b)~No attack vs. $S_{1}$ attacked; (c)~No attack vs. $S_{2}$ attacked; (d)~No attack vs. $S_{0},S_{1}$ attacked; (e)~No attack vs. $S_{0},S_{2}$ attacked; (f)~No attack vs. $S_{1},S_{2}$ attacked; (g)~No attack vs. $S_{0},S_{1},S_{2}$ attacked.}
	\label{fig:scen_2_exp_d}
	\vspace{-0.15in}
\end{figure*}

\begin{figure*}
	\centering
	\subfloat[]{
		\label{fig:s0_2_r}
		\includegraphics[width=0.23\textwidth]{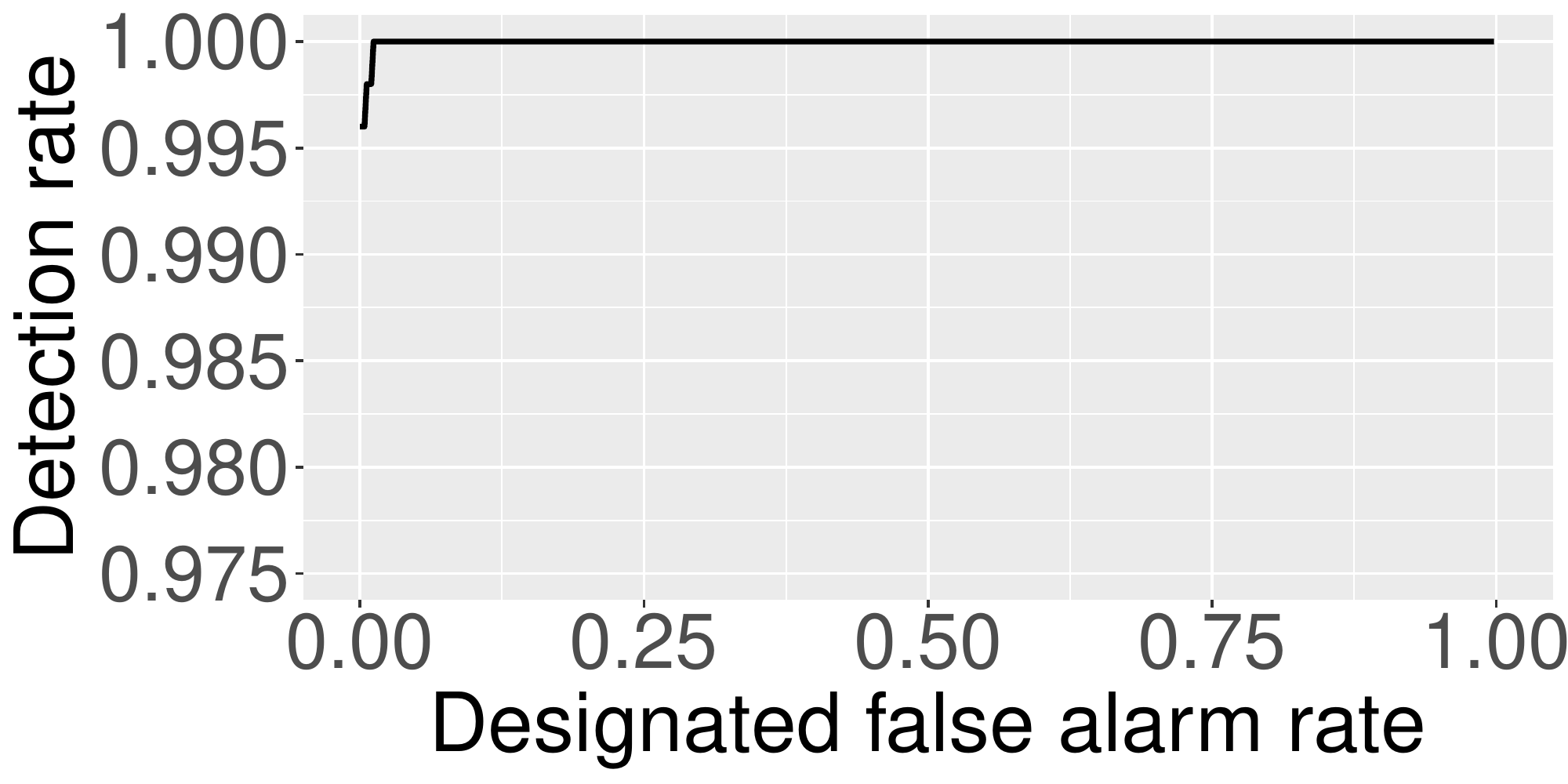}}
	\hfil
	\subfloat[]{
		\label{fig:s1_2_r}
		\includegraphics[width=0.23\textwidth]{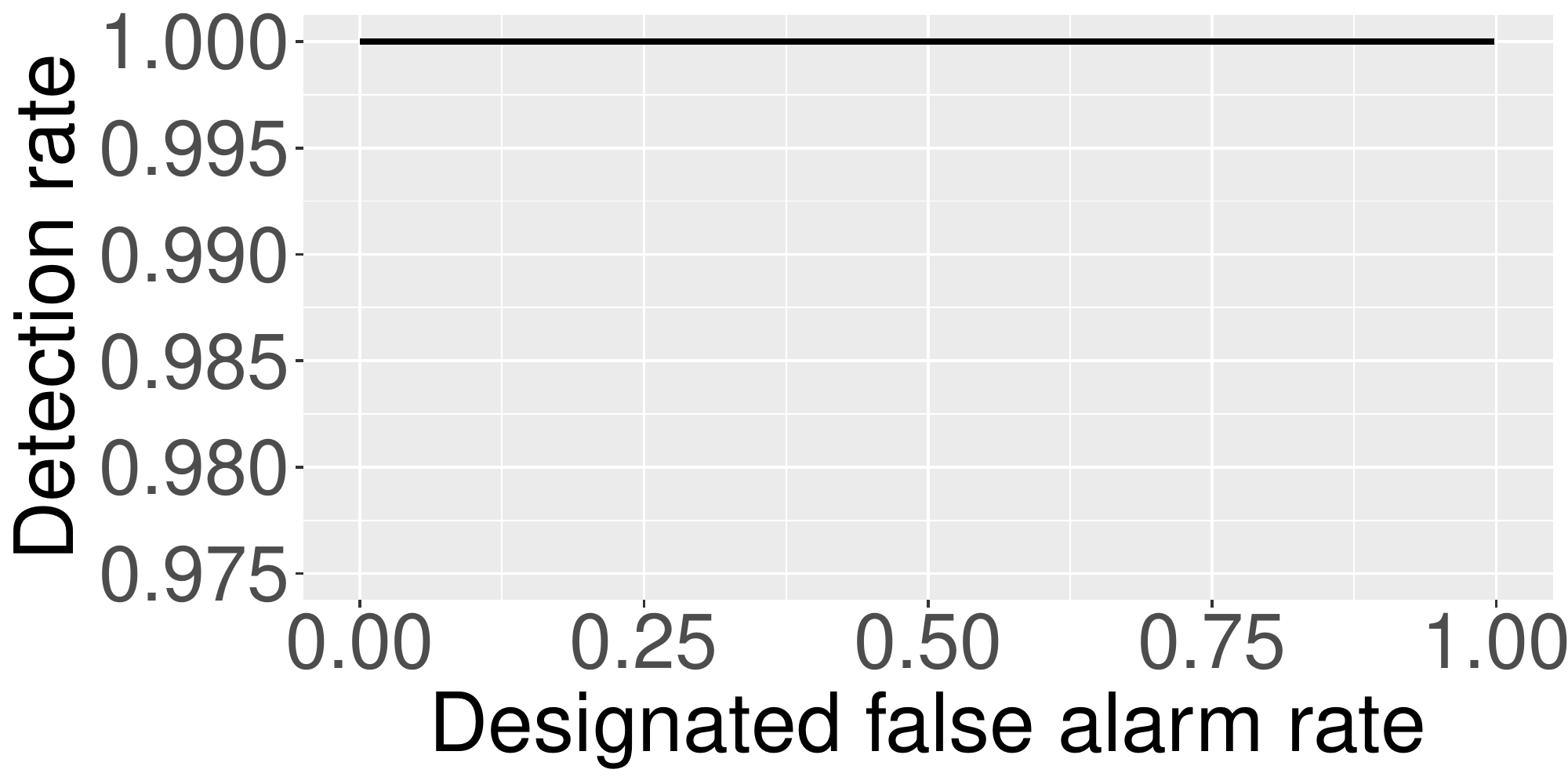}}
	\hfil
	\subfloat[]{
		\label{fig:s2_2_r}
		\includegraphics[width=0.23\textwidth]{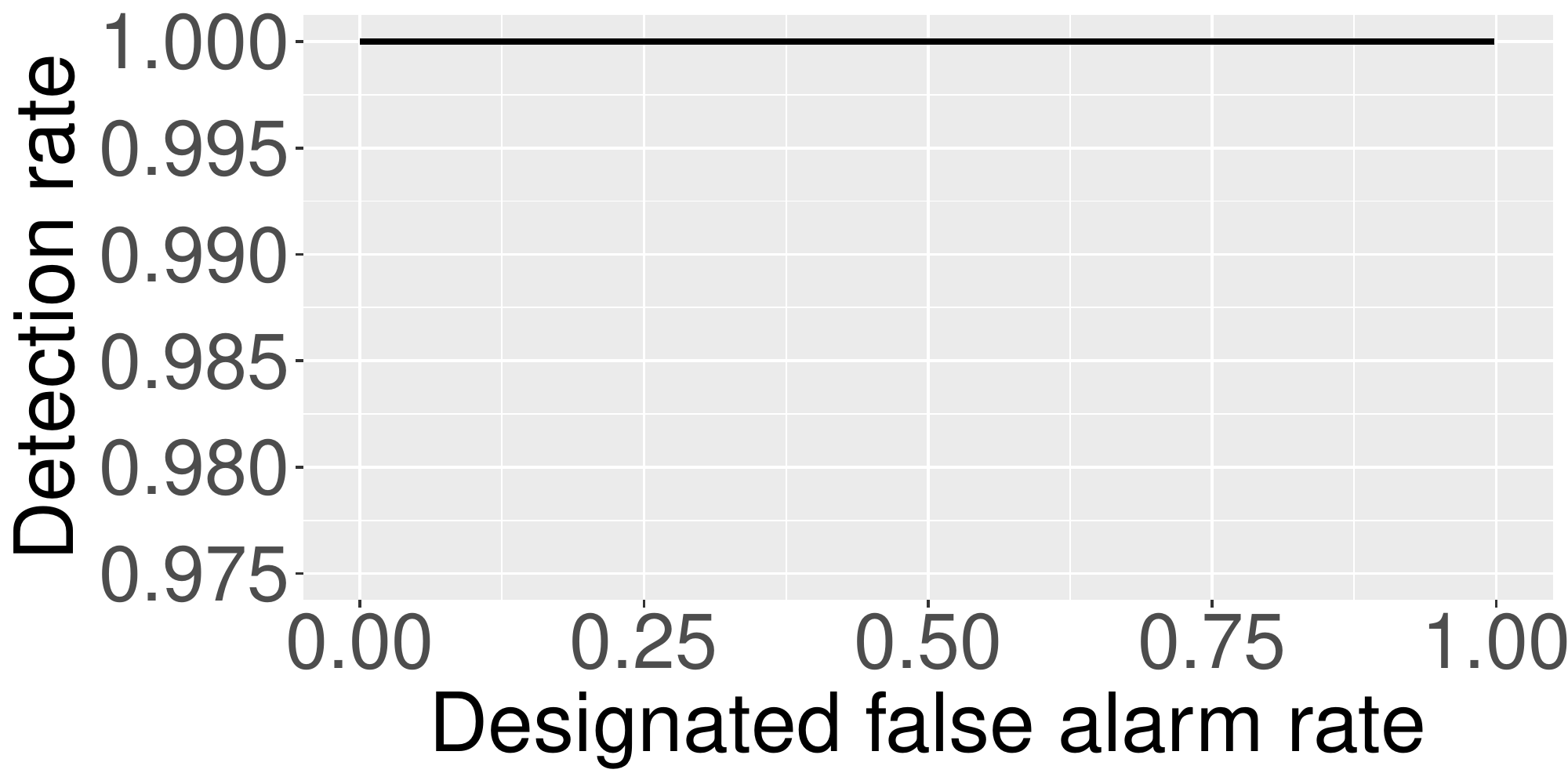}}
	\hfil
	\subfloat[]{
		\label{fig:s0_s1_2_r}
		\includegraphics[width=0.23\textwidth]{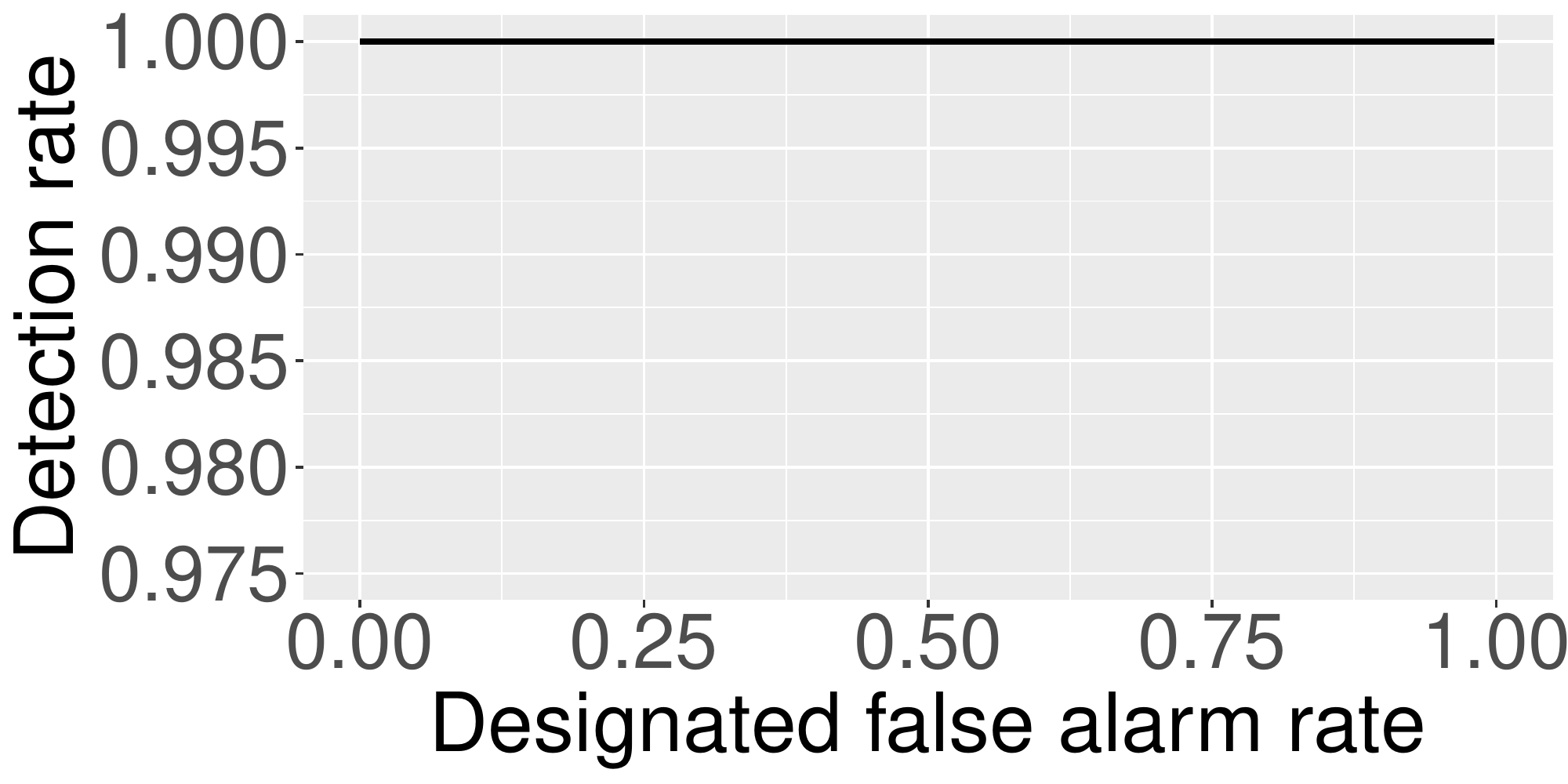}}
	\hfil
	\vspace{-0.1in}
	\subfloat[]{
		\label{fig:s0_s2_2_r}
		\includegraphics[width=0.23\textwidth]{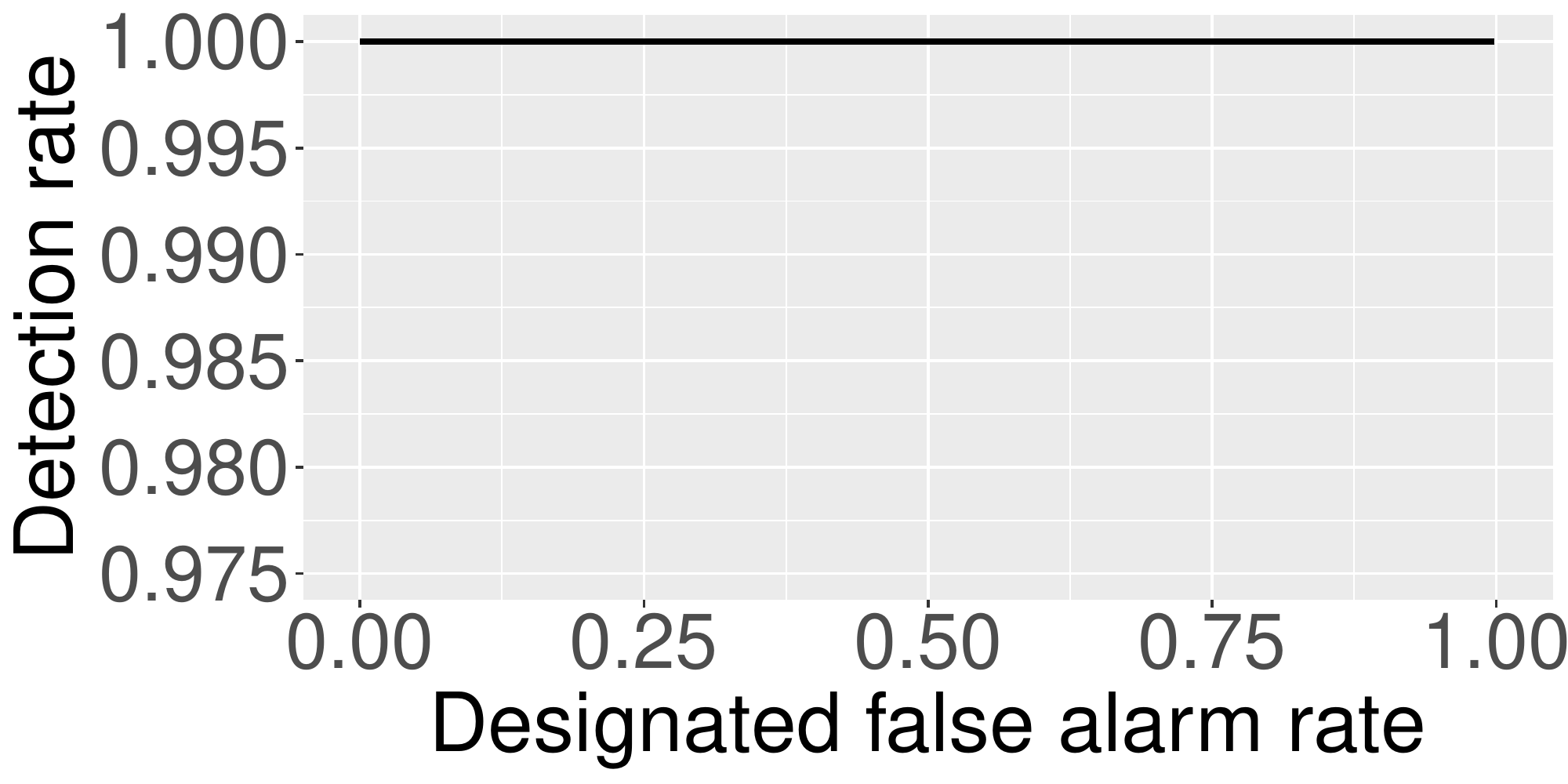}}
	\hfil
	\subfloat[]{
		\label{fig:s1_s2_2_r}
		\includegraphics[width=0.23\textwidth]{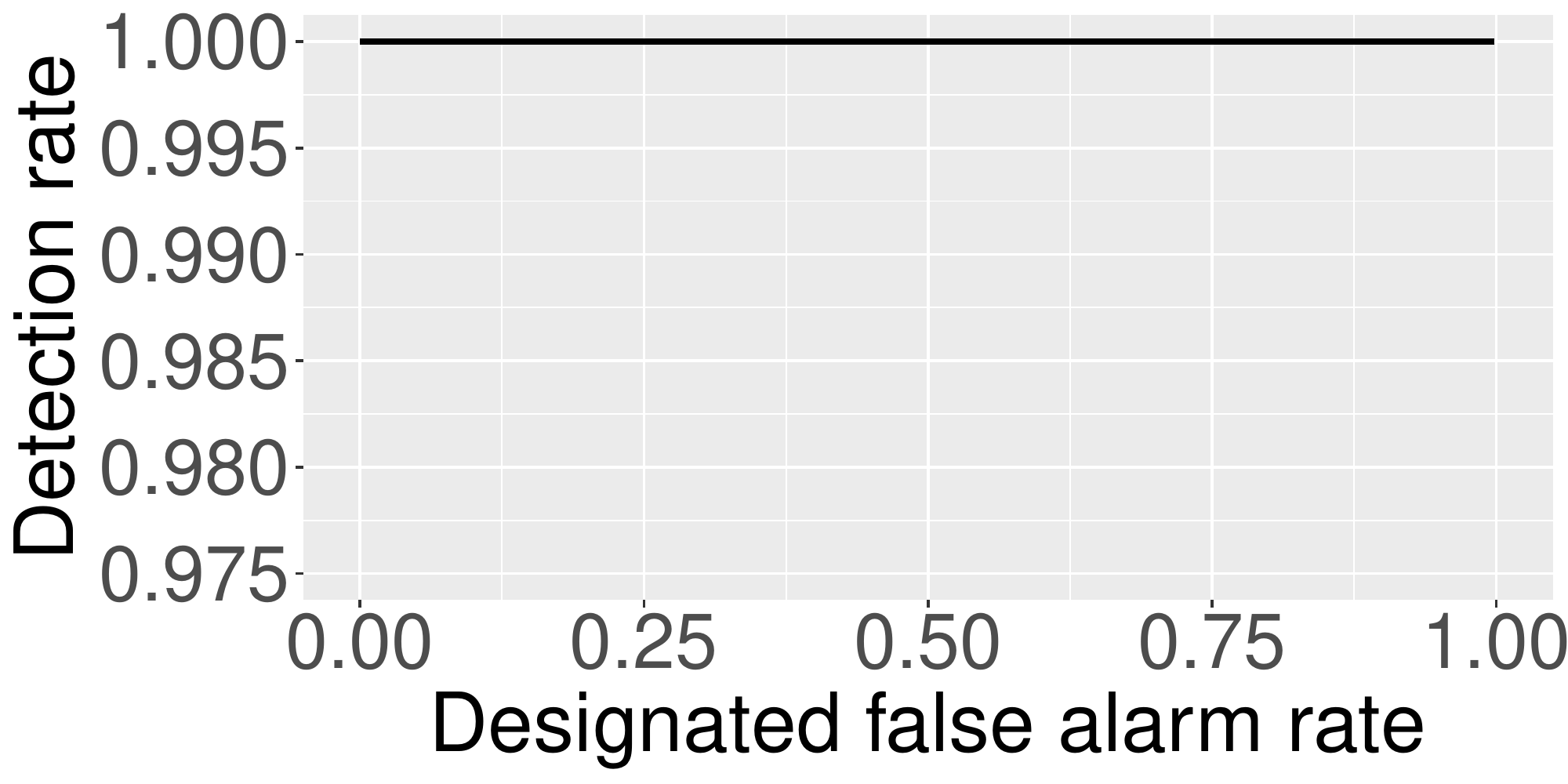}}
	\hfil
	\subfloat[]{
		\label{fig:s0_s1_s2_2_r}
		\includegraphics[width=0.23\textwidth]{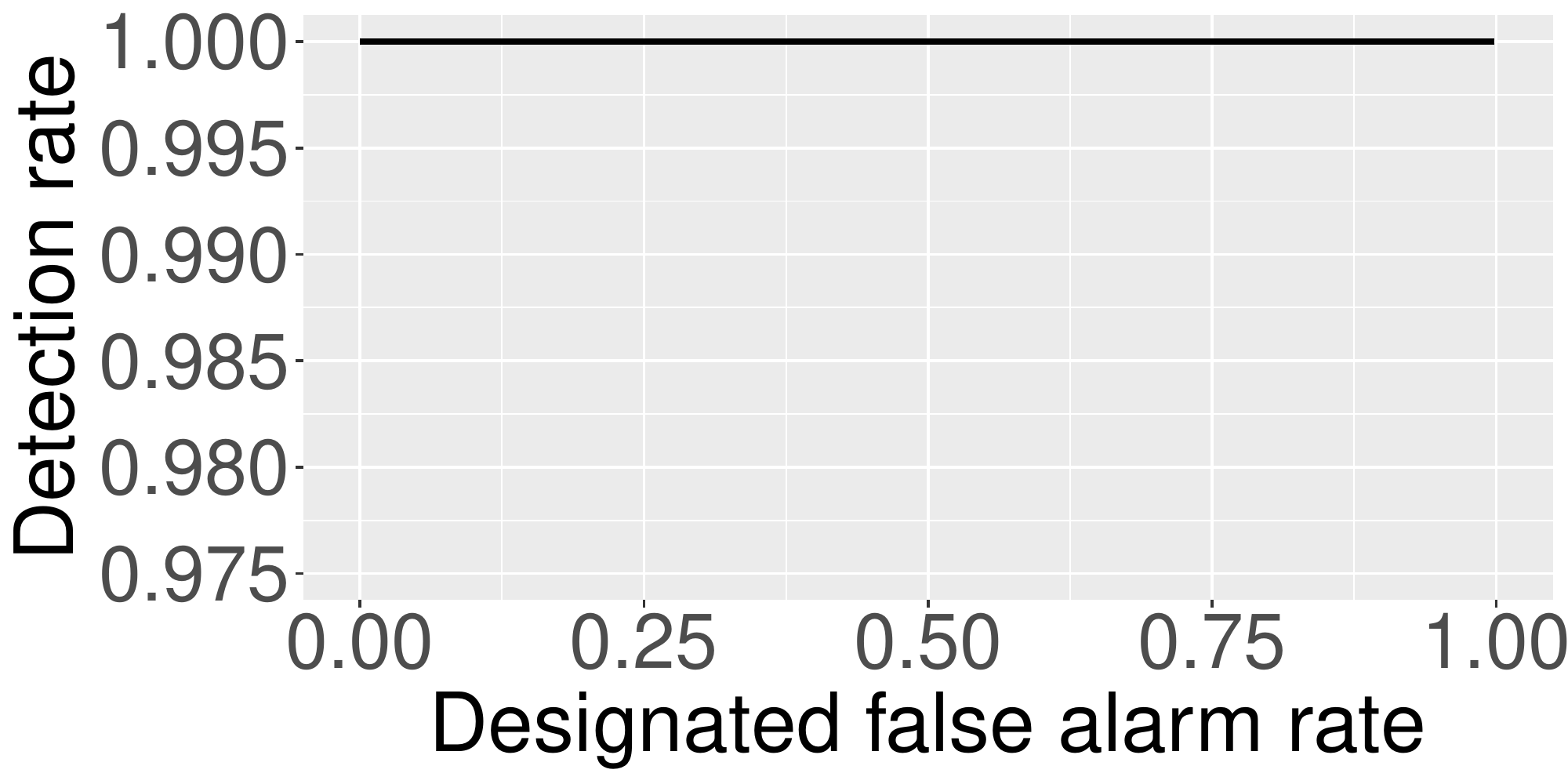}}
	\caption{Detection rate varies with the designated false alarm rate $r$ in each attack case for Scenario 2. (a)~$S_{0}$ is attacked; (b)~$S_{1}$ is attacked; (c)~$S_{2}$ is attacked; (d)~$S_{0}$ and $S_{1}$ are attacked; (e)~$S_{0}$ and $S_{2}$ are attacked; (f)~$S_{1}$ and $S_{2}$ are attacked; (g)~$S_{0}$, $S_{1}$ and $S_{2}$ are attacked.}
	\label{fig:scen_2_exp_r}
	\vspace{-0.15in}
\end{figure*}

\subsubsection{Results}

The experimental results are shown in Fig.~\ref{fig:scen_1_exp_d} and Fig.~\ref{fig:scen_1_exp_r}. In Fig.~\ref{fig:scen_1_exp_d}, we compare the distributions of disparity errors between the normal case (i.e., no attack, in cyan bars) with each one of the seven possible attack cases (red bars). We first observe that the disparity errors are smaller than $20\%$ in most normal cases. By comparison, the disparity errors in most attack scenarios are larger than $25\%$. These results indicate that our detection scheme is sensitive enough, so that there is almost no overlap between the distribution in the normal case and the distributions in those attack cases.

In Fig.~\ref{fig:scen_1_exp_r}, we adjust the threshold $\theta_{0,1,2}$ used to declare attacks by varying the designated false alarm rate $r$, and evaluate the attack detection rate versus $r$. As we can see from the figures, among seven attack scenarios, the performance is perfect in five cases, where the detection rate hits $100\%$ for all possible values of $r$. And even in the non-perfect cases (i.e., (b) and (f)), the proposed detection system can obtain more than $99.5\%$ detection rate with less than $5\%$ false alarm. Such results confirm our hypothesis for such a three-sensor system.

We also show the detection rate comparison between our proposed detection method ($r \leq 5\%$) and the baseline for Scenario 1 in Table~\ref{tbl:dete_exp_r}, where we can observe that our method outperforms the baseline by about $35\%$ with IoU $=0.5$ and about $40\%$ with IoU $=0.7$. The reason is that our method detects optical attack by measuring \emph{pixel-level} disparity inconsistencies, which is much denser and more fine-grained than the \emph{object-level} detection used in the baseline. In most cases where the attack only partially occludes important objects or does not occlude them at all, the \emph{object-level} detection is highly likely to fail, while our proposed method can still function normally.

\vspace{-0.15in}
\subsection{Scenario 2: Three Cameras}

For this scenario, we consider a three-sensor system that consists of three cameras, denotes as $S_{0}$, $S_{1}$, and $S_{2}$, from the right to the left. Similar to the previous scenario, we also consider the data generated by the sensors as $D_{0}$, $D_{1}$, and $D_{2}$. The detection system that we design for this scenario is shown as the combination of Block B and Block C in Fig.~\ref{fig:detection}.

In the system illustrated in Fig.~\ref{fig:detection} (Block B \& Block C), we designate camera $S_2$ as the reference camera to generate two disparity maps. In our experiments, we find it more convenient to implement our detection scheme when the leftmost or rightmost camera is used as the reference camera.

Since the sensor data of this scenario are all images, to generate the two disparity maps, we feed $D_{2}$ with $D_{0}$ and $D_{1}$ to the depth estimation model, respectively. It shall be noted that, since the distance between $S_0$ and $S_2$ is usually different from the distance between $S_1$ and $S_2$, we need to adjust the disparity in $DM_{0,2}$ by updating the baseline $b$ accordingly. After the disparity maps $DM_{0,2}$ and $DM_{1,2}$ are generated, the rest of the procedures in the detection method are the same as those in the previous scenario.

\vspace{-0.15in}
\subsection{Experiments for Scenario 2}

\subsubsection{Setup}

Here, we use the data of three cameras in the customized KITTI raw dataset~\cite{geiger13vision} and also consider all possible attack cases. The baseline method is implemented with the backbone of Faster R-CNN~\cite{ren15faster}. The rest of settings are the same as those in the experiments for Scenario~1.

\subsubsection{Results}

We show the experiment results in Fig.~\ref{fig:scen_2_exp_d} and Fig.~\ref{fig:scen_2_exp_r}. In Fig.~\ref{fig:scen_2_exp_d}, we compare the distributions of disparity errors between the normal case (i.e., no attack, in cyan bars) with each one of the seven possible attack cases (red bars). Similar to Scenario~1, we observe that the disparity errors are less than $10\%$ in virtually all normal cases. By comparison, the disparity errors in $99\%$ of attack scenarios are larger than $12.5\%$.

In Fig.~\ref{fig:scen_2_exp_r}, we vary the threshold $\theta_{0,1,2}$ used to declare attacks via $r$, and evaluate the attack detection rate versus the designated false alarm rate. We can observe that the detection performance is perfect in all cases, where the detection rate remains $100\%$ when $r$ varies from $0$ to $1$. Comparing the results in Scenario 2 with results in Scenarios 1, we notice that the detection performance in scenario 2 is slightly better. We believe this is due to the facts that the disparity maps in this scenario are generated using the same method and there are more valid pixels in the comparison.

In Table~\ref{tbl:dete_exp_r}, the performance comparison for this scenario still shows that our proposed detection method outperforms the baseline by a large margin (more than $30\%$), which again shows the merit of \emph{pixel-level} detection.

\vspace{-0.2in}
\subsection{Empirical Findings}

To briefly summarize, the findings from the attack detection experiments for the two three-sensor systems are listed as follows:

\begin{itemize}
    \item 
    The experimental results confirm our hypothesis that there exists a detection system that can detect optical attacks on the two three-sensor systems with high accuracy and low false alarm rate.
    
    \item
    The detection rate is insensitive to the designated false alarm rate. As long as the detection rate is maintained at a high level, the designated false alarm rate should be set as low as possible, empirically less than $5\%$.
    
    \item
    In those two three-sensor systems, any sensor or any combination of sensors being attacked can cause the disparity error beyond the threshold.
\end{itemize}

Based on these findings, we further develop the identification approach for the proposed framework.

\vspace{-0.1in}
\section{Attack Identification} \label{sec.identification}

In this section, we present the second procedure of our framework which identifies the compromised sensors in a system with one LiDAR $S_{0}$ and $n$ cameras, namely, $S_{1}$ to $S_{n}$ from the right to the left, where $n \geq 3$, based on the empirical findings from the detection method. This method is inspired by Error Correction Codes (ECC) and can achieve the identification as long as there are no more than $n-2$ sensors being attacked simultaneously. In addition, we also demonstrate the proof of the correctness of our identification method, as well as show its effectiveness and accuracy via experiments.

We now introduce a few definitions that are used in the rest of this section. For every sensor $S_{i}$, its state $s_{i}$ can switch between normal state and attack state
\begin{equation}\label{eq:sensor_state}
s_{i}=\begin{cases}
    1, \text{ if } S_{i} \text{ is attacked,}\\
    0, \text{ otherwise,}
\end{cases}
\end{equation}
where $i\in\{0,1,\cdots,n\},n \geq 3$. The sensor state vector in the system is defined as:
\begin{equation}\label{eq:sensor_state_vector}
\textbf{\textit{s}}:=[s_{0},s_{1},\cdots,s_{n}],
\end{equation}
which is the hidden ground truth that we try to identify.

For disparity error $E_{i,j,k}$ among sensors $S_{i},S_{j}$, and $S_{k}$, we use $e_{i,j,k}$ to indicate whether $E_{i,j,k}$ exceeds the corresponding threshold $\theta_{i,j,k}$ in which
\begin{equation}\label{eq:error_state}
e_{i,j,k}=\begin{cases}
    1, \text{ if }E_{i,j,k}>\theta_{i,j,k},\\
    0, \text{ otherwise},
\end{cases}
\end{equation}
where $i,j,k \in \{0,1,\cdots,n\},i<j<k$. And similarly, we use the disparity error state vector $\textbf{\textit{e}}$ to represent the states of disparity errors of three-sensor combinations in the system.

Since the system consists of one LiDAR and $n$ cameras, the combination of any three sensors from it must be either one LiDAR with two cameras or three cameras. According to the empirical finding drawn from the experiments in the previous section, any sensor or any combination of sensors from such three-sensor sets being compromised leads to the corresponding disparity error higher than the threshold. Therefore, based on the definitions in Eqn.~(\ref{eq:sensor_state}) and Eqn.~(\ref{eq:error_state}), we have
\begin{equation}\label{eq:finding}
e_{i,j,k}=(s_{i} \lor s_{j} \lor s_{k}),
\end{equation}
where $i,j,k \in \{0,1,\cdots,n\},i<j<k$, and $\lor$ is logical OR operation.

\vspace{-0.1in}
\subsection{Calculation of Disparity Error State Vector}

Given $n+1$ sensors, we use the leftmost sensor $S_{n}$ as the reference camera and calculate disparity error state vector
\begin{equation}\label{eq:error_state_vector}
\textbf{\textit{e}}:=[e_{0,1,n},e_{0,2,n},\cdots,e_{n-2,n-1,n}],
\end{equation}
where $n \geq 3$. In the calculation, the disparity maps generated by $S_{n}$ with every remaining sensor are compared with each other using the same standard described by Eqn.~(\ref{eq:disparity_error}). Then, by the definition in Eqn.~(\ref{eq:error_state}), $\textbf{\textit{e}}$ is obtained via thresholding the resulted disparity errors from comparison. We also show this calculation process in the form of pseudocode in Algorithm~\ref{algo:cal_e} that takes the data of one LiDAR and $n$ cameras and a list of thresholds as inputs and outputs the disparity error state vector $\textbf{\textit{e}}$. Specifically, it first generates $n$ disparity maps with the sensor data and compares each two of them to obtain disparity errors, and then calculates $\textbf{\textit{e}}$ by encoding disparity errors with thresholds.

Note that the thresholds for calculating $\textbf{\textit{e}}$ are also determined offline using one designated false alarm rate $r$. The approach is similar as in the detection procedure. For every disparity error, we collect sufficient samples when the system is safe and consider $r \times 100\%$ of samples as virtual outliers. The thresholds are then set to the maximal values of inliers. Hence, Eqn.~(\ref{eq:detection_threshold}) can be rewritten as:
\begin{equation}
    \frac{\text{\# samples of } E_{i,j,n} > \theta_{i,j,n}}{\text{\# samples of } E_{i,j,n}} = r,
\end{equation}
where $r \in [0,1]$, $i,j \in \{0,1,\cdots,n-1\},i<j$. Unlike the detection rate which is insensitive to $r$, our subsequent experiments indicate that the identification rate drops linearly with $r$ increasing and the best identifying performance is achieved when $r=1\%$.

\vspace{-0.15in}
\subsection{Identification of Sensor State Vector}

\setlength{\textfloatsep}{6pt}

\begin{algorithm}[t]
\caption{\textit{Calculation of disparity error state vector $\textbf{\textit{e}}$ }}\label{algo:cal_e}
\KwIn{$\textbf{\textit{D}}$: a list of sensor data with length equal to $|\textbf{\textit{D}}|$, where $D_{0}$ is point cloud and the rest are images; $\boldsymbol{\theta}$: a list of thresholds.}
\KwOut{$\textbf{\textit{e}}$: the disparity error state vector.}
select $D_{n}$ as reference image, where $n=|\textbf{\textit{D}}|-1$\;
get disparity map $DM_{0,n}$ using $D_{0}$ and $D_{n}$ (Scenario 1 in Section~\ref{sec.detection})\;
\For{$i\gets{1}$ \KwTo $n-1$}{
    get disparity map $DM_{i,n}$ using $D_{i}$ and $D_{n}$ (Scenario 2 in Section~\ref{sec.detection})\;
}
\For{$i\gets{0}$ \KwTo $n-2$}{
    \For{$j\gets{i+1}$ \KwTo $n-1$}{
        get disparity error $E_{i,j,n}$ by comparing $DM_{i,n}$ with $DM_{j,n}$ (Section~\ref{sec.detection})\;
        \If{\text{\normalfont $E_{i,j,n}$ exceeds threshold $\theta_{i,j,n}$}}{
            assign $1$ to disparity error state $e_{i,j,n}$\;
        }
        \Else{
            assign $0$ to $e_{i,j,n}$\;
        }
    }
}
\Return{$\textbf{\textit{e}}$}\;
\vspace{-0.05in}
\end{algorithm}

We now elaborate on how to infer $\textbf{\textit{s}}$ according to $\textbf{\textit{e}}$. We consider the following three cases of $\textbf{\textit{e}}$:
\begin{itemize}
    \item If all elements in $\textbf{\textit{e}}$ are $0$s, according to Eqn.~(\ref{eq:finding}) and Eqn.~(\ref{eq:error_state_vector}), $s_i=0$ for $0\leq i\leq n$. In other words, no sensor is attacked. 
    \item
    If only some elements in $\textbf{\textit{e}}$ are $0$s, we have the following Lemma~\ref{thrm:lem_1} to identify all attacked sensors.
    \item
    If all elements in $\textbf{\textit{e}}$ are $1$s and no more than $n-2$ sensors are attacked simultaneously, we can repeatedly use Lemma~\ref{thrm:lem_1} and Lemma~\ref{thrm:lem_2} to identify all attacked sensors.
\end{itemize}

\begin{lemma}\label{thrm:lem_1}
In a system with $n+1$ sensors, if there exist $i_{0},j_{0}$ such that $e_{i_{0},j_{0},n}=0$, then $s_i=e_{i,i_{0},n}$ for  $0 \leq i < i_{0}$, and $s_{i}=e_{i_{0},i,l-1}$ for $i_0 <i < n$.
\end{lemma}

\begin{figure*}[!t]
    \centering
	\includegraphics[width=0.98\textwidth]{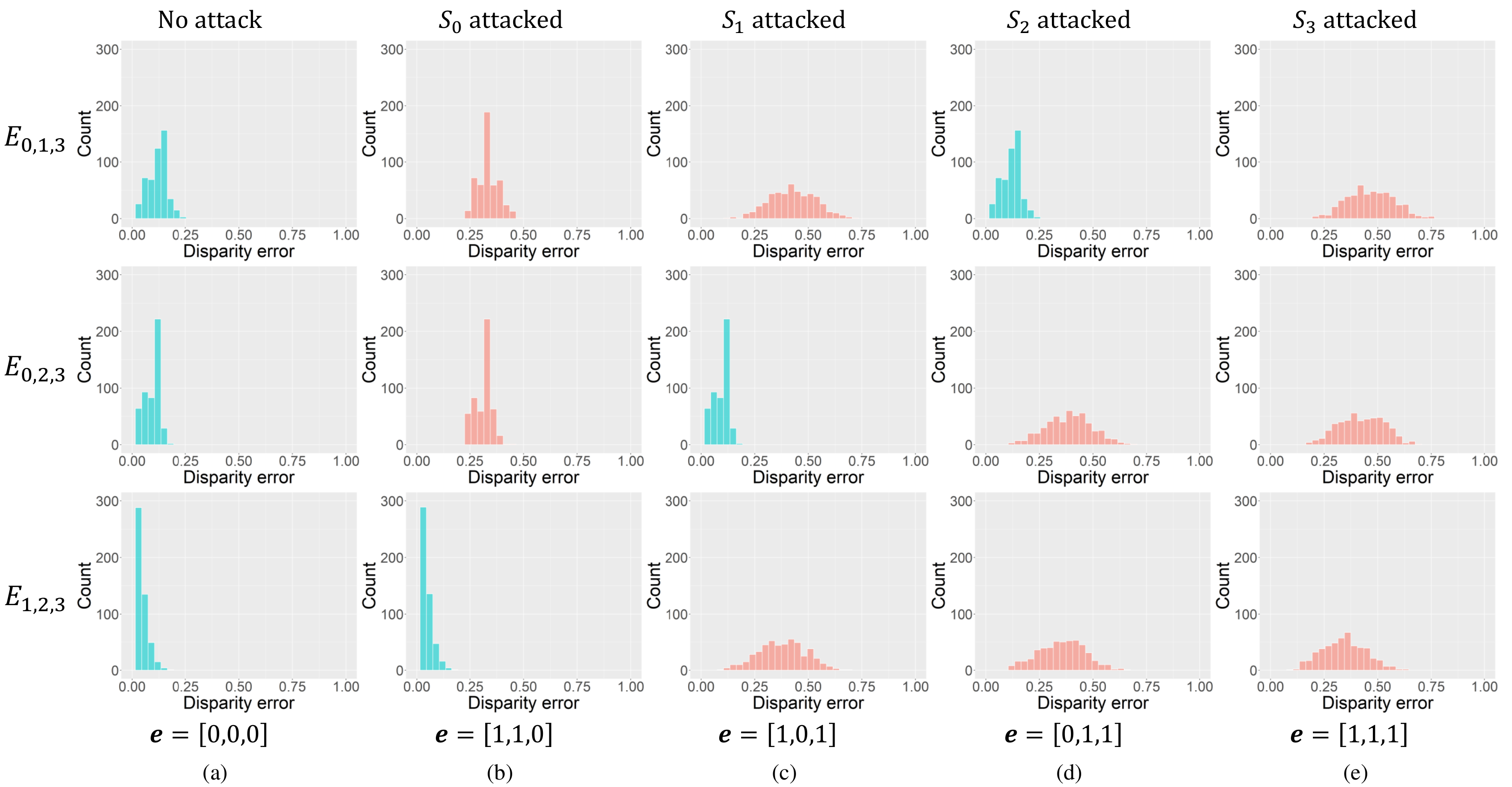}
	\caption{Distributions of disparity error $E_{0,1,3}$ (1st row), disparity error $E_{0,2,3}$ (2nd row) and disparity error $E_{1,2,3}$ (3rd row) in normal scenario (1st column) and in attack scenarios (2nd--5th columns). Cyan bars mean that the disparity errors in a scenario involve no attacked sensor, while red bars indicate that the disparity errors in a scenario involve attacked sensor. (a)~No disparity errors exceed thresholds, so $\textbf{\textit{e}}=[0,0,0]$ indicating no optical attack; (b)~$E_{0,1,3}$ and $E_{0,2,3}$ exceed thresholds, so $\textbf{\textit{e}}=[1,1,0]$ indicating that $S_{0}$ is attacked; (c)~$E_{0,1,3}$ and $E_{1,2,3}$ exceed thresholds, so $\textbf{\textit{e}}=[1,0,1]$ indicating that $S_{1}$ is attacked; (d)~$E_{0,2,3}$ and $E_{1,2,3}$ exceed thresholds, so $\textbf{\textit{e}}=[0,1,1]$ indicating that $S_{2}$ is attacked; (e)~All three disparity errors exceed thresholds, so $\textbf{\textit{e}}=[1,1,1]$ indicating that $S_{3}$ is attacked.}
	\label{fig:iden_exp_d}
	\vspace{-0.15in}
\end{figure*}

\begin{figure*}
	\centering
	\subfloat[]{
		\label{fig:s0_i_r}
		\includegraphics[width=0.23\textwidth]{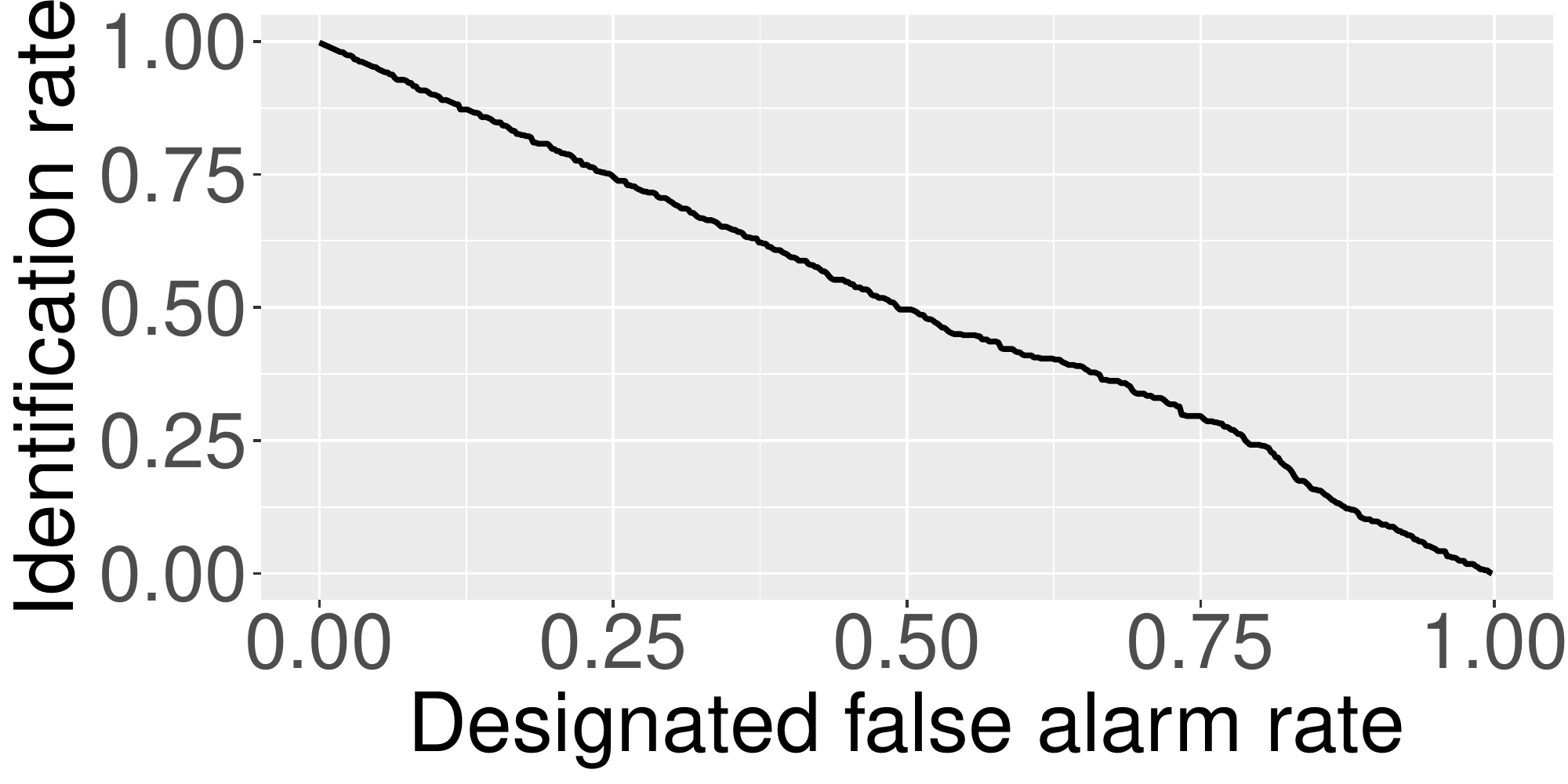}}
	\hfil
	\subfloat[]{
		\label{fig:s1_i_r}
		\includegraphics[width=0.23\textwidth]{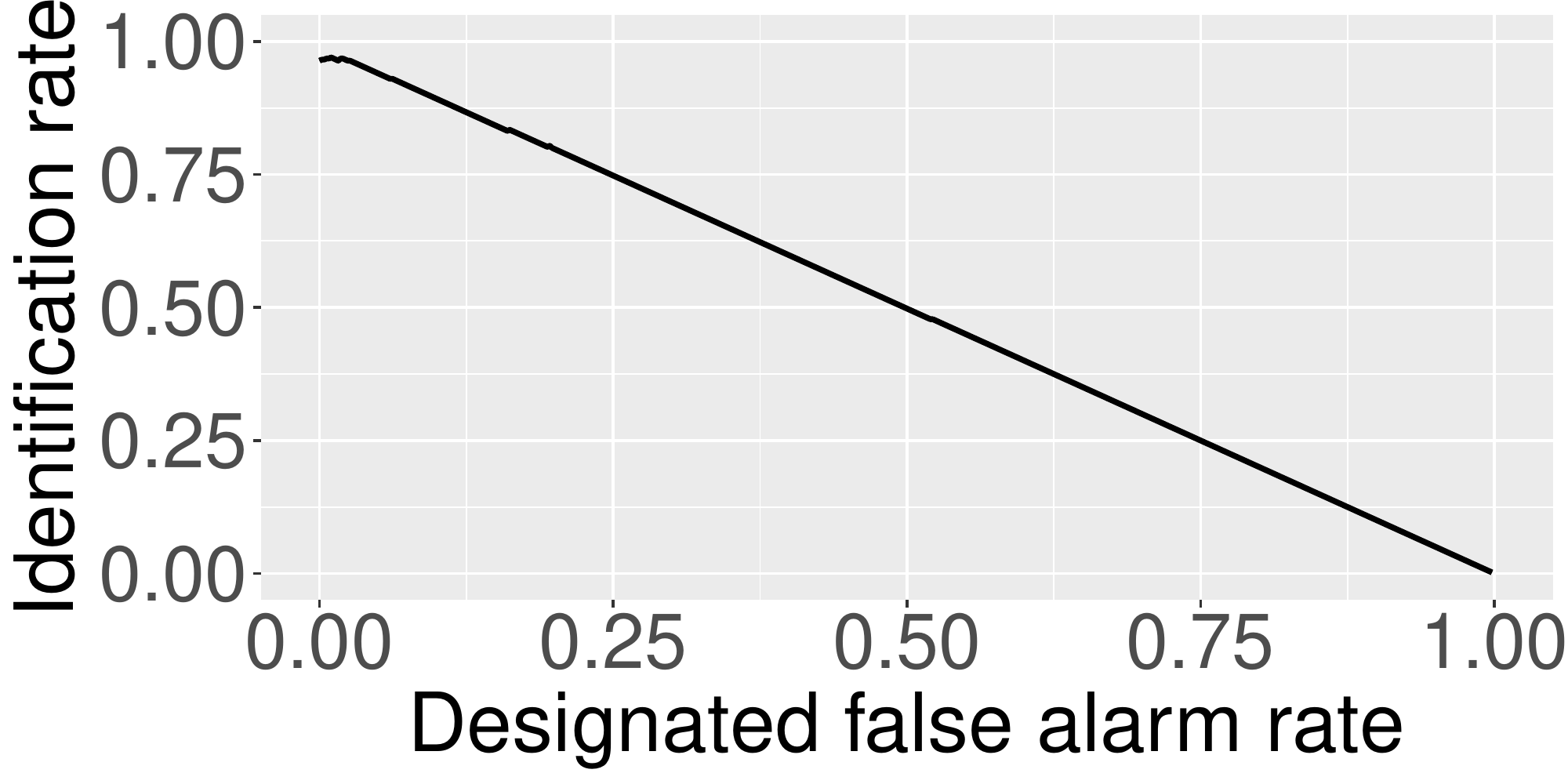}}
	\hfil
	\subfloat[]{
		\label{fig:s2_i_r}
		\includegraphics[width=0.23\textwidth]{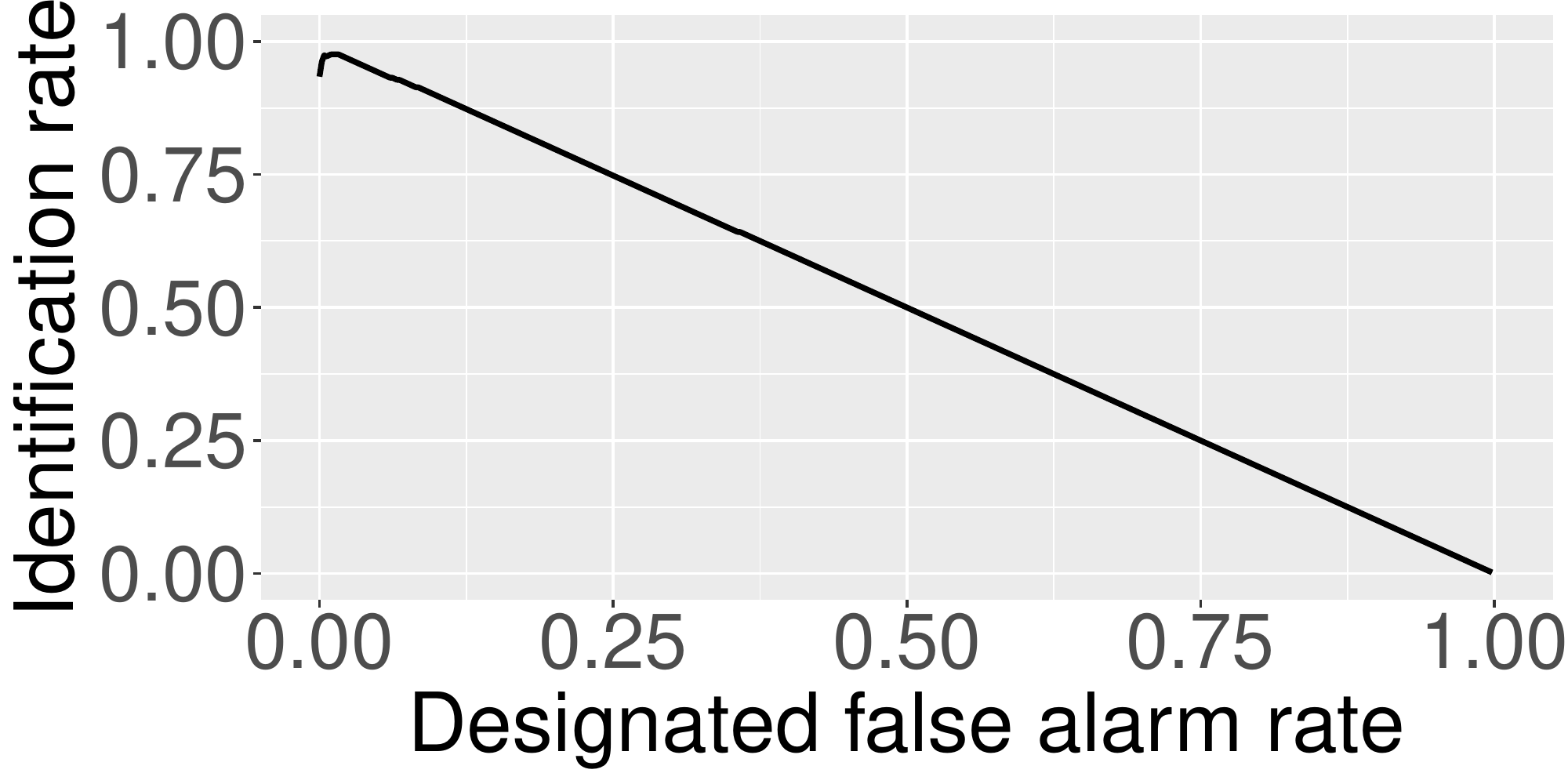}}
	\hfil
	\subfloat[]{
		\label{fig:s3_i_r}
		\includegraphics[width=0.23\textwidth]{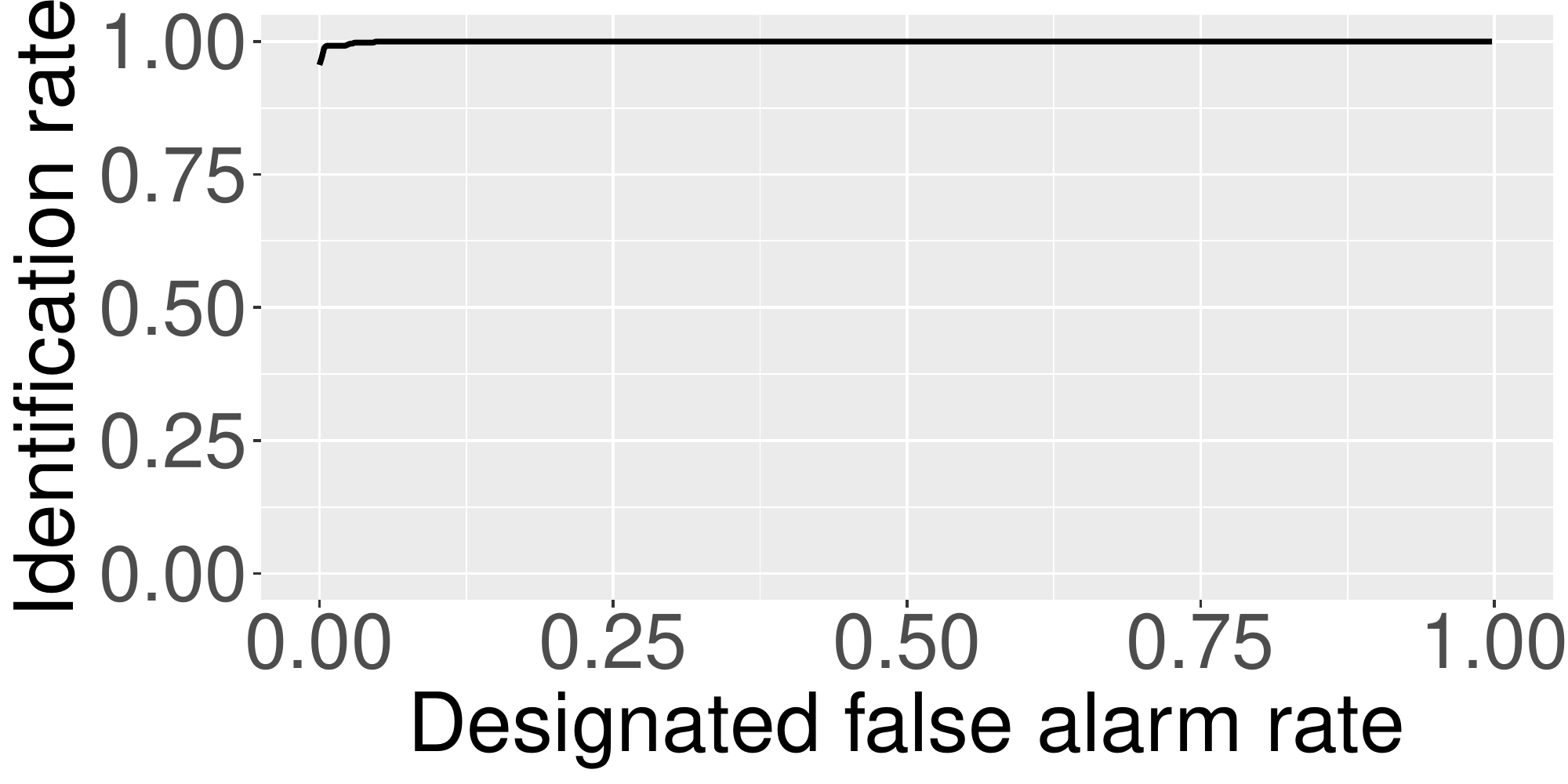}}
	\caption{Identification rate varies with the designated false alarm rate $r$ in each attack scenario. (a)~$S_{0}$ is attacked; (b)~$S_{1}$ is attacked; (c)~$S_{2}$ is attacked; (d)~$S_{3}$ is attacked.}
	\label{fig:iden_exp_r}
	\vspace{-0.15in}
\end{figure*}

\begin{IEEEproof}
If there exist $i_{0},j_{0}$ such that $e_{i_{0},j_{0},n}=0$,
according to Eqn.~(\ref{eq:finding}), we can have:
\begin{equation}
e_{i_{0},j_{0},n}=(s_{i_{0}} \lor s_{j_{0}} \lor s_{n})=0
\end{equation}
which implies
\begin{equation}
s_{i_{0}}=s_{j_{0}}=s_{n}=0
\end{equation}
Now we only need to focus on the state of $s_{0},s_{1}, \ldots, s_{n-1}$.

For any $ i \in \{0,1,\cdots,i_{0}-1\}$,
\begin{equation}
e_{i,i_{0},l-1} = (s_{i} \lor s_{i_{0}} \lor s_{n}) = (s_{i} \lor 0 \lor 0) = s_{i},
\end{equation}
and  for any $ i \in \{i_{0}+1,i_{0}+2,\cdots,n-1\}$,
\begin{equation}
e_{i_{0},i,n} = (s_{i_{0}} \lor s_{i} \lor s_{n}) = (0 \lor s_{i}\lor 0) = s_{i}.
\end{equation}
\end{IEEEproof}

Lemma~\ref{thrm:lem_1} shows that we can identify the sensor state vector $\textbf{\textit{s}}$ if at least one element in $\textbf{\textit{e}}$ is $0$. For the case where all elements in $\textbf{\textit{e}}$ are $1$s, we have the following lemma.

\begin{lemma}\label{thrm:lem_2}
In a system with $n+1$ sensors, when there are no more than $n-2$ sensors being compromised simultaneously, if the elements of $\textbf{\textit{e}}$ are all $1$s, then $s_{n}=1$.
\end{lemma}
 
\begin{IEEEproof}
Since there are no more than $n-2$ sensors being attacked, the states of at least three sensors are $0$s. If $S_{n}$ is normal, then there exist $i^*$ and $j^*$, where $i^*<j^*$, such that sensors $S_{i^*}$ and $S_{j^*}$ are normal, i.e., $s_{i^*}=s_{j^*}=s_{n}=0$. In this case, $e_{i^*,j^*,n}=0$, which contradicts the fact that all the elements of $\textbf{\textit{e}}$ are $1$s. Therefore, $s_{n}=1$.
\end{IEEEproof}

\begin{algorithm}[t]
\caption{\textit{Identification of sensor state vector $\textbf{\textit{s}}$ }}\label{algo:iden_s}
\KwIn {$\textbf{\textit{D}}$: a list of sensor data with length equal to $|\textbf{\textit{D}}|$, where $D_{0}$ is point cloud and the rest are images; $\boldsymbol{\theta}$: a list of thresholds.}
\KwOut{$\textbf{\textit{A}}$: the list of compromised sensors.}
select $D_{n}$ as reference image, where $n=|\textbf{\textit{D}}|-1$\;
calculate $\textbf{\textit{e}}$ using Algorithm~\ref{algo:cal_e} with $\textbf{\textit{D}}$ and $\boldsymbol{\theta}$\;
\If{\text{\normalfont there exists 1 in $\textbf{\textit{e}}$}}{
    \If{\text{\normalfont there exists 0 in $\textbf{\textit{e}}$}}{
        \tcp{use Lemma~\ref{thrm:lem_1} to infer $\textbf{\textit{s}}$}
        find $i_{0},j_{0}$ which satisfy $e_{i_{0},j_{0},n}=0$\;
        \For{$i\gets{0}$ \KwTo $n-1$}{
            assign $\min(i, i_0)$ to $k_1$, $\max(i, i_0)$ to $k_2$\;
            \If{\text{$e_{k_1,k_2,n}=1$ {\normalfont \textbf{and}} $k_1 \neq k_2$}}{
                sensor state $s_{i}$ is 1\;
                push attacked sensor $S_{i}$ into $\textbf{\textit{A}}$\;
            }
        }
    }
    \Else{
        \tcp{use Lemma~\ref{thrm:lem_2} to infer $s_{n}$}
        sensor state $s_{n}$ is 1\;
        push attacked sensor $S_{n}$ into $\textbf{\textit{A}}$\;
        \tcp{infer rest sensors recursively}
        \If{$n>3$}{
            remove $D_{n}$ from $\textbf{\textit{D}}$\;
            go to line $1$ to rerun with updated $\textbf{\textit{D}}$\;
        }
    }
}
\Return{$\textbf{\textit{A}}$}\;
\vspace{-0.05in}
\end{algorithm}

When all elements of $\textbf{\textit{e}}$ in a system with $n+1$ sensors are $1$, though we cannot directly find out the states of all sensors, Lemma~\ref{thrm:lem_2} can identify the last sensor's state. After that, we can virtually remove the last sensor and consider a system with $n$ sensors. We recalculate the $\textbf{\textit{e}}$ by Algorithm~1 for such $n$ sensors, then determine $\textbf{\textit{s}}$ according to Lemma~\ref{thrm:lem_1} and Lemma~\ref{thrm:lem_2}. We repeat this process until the states of all sensors are identified. We also present this identification algorithm as pseudocode in Algorithm~\ref{algo:iden_s} that takes the same inputs as Algorithm~\ref{algo:cal_e} and outputs a list of attacked sensors. Specifically, it begins with calculating $\textbf{\textit{e}}$ using Algorithm~\ref{algo:cal_e} with the inputs, and then infers the sensor state vector $\textbf{\textit{s}}$ using Lemma~\ref{thrm:lem_1} when some elements in $\textbf{\textit{e}}$ are $0$s. When all elements in $\textbf{\textit{e}}$ are $1$s, it first infers $s_{n}$, the state of the reference camera, using Lemma~\ref{thrm:lem_2}, and then excludes the data of the reference camera from the inputs and infers the rest of sensor states by rerunning Algorithm~\ref{algo:iden_s} with the updated inputs.

\vspace{-0.15in}
\subsection{Experiments}

\setlength{\textfloatsep}{18pt}

\begin{table}[t]
    \centering
	\caption{Identification rate at particular values of the designated false alarm rate in attack scenarios}\label{tbl:iden_exp_r}
	\begin{tabular}{|l|c c c c|l|}
	    \hline
	        \multirow{2}{1em}{$~r$}& \multicolumn{4}{c|}{Attacked Sensor} & \multirow{2}{0em}{Average} \\
		\cline{2-5}
		    & $S_{0}$ & $S_{1}$ & $S_{2}$ & $S_{3}$ & \\
		\hline
    	\hline
    	$0\%$ & $\textbf{99.80\%}$ & $96.40\%$ & $93.40\%$ & $95.60\%$ & $96.30\%$ \\
    	$1\%$ & $98.80\%$ & $\textbf{97.00\%}$ & $\textbf{97.60\%}$ & $99.20\%$ & $\textbf{98.15\%}$ \\
    	$2\%$ & $98.00\%$ & $96.80\%$ & $97.20\%$ & $99.20\%$ & $97.80\%$ \\
    	$3\%$ & $96.60\%$ & $96.00\%$ & $96.20\%$ & $99.80\%$ & $97.15\%$ \\
    	$5\%$ & $94.80\%$ & $94.00\%$ & $94.20\%$ & $\textbf{100\%}$ & $95.75\%$\\
    	\hline
	\end{tabular}
	\vspace{-0.15in}
\end{table}

\subsubsection{Setup}

To verify the effectiveness and evaluate the precision of our identification scheme, we conduct substantial experiments. Since the identification scheme is designed to be in a form of recursion, the experiments here are for the base case where a system consists of one LiDAR ($S_{0}$) and $n=3$ cameras ($S_{1}$, $S_{2}$, and $S_{3}$). And according to the constraint, there is at most one attacked sensor in the system.

In the experiments, we consider the normal scenario and all possible attack scenarios where each one of the four sensors gets attacked. We use the data of one LiDAR and three cameras from the customized KITTI raw dataset~\cite{geiger13vision}. To generate the compromised sensor data for each attack scenario, we do the same as in Section~\ref{subsubsec:setup}. The depth estimation model is provided by~\cite{wang19pseudo}. As for metrics, we measure the disparity error distribution and the rate of correct identification for each attack scenario.

\subsubsection{Results}

We present the experimental results in Fig.~\ref{fig:iden_exp_d}, Fig.~\ref{fig:iden_exp_r}, and Table~\ref{tbl:iden_exp_r}. In Fig.~\ref{fig:iden_exp_d}, the three rows of sub-figures represent the distributions of disparity error $E_{0,1,3}$ (1st row), disparity error $E_{0,2,3}$ (2nd row), and disparity error $E_{1,2,3}$ (3rd row), respectively. And the columns in Fig.~\ref{fig:iden_exp_d} represent five scenarios.

In Fig.~\ref{fig:iden_exp_d}, we can first compare the disparity errors in each row. Similar to the results in the last section, the results clearly illustrate that the distribution of disparity errors can help to detect whether there is an attack. For example, in the first row for $E_{0,1,3}$, when there is no attack on $S_0$, $S_1$, and $S_3$, the disparity errors (cyan bars) are mostly less than $23\%$. By comparison, the disparity errors (red bars) are larger than $23\%$ when any of the sensors is attacked. Such results affirm the  feasibility and correctness of defining attacks according to the disparity error state. Three sub-figures in each of the five attack scenarios clearly show that there is a unique pattern of the combination of $E_{0,1,3}$, $E_{0,2,3}$, and $E_{1,2,3}$ for each attack scenario. For instance, when there is no attack launched, the three disparity errors are all within certain boundaries, representing the disparity error state vector $\textbf{\textit{e}}=[0,0,0]$. And if any one of the sensors is attacked, the disparity errors involving that sensor will exceed boundaries, leading to another unique $\textbf{\textit{e}}$.

In Fig.~\ref{fig:iden_exp_r}, we vary the thresholds by adjusting the designated false alarm rate $r$ and compute the corresponding identification rate in four attack scenarios. It is obvious that, if the attacked sensor is not the reference, then the identification rate drops linearly when $r$ increases from $0$ to $1$. On the other hand, when the reference sensor $S_3$ is attacked, then the identification rate remains very close to $100\%$. 

Such observations suggest that choosing a small $r$ may lead to the best performance overall. To identify the best $r$, we conduct some experiments to investigate the impact of $r$, when it is within the range of $0$ to $5\%$. The numerical results are shown in Table~\ref{tbl:iden_exp_r}. As we can see, the best average identification rate for the four attack scenarios occurs at $r=1\%$.

\vspace{-0.1in}
\subsection{Discussion}

Though the identification method of our framework can accurately identify attacked sensors, it is limited to the condition where no more than $n-2$ sensors are attacked simultaneously in a system with $n+1$ sensors. We plan to address this limitation through cross-vehicle sensing data validation in our future studies.

\begin{figure}[t]
	\centering
	\subfloat[]{
		\label{fig:l_dr}
		\includegraphics[width=0.4\textwidth]{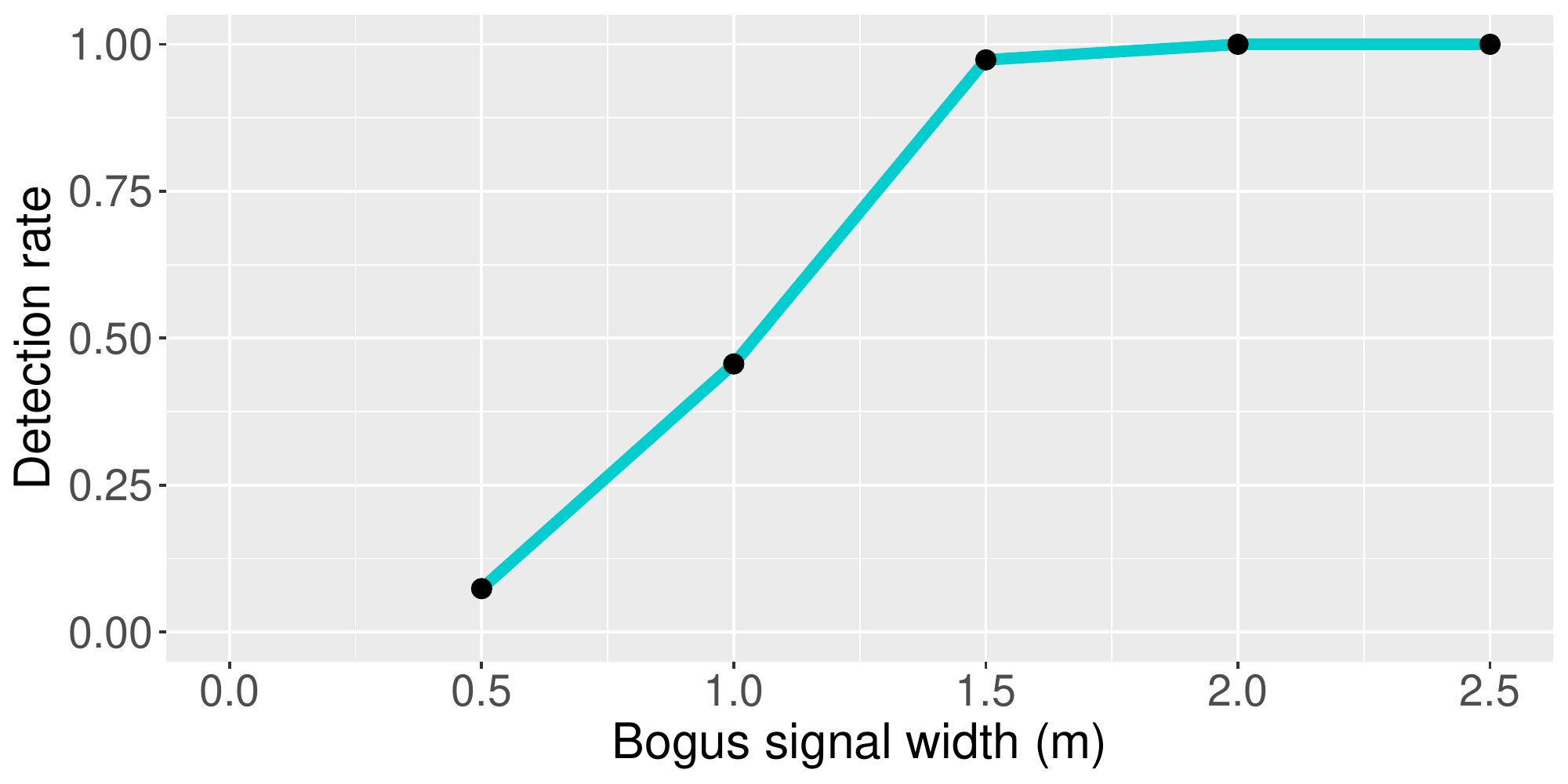}}
	\hfil
	\subfloat[]{
		\label{fig:l_ap}
		\includegraphics[width=0.4\textwidth]{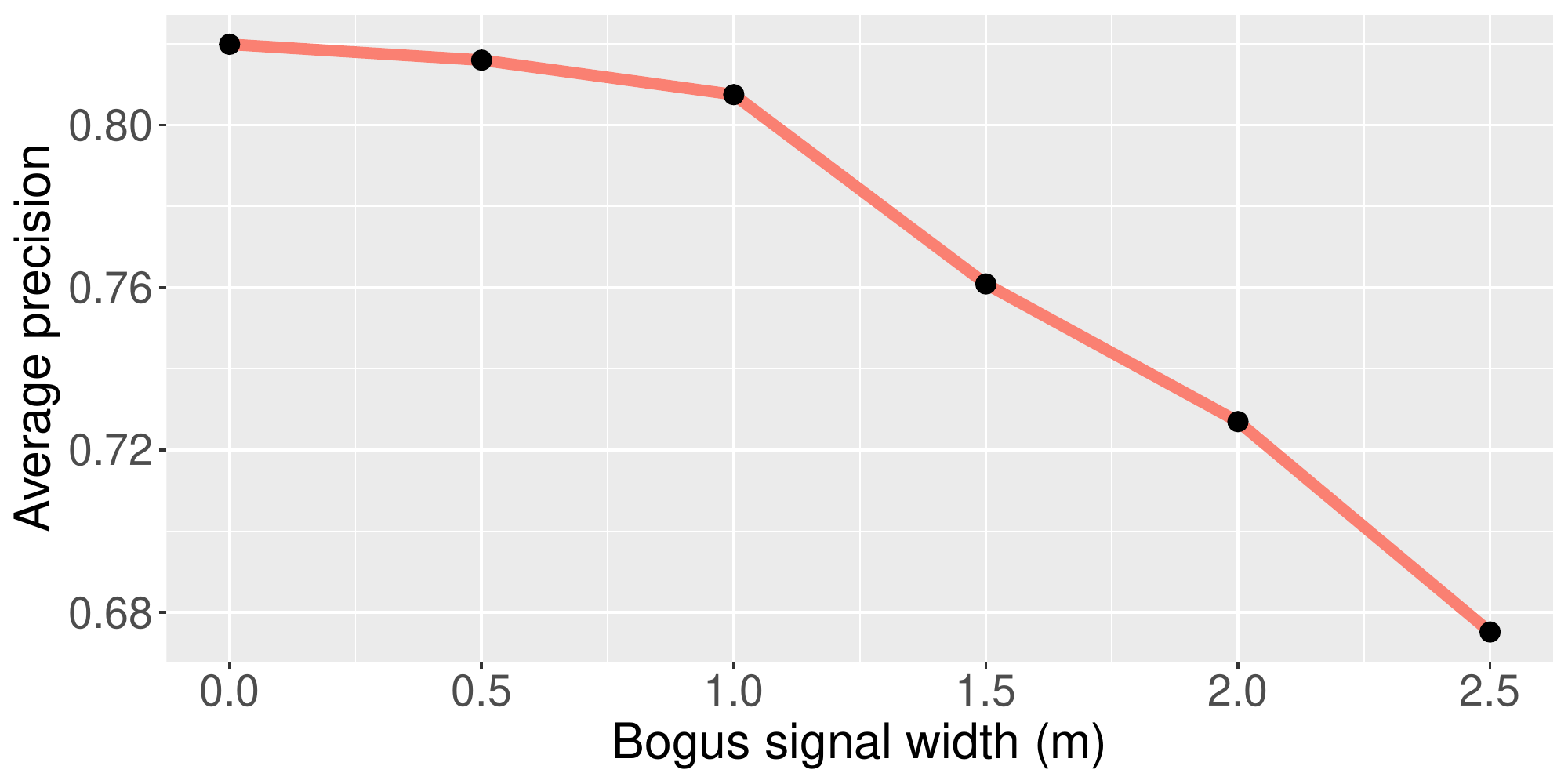}}
	\caption{Sensitivity of our framework and AP of PointRCNN when the attack is against LiDAR. (a)~Detection rate for attacks against LiDAR varies with the width of bogus signal; (b)~Average precision of PointRCNN varies with the width of bogus signal.}
	\label{fig:defense_l}
	\vspace{-0.15in}
\end{figure}

\vspace{-0.05in}
\section{Framework Sensitivity}
\label{sec.sensitivity}

With the best designated false alarm rate $r$ determined, we conduct further experiments to investigate how sensitive our framework is, namely, for optical attacks with what range of settings (width of bogus signal, size of facula) our framework works effectively. Empirically, the milder optical attacks are, the more difficult they get detected. Meanwhile, we also measure how much the perception function of AVs is influenced by the optical attacks with different settings using state-of-the-art object detection algorithms. Those algorithms usually possess a certain degree of resistance to minor optical attacks, so our framework does not have to be universally sensitive.

In this section, we use experiments to demonstrate that our framework has excellent sensitivity to the attacks on LiDAR with settings that object detection model cannot overcome. As for the attacks on camera, our framework is also sensitive in most cases, but shows limit when the attack is too mild. The experiments consist of two parts: the first part is for the sensitivity to the attacks on LiDAR, and the second part for the sensitivity to the attacks on camera.

\vspace{-0.15in}
\subsection{Metrics}

To measure the sensitivity, we use the detection rate of our framework with $r=1\%$, since detection rate can also reflect the performance of identification procedure. As described in Section~\ref{sec.identification}, the identification procedure of our framework is directly developed upon multiple detection processes, so the identification rate is highly correlated with the detection rate.

As for evaluating the corresponding performance of object detection algorithms used in AVs, we follow the KITTI object detection benchmark~\cite{geiger12are} and calculate the average precision of vehicle detection with IoU threshold equal to $70\%$.

\begin{figure}[!t]
	\centering
	\subfloat[]{
		\label{fig:c_dr}
		\includegraphics[width=0.4\textwidth]{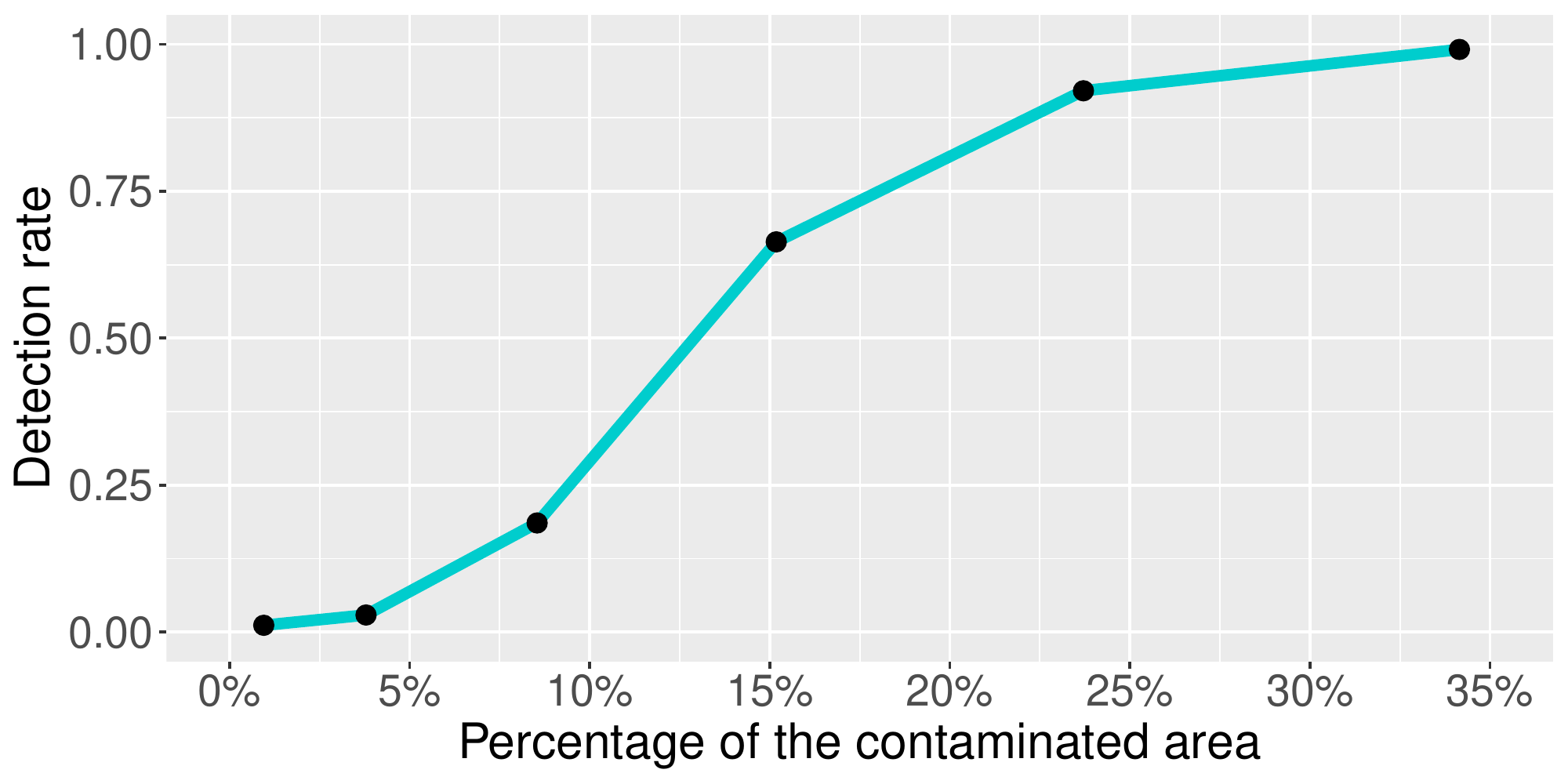}}
	\hfil
	\subfloat[]{
		\label{fig:c_ap}
		\includegraphics[width=0.4\textwidth]{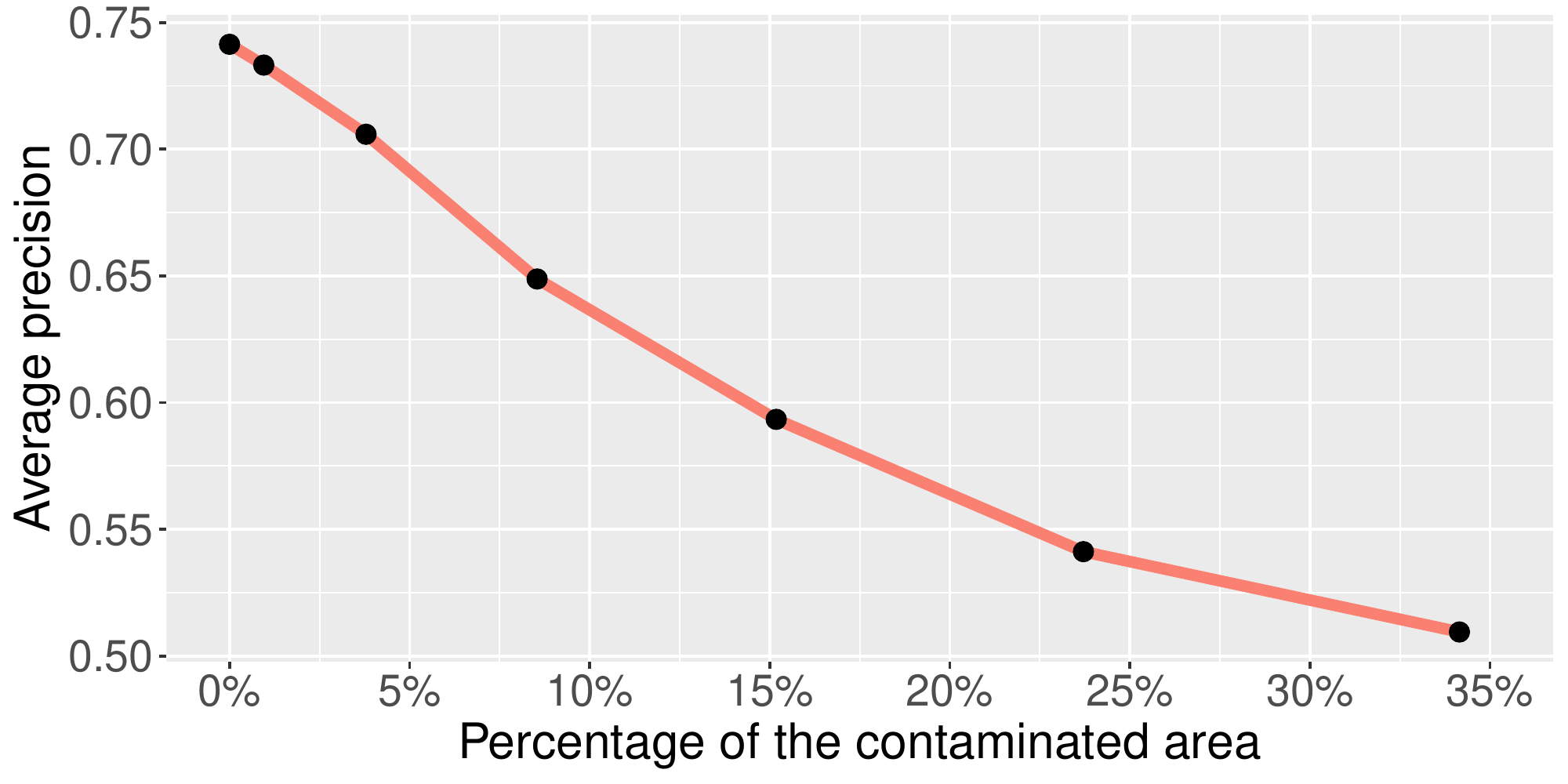}}
	\caption{Sensitivity of our framework and AP of YOLOv3 when the attack is against camera. (a)~Detection rate for attacks against camera varies with the percentage of the area contaminated by facula; (b)~Average precision of YOLOv3 varies with the percentage of the area contaminated by facula.}
	\label{fig:defense_c}
	\vspace{-0.15in}
\end{figure}

\vspace{-0.15in}
\subsection{Experimental Setup}

We conduct our experiments on the dataset provided in the KITTI object detection benchmark~\cite{geiger12are} which contains sensor data of one LiDAR and two cameras. As described in Section~\ref{subsubsec:setup}, we divide the labeled part of the dataset into training set and test set. The training set is used to train the object detection models, while the test set is for generating compromised sensor data.

To find out the sensitivity of our framework to the optical attacks on LiDAR, we produce five affected point clouds for every set of sensor data in the test set. The five affected point clouds contain a bogus signal with a width of 0.5 meter, 1.0 meter, 1.5 meters, 2.0 meters, and 2.5 meters, respectively. And the object detection algorithm chosen for this part of experiments is PointRCNN~\cite{shi19pointrcnn}, a state-of-the-art 3D object detection algorithm that takes a point cloud as input.

In term of the experiment setup for evaluating sensitivity to the attacks on camera, we generate six pairs of compromised stereo pictures for each set of sensor data in the test set. The left pictures of the six pairs are overlaid with a Gaussian facula with radius of 37.5 pixels, 75 pixels, 112.5 pixels, 150 pixels, 187.5 pixels, and 225 pixels, respectively. And the corresponding percentages of the contaminated area in images are 0.95\%, 3.79\%, 8.54\%, 15.18\%, 23.71\%, and 34.15\%. The object detection algorithm for this part of experiments is YOLOv3~\cite{redmon18yolov3}, which is one of the most popular real-time object detection algorithms using images as input.

In the experiments, we feed the compromised sensor data to our framework and the selected object detection model, and then evaluate them via the aforementioned metrics. The PSMNet model used in the framework is provided by~\cite{wang19pseudo}.

\vspace{-0.15in}
\subsection{Experiment Results}

\subsubsection{Sensitivity to the Attacks on LiDAR}

As shown in Fig.~\ref{fig:defense_l}, with the increase of the width of the bogus signal, the detection rate of our framework surges, while the average precision of PointRCNN declines. The AP of PointRCNN decreases very slightly when the width of the bogus signal is within $1.0$ meter, which implies that PointRCNN exhibits some resistance to minor disturbing signals. On the other hand, when the size of the bogus signal is larger, the average precision starts dropping rapidly. However, it should be noted that when the width is at $1.5$ meters, the total decline of AP is only $0.06$, while the detection rate of our framework already reaches nearly $100\%$. These results show clearly that our framework is highly sensitive to the attacks on LiDAR that cannot be mitigated by object detection algorithms.

\subsubsection{Sensitivity to the Attacks on Camera}

The experiment results for this part are illustrated in Fig.~\ref{fig:defense_c}. The tendencies of the detection rate of our framework and the AP of object detection model are similar as those in the first part of the experiments. Particularly, when the percentage of the contaminated area is within the range of $5\%$ to $15\%$, although our framework has a small detection rate, the AP of YOLOv3 maintains at a high level, which means that the perception has not been compromised due to small attacks. When the percentage of the contaminated area in images increases to $23.71\%$, the detection rate of our framework surpasses $90\%$. In the meantime, the AP of YOLOv3 drops significantly to $54.12\%$. These results suggest that our framework has a strong sensitivity to the attacks on camera when the contaminated area in images is greater than $20\%$. Less than that, the framework may show some limitations.

\vspace{-0.1in}
\section{Conclusion} \label{sec.conclusion}

In this paper, we have systematically investigated the mitigation of attacks on optical devices (LiDAR and camera) that are essential to perform accurate object and event detection and response (OEDR) tasks in autonomous driving systems. Specifically, we proposed a framework to detect and identify sensors that are under attack. For the attack detection, we considered two common three-sensor systems, (1) one LiDAR and two cameras, and (2) three cameras, and we developed effective procedures to detect any attack on each of them. Using real datasets and the state-of-the-art machine learning model, we conducted extensive experiments confirming that our detection scheme can detect attacks with high accuracy and a low false alarm rate. Based on the detection models, we further developed an identification model that is capable of identifying up to $n-2$ attacked sensors in a system with one LiDAR and $n$ cameras. For the identification procedure, we proved its correctness and used experiments to validate its performance. At last, we investigated the sensitivity of our framework and showed its excellence in this aspect.

In the future, we plan to study methods to further identify the damaged portion in image and point cloud and perform data recovery for the damaged portion using intact data from other sensors of the ego-vehicle, nearby infrastructure, or vehicles in vicinity. We also plan to investigate the identification task for autonomous driving system based on Multi-Access Edge Computing (MEC) and 5G in which sensor data of \emph{multiple} vehicles can be exploited to overcome the limitation of our proposed identification solution.

\vspace{-0.1in}
\bibliographystyle{IEEEtran}
\bibliography{auto-driving_2019}

\end{document}